\documentclass{article} 

\usepackage{amssymb,amsfonts,amsmath,amsthm,amscd,dsfont,mathrsfs}
\usepackage{graphicx,float,psfrag,epsfig}
\usepackage{wrapfig}
\usepackage{pgf,pgfarrows,pgfnodes}

\DeclareMathAlphabet{\mathpzc}{OT1}{pzc}{m}{it}

\usepackage{epstopdf}
\usepackage{caption}
\usepackage{subcaption}
\usepackage{array}
\newcolumntype{C}[1]{>{\centering\let\newline\\\arraybackslash\hspace{0pt}}m{#1}}

\newtheorem{propo}{Proposition}
\newtheorem{lemma}[propo]{Lemma}
\newtheorem{definition}[propo]{Definition}

\newtheorem{corollary}[propo]{Corollary}

\newtheorem{theorem}[propo]{Theorem}

\def\L{\mathcal{L}}
\def\cG{\mathcal{G}}
\def\cL{\mathcal{L}}
\def\Lrb{\mathcal{L_{\rm RB}}}
\def\I{\mathbb{I}}

\def\i{i^\prime}
\def\j{j^\prime}

\def\k{k^\prime}

\def\deg{{\rm deg}}

\def\cS{{\mathcal{S}}}
\def\cP{{\mathcal{P}}}
\def\Tr{{ \rm Tr}}
\def\diag{{\rm diag}}
\def\E{{\mathbb{E}}}
\def\P{{\mathbb{P}}}
\def\H{{\mathcal{H}}}
\def\reals{{\mathbb{R}}}
\def\prob{{\mathbb{P}}}
\def\<{\langle}
\def\>{\rangle}
\def\Gja{G_{j,a}}

\def\ltheta{{\widetilde{\theta}}}
\def\lOmega{{\widetilde{\Omega}}}
\def\le{{\widetilde{e}}}
\def\lM{{\widetilde{M}}}
\def\lL{{\widetilde{L}}}
\def\lalpha{{\widetilde{\alpha}}}
\def\lW{{\widetilde{W}}}
\def\ld{{\widetilde{d}}}

\newcommand{\norm}[1]{\|#1\|}
\newcommand{\vect}[1]{\boldsymbol{#1}} 
\newcommand{\ceil}[1]{\left \lceil{#1} \right \rceil}
\newcommand{\floor}[1]{\left \lfloor{#1} \right \rfloor}

\begin{document}
\date{}
\title{Data-driven Rank Breaking for Efficient Rank Aggregation}

\author{
{Ashish Khetan and Sewoong Oh }\\
{Department of ISE, University of Illinois at Urbana-Champaign}\\
{Email: $\{$khetan2,swoh$\}$@illinois.edu}
}

\maketitle

\begin{abstract}
 Rank aggregation systems collect ordinal preferences from individuals to produce a global ranking that represents the social preference.  
 Rank-breaking is a common practice to reduce the computational complexity of learning the global ranking. 
 The individual preferences are broken into pairwise comparisons and 
 applied to efficient algorithms tailored for independent paired comparisons. 
 However, due to the ignored dependencies in the data, 
 naive rank-breaking approaches can result in inconsistent estimates. 
 The key idea to produce accurate and  consistent estimates is to treat the pairwise comparisons unequally, depending on the topology of the collected data. 
In this paper, we provide the optimal rank-breaking estimator, which not only achieves consistency but also achieves the best error bound.  
This allows us to characterize the fundamental tradeoff between accuracy and complexity. 
Further, the analysis identifies how the accuracy depends on the spectral gap of a corresponding comparison graph. 

\end{abstract}

\section{Introduction} 
\label{sec:intro} 

In several applications such as  electing officials, choosing policies, or making recommendations, we are 
given partial preferences from individuals over a set of alternatives, with the goal of producing a global ranking that represents the collective preference of the population or the
society. 
This process is referred to as {\em rank aggregation}. 
One popular approach is {\em learning to rank}. 
Economists have modeled each individual as 
a rational being maximizing his/her perceived utility. 
Parametric probabilistic models, known collectively as Random Utility Models (RUMs), 
have been proposed to model such individual choices and preferences \cite{McF80}. 
This allows one to infer the    global ranking  
by learning the inherent utility from individuals' revealed preferences, 
which are noisy manifestations of the underlying true utility of the alternatives.

Traditionally,  learning to rank has been studied under 
the following  data collection scenarios:  
 pairwise comparisons,  best-out-of-$k$ comparisons, and 
$k$-way comparisons. 
{\em Pairwise comparisons} are commonly studied in the classical context of sports matches as well as more recent applications in crowdsourcing, where each worker is presented with a  pair of choices and asked to choose the more favorable one. 
{\em Best-out-of-$k$ comparisons} data sets are commonly available from purchase history of customers. 
Typically, a set of $k$ alternatives are offered among which one is chosen or purchased
 by each customer. 
This has been widely studied in operations research in the context of modeling 
customer choices  for revenue management and assortment optimization. 
The {\em $k$-way comparisons} are assumed in traditional rank aggregation scenarios,  where 
each person reveals his/her preference as a ranked list over a set of $k$ items. 
In some real-world elections, voters provide ranked preferences over the whole set of candidates \cite{Lun07}. 
We refer to these three types of ordinal data collection scenarios as `traditional' throughout this paper.
 
For such traditional data sets, 
there are several computationally efficient inference algorithms for finding the Maximum Likelihood (ML) estimates 
that provably achieve the minimax optimal performance \cite{NOS12,SBB15,HOX14}. 
However, modern data sets can be unstructured. 
Individual's revealed ordinal preferences can be 
 implicit, such as movie ratings, time spent on the news articles, and 
 whether the user finished watching the movie or not. 
In crowdsourcing, it has also been observed that humans are more efficient at performing batch comparisons \cite{NIPS2011_4187}, 
as opposed to providing the full ranking or choosing the top item. 
This calls for more flexible approaches for rank aggregation that can 
take such diverse forms of ordinal data into account. 
For such non-traditional data sets, 
finding the ML estimate can become significantly more challenging, requiring run-time exponential in the 
problem parameters.

To avoid such a computational bottleneck, a common heuristic  is to resort to {\em rank-breaking}.  
The collected ordinal data is first transformed into a bag of pairwise comparisons, 
ignoring the dependencies that were present in the original data.  
This is then processed via existing inference algorithms tailored for {\em independent} pairwise comparisons, 
hoping that the dependency present in the input data does not lead to inconsistency in estimation. 
This idea is one of the main motivations 
 for numerous approaches  specializing in learning to rank from 
pairwise comparisons, e.g., \cite{Ford57,NOS14,ACPX13}. 
However, such a heuristic of full rank-breaking defined explicitly in \eqref{eq:fullrankbreaking},
where all pairwise comparisons are weighted and treated equally ignoring their dependencies, 
has been recently shown to introduce inconsistency  \cite{APX14a}. 

The key idea to produce accurate and consistent estimates is to treat the pairwise comparisons unequally, depending on the topology of the collected data. 
A fundamental question of interest to practitioners is 
how to choose the weight of each pairwise comparison in order to achieve not only consistency but also the best accuracy, among those consistent estimators using rank-breaking. 
We study how the accuracy of resulting estimate depends on the topology of the data and 
the weights on the pairwise comparisons. 
This provides a guideline for the optimal choice of the weights, driven by the topology of the data, 
that leads to accurate estimates. 

\bigskip
{\bf Problem formulation.} 
The premise in the current race to  collect  more data on  user activities
is that, a hidden true preference manifests in the user's  activities and choices. 
Such data can be explicit, as in ratings, ranked lists, pairwise comparisons, and like/dislike buttons.  
Others are more implicit, such as purchase history and viewing times. 
While more data in general allows for a more accurate inference,  
the heterogeneity of user activities makes it difficult to infer the underlying preferences directly. 
Further, each user reveals her preference on only  a few contents. 

Traditional collaborative filtering fails to capture the diversity of modern data sets. 
The sparsity and heterogeneity of the data renders typical  similarity measures ineffective in the nearest-neighbor methods. 
Consequently, simple measures of similarity prevail in practice, as in Amazon's ``people who bought ... also bought ...'' scheme. 
Score-based methods require translating heterogeneous data into numeric scores, which is a priori a difficult task. 
Even if explicit ratings are observed, those are often unreliable and the scale of such ratings vary from user to user.  

We propose aggregating ordinal data based on users' revealed preferences that 
are expressed in the form of {\em partial orderings} 
(notice that our use of the term is slightly different from its original use in revealed preference theory). 
We interpret user activities as manifestation of the hidden  preferences according to discrete choice models (in particular the Plackett-Luce model defined in \eqref{eq:rum}). 
This provides a  
more reliable,  scale-free, and widely applicable representation of the heterogeneous data as partial orderings, as well as  
 a probabilistic interpretation of how preferences manifest. 
In full generality, the data collected from each individual can be represented by a {\em partially ordered set (poset)}.
Assuming consistency in a user's revealed preferences, 
any ordered relations can be seamlessly translated into a poset, 
represented as a Hasse diagram by a directed acyclic graph (DAG). 
The DAG below represents ordered relations $a>\{b,d\}$, $b>c$, $\{c,d\}>e$, and $e>f$. 
For example, this could have been translated from two sources: 
a five star rating on $a$ and a three star ratings on $b,c,d$, a two star rating on $e$, and a one star rating on $f$; 
and the item $b$ being purchased after reviewing $c$ as well.  
\begin{figure}[h]
	\begin{center}
	\includegraphics[width=.2\textwidth]{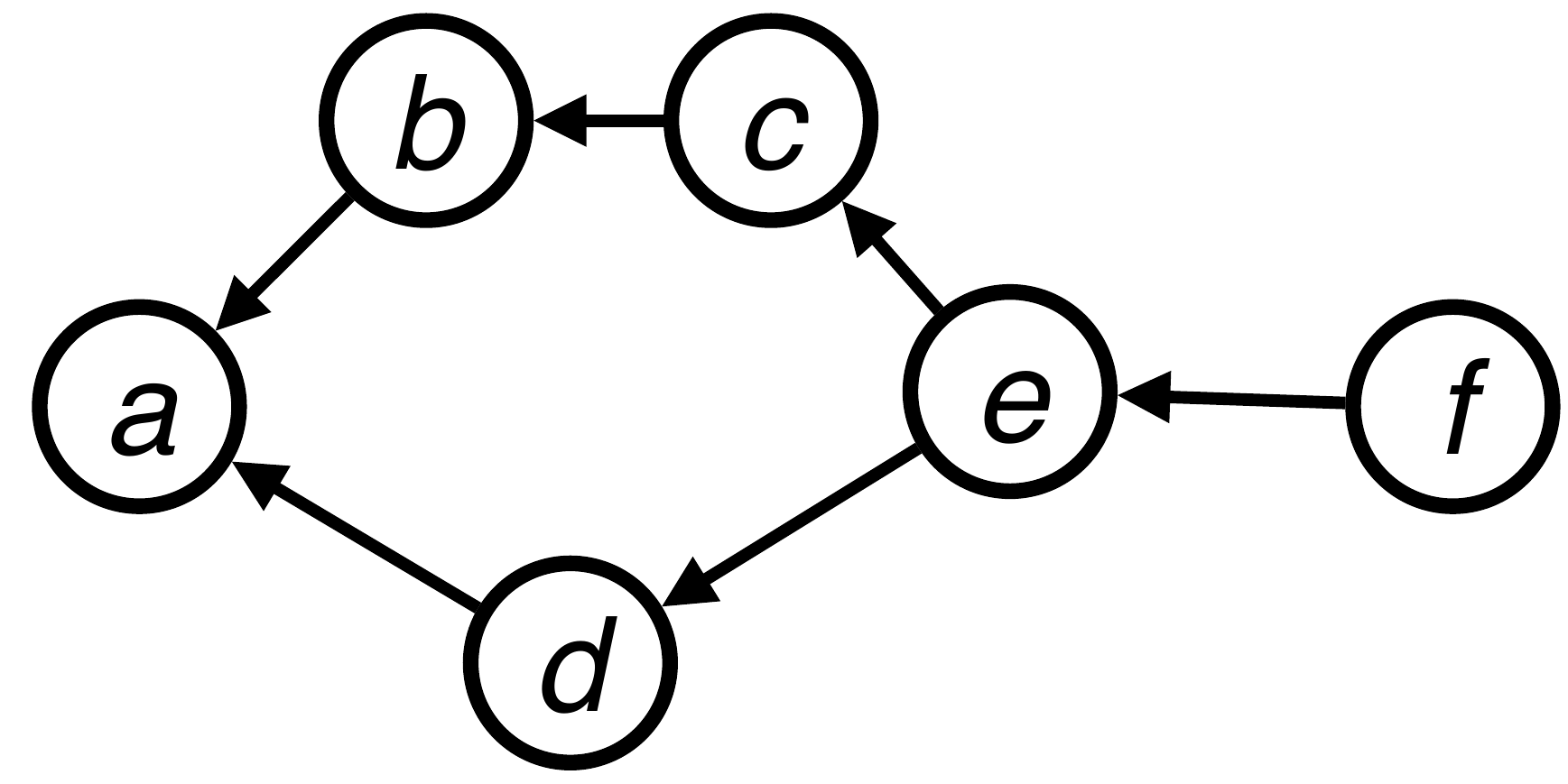}
	\put(-50,60){${\cal G}_j$}
	\hspace{1cm}
	\includegraphics[width=.3\textwidth]{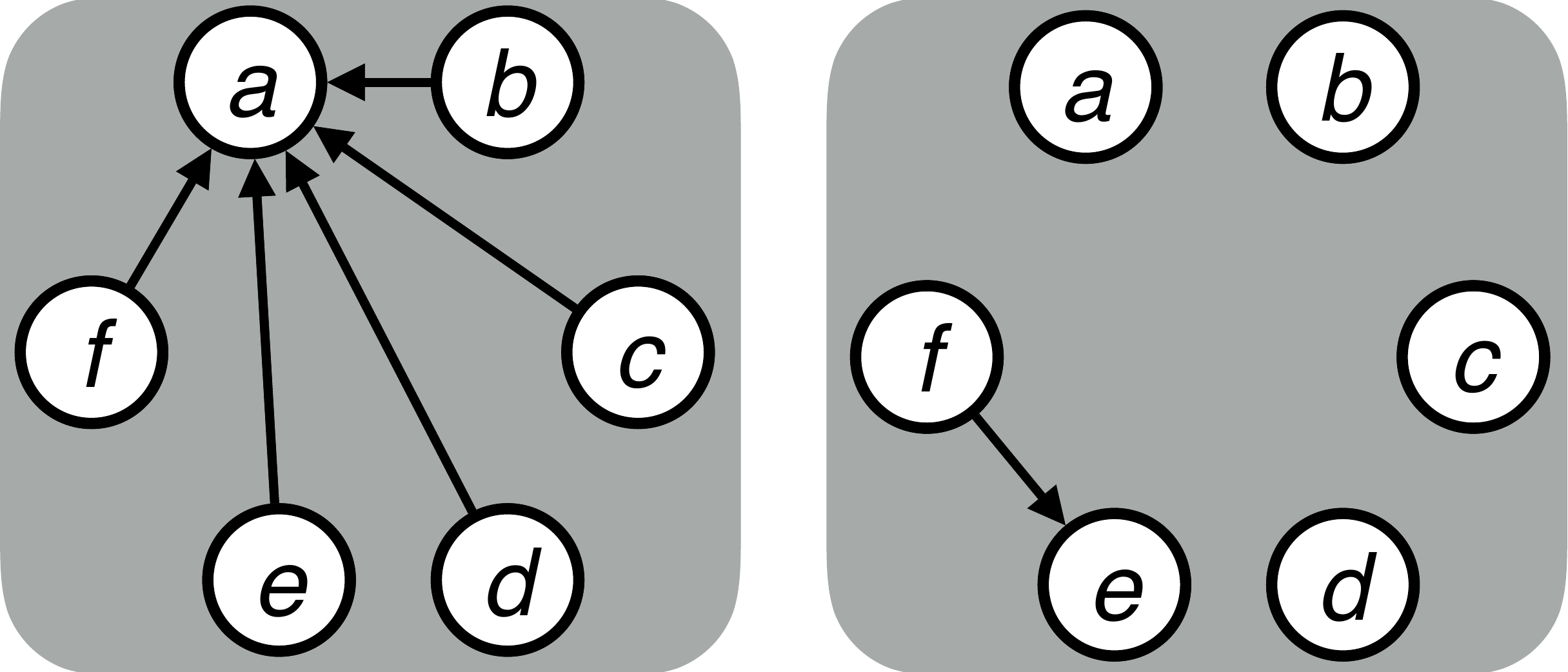}
	\put(-115,70){$G_{j,1}$}
	\put(-40,70){$G_{j,2}$}
	\end{center}
	\caption{ A DAG representation of consistent partial ordering of a user $j$, also called a Hasse diagram (left). 
	A set of rank-breaking graphs extracted from the Hasse diagram for the separator item $a$ and $e$, respectively (right).}
	\label{fig:hasse}
\end{figure}

There are $n$ users or agents, and each agent $j$  provides his/her ordinal evaluation on a subset $S_j$ of 
$d$ items or alternatives. We refer to $S_j\subset\{1,2,\ldots,d\}$ as {\em offerings} provided to $j$, and use $\kappa_j=|S_j|$ to denote the 
size of the offerings.  
We assume that the partial ordering over the offerings is 
a manifestation of her preferences as per a popular choice model known as Plackett-Luce (PL) 
model. 
As we explain in detail below,  the PL model produces total orderings (rather than partial ones).
The data collector queries each user for a partial ranking in the form of a poset over $S_j$. 
  For example, the data collector can ask for 
  the top item, unordered subset of three next preferred items, the fifth item, and the least preferred item. In this case, an example of such poset could be 
  $a < \{b,c,d\} < e < f$, which could have been generated from a total ordering produced by the PL model and taking the corresponding partial ordering from the total ordering. 
  Notice that we fix the topology of the DAG first and ask the user to fill in the node identities corresponding to her total ordering as (randomly) generated by the PL model. 
  Hence, the structure of the poset is considered deterministic, 
  and only the identity of the nodes in the poset is considered random. 
Alternatively, one could consider a different scenario where the topology of the poset is also random and depends on the outcome of the preference, which is out-side the scope of this paper and provides an interesting future research direction.

The PL model  is  a special case of {\em random utility models}, defined as follows \cite{WB02,APX12}. 
Each item $i$ has a real-valued latent utility $\theta_i$. 
When presented with a set of items, 
a user's reveled preference is a partial ordering according to 
noisy manifestation of the utilities, i.e. i.i.d. noise added to the true utility $\theta_i$'s. 
The PL model is a special case where the noise follows the standard Gumbel distribution, and is one of the most popular model in social choice theory \cite{McF73,MT00}. 
PL has several important properties,  
making this model realistic in various domains, including marketing 
\cite{GL83}, transportation \cite{McF80,BL85}, biology \cite{BIOLOGY}, and natural language processing \cite{MCCD13}. 
Precisely, each user $j$, when presented with a set $S_j$ of items, 
draws a noisy utility of each item $i$ according to
\begin{eqnarray*}
	u_i &=& \theta_i + Z_i\;, \label{eq:rum}
\end{eqnarray*}
where $Z_i$'s follow the independent standard Gumbel distribution. 
Then we observe the ranking resulting from sorting the items as per noisy observed utilities $u_j$'s. 
Alternatively, the PL model is also equivalent to 
the following random process. 
For a set of alternatives $S_j$, a ranking $\sigma_j:[|S|] \to S$ is generated 
in two steps: $(1)$ independently assign each item $i \in S_j$ an unobserved value $X_i$, exponentially distributed with mean $e^{-\theta_i}$; $(2)$ select a ranking $\sigma_j$ so that $X_{\sigma_j(1)} \leq X_{\sigma_j(2)} \leq \cdots \leq X_{\sigma_j(|S_j|)}$.

The PL model 
$(i)$ satisfies Luce's `independence of irrelevant alternatives' in social choice theory \cite{Ray73}, and has a simple characterization as sequential (random) choices as explained below; and
$(ii)$ has a maximum likelihood estimator (MLE) which is a convex program in $\theta$ in the traditional scenarios of pairwise, best-out-of-$k$ and $k$-way comparisons.
Let $\prob(a>\{b,c,d\})$ denote the probability $a$ was chosen as the best alternative among the set $\{a,b,c,d\}$. 
Then, the probability that a user reveals a linear order $(a>b>c>d)$ is 
equivalent as making sequential choice from the top to bottom: 
\begin{eqnarray*}
	\prob(a>b>c>d) &=& \prob(a>\{b,c,d\})\, \,\prob(b>\{c,d\})\, \,\prob(c>d) \nonumber \\ 
		&=&	\frac{e^{\theta_a}}{(e^{\theta_a}+e^{\theta_b}+e^{\theta_c}+e^{\theta_d} ) } \, 
		\frac{e^{\theta_b}}{(e^{\theta_b}+e^{\theta_c}+e^{\theta_d} )} \, 
		\frac{e^{\theta_c}}{(e^{\theta_c}+e^{\theta_d})} \, \;.
\end{eqnarray*}
We use the notation $(a>b)$ to denote the event that $a$ is preferred over $b$. 
In general, for user $j$ presented with offerings $S_j$, the probability that the revealed preference is a 
total ordering $\sigma_j$ is 
$\prob(\sigma_j) = \prod_{i\in \{1,\ldots,\kappa_j-1\}}   (e^{\theta_{\sigma^{-1}(i)}})/(\sum_{i'=i}^{\kappa_j}e^{\theta_{\sigma^{-1}(i')}}) $. 
We consider the true utility $\theta^* \in \Omega_b$, where we define    $\Omega_b$ as 
\begin{eqnarray*}
	\Omega_b &\equiv& \Big\{ \, \theta \in \reals^d \,\big|\, \sum_{i\in[d]} \theta_i=0 \,,\, |\theta_i| \leq b \text{ for all $i\in[d]$ } \,\Big\} \;.
\end{eqnarray*}
Note that by definition, the PL model is invariant under shifting the 
utility $\theta_i$'s. Hence,  the centering ensures uniqueness of the parameters for each PL model. 
The bound $b$ on the dynamic range is not a restriction, but is written explicitly to 
capture the dependence of the accuracy  in our main results. 

We have $n$ users each providing a partial ordering of a set of offerings $S_j$ 
according to the PL model. Let ${\cG}_j$ denote both the DAG representing the partial ordering 
from user $j$'s preferences. With a slight abuse of notations, we also let $\cG_j$ denote 
 the set of rankings that are consistent with this DAG. 
For general partial orderings, the probability of observing $\cG_j$ is the sum of all total orderings that is consistent with the observation, i.e.  $\prob(\cG_j)=\sum_{\sigma \in \cG_j} \prob(\sigma)$.
The goal is to efficiently learn the true utility $\theta^*\in\Omega_b$, from the $n$ sampled partial orderings.
One popular approach is to compute the maximum likelihood estimate (MLE) by solving the 
following optimization: 
\begin{eqnarray*}
	\underset{\theta \in \Omega_b }{\text{maximize}} && \sum_{j=1}^n \; \log \prob(\cG_j) \;. 
	\label{eq:mle}
\end{eqnarray*}
 This optimization is a simple convex optimization, in particular a logit regression, when the structure of the data $\{\cG_j\}_{j\in[n]}$ is traditional. 
This is  one of the reasons the PL model is attractive. 
However, for general posets, this can be computationally challenging. 
Consider an example of position-$p$ ranking, where each user provides which item is at $p$-th position in his/her ranking. 
Each term in the log-likelihood for this data involves summation over $O((p-1)!)$ rankings, 
which takes $O(n\,(p-1)!)$ operations to  evaluate the objective function. Since $p$ can be as large as $d$, 
such a computational blow-up renders MLE approach impractical. 
A common remedy is to resort to rank-breaking, which might result in inconsistent estimates.

\bigskip
{\bf Rank-breaking.} 
	Rank-breaking refers to the idea of extracting a set of pairwise comparisons from the observed 
	partial orderings and applying estimators tailored for paired comparisons treating each piece of comparisons as independent. 
	Both the choice of which paired comparisons to extract and 
	the choice of parameters in the estimator, which we call {\em weights}, turns out to be 
	crucial as we will show. 
	Inappropriate selection of the paired comparisons can 
	lead to inconsistent estimators as proved in \cite{APX14a}, and 
	the standard choice of the parameters can lead to a significantly suboptimal performance. 
	
	A naive rank-breaking that is widely used in practice is to apply rank-breaking to all possible pairwise relations that one can read from the partial ordering and weighing them equally. 
	We refer to this practice as {\em full rank-breaking}. In the example in Figure \ref{fig:hasse}, 
	full rank-breaking first extracts 
	the bag of comparisons ${\cal C}=\{(a>b),(a>c),(a>d),(a>e),(a>f),\ldots,(e>f)\}$ with 13 paired comparison outcomes, and 
	apply the maximum likelihood estimator treating each paired outcome as independent. Precisely, 
	the {\em full rank-breaking  estimator} solves the convex optimization of 
	\begin{eqnarray}
	\label{eq:fullrankbreaking}
	\widehat\theta &\in & \arg\max_{\theta \in \Omega_b} \;
	\sum_{ (i>i') \in {\cal C} }   \Big(\theta_i   - \log \Big(e^{\theta_i} + e^{\theta_{\i}}\Big) \Big)\;.
	\end{eqnarray} 
There are several efficient implementation tailored for this problem \cite{Ford57,Hun04,NOS12,MG15}, and 
under the traditional scenarios, these approaches provably achieve 
the minimax optimal rate \cite{HOX14,SBB15}.
For general non-traditional data sets, 
there is a significant gain in computational complexity. 
In the case of position-$p$ ranking, where each of the $n$ users 
report his/her $p$-th ranking item among $\kappa$ items, the computational complexity reduces from $O(n\,(p-1)!)$ for the MLE in \eqref{eq:mle} 
to 
$O( n\,p\,(\kappa-p))$ for the full rank-breaking estimator in \eqref{eq:fullrankbreaking}. 
However, this gain comes at the cost of accuracy. 
It is known that  the full-rank breaking estimator is inconsistent \cite{APX14a}; 
the error is strictly bounded away from zero even with infinite samples. 
	
Perhaps surprisingly, Azari Soufiani et al. 
\cite{APX14a} recently characterized 
the entire set of consistent rank-breaking estimators. 
Instead of using the bag of paired comparisons, 
the sufficient information for consistent rank-breaking is a set of rank-breaking graphs defined as follows.

Recall that a user $j$ provides his/her preference as a  poset represented by a DAG $\cG_j$. 
Consistent rank-breaking first identifies all {\em separators} in the DAG. 
A node in the DAG is a separator if one can partition the rest of the nodes  into two parts. 
A partition $A_{\rm top}$ which is the set of items that are preferred over the separator item, and 
a partition $A_{\rm bottom}$ which is the set of items that are less preferred than the separator item. 
One caveat is that we allow $A_{\rm top}$ to be empty, but $A_{\rm bottom}$ must have at least one item. 
In the example in Figure \ref{fig:hasse}, there are two separators: the item $a$ and the item $e$. 
Using these separators, one can extract the following partial ordering from the original poset: $(a>\{b,c,d\}>e>f)$. 
The items $a$ and $e$ separate the set of offerings into partitions, hence the name separator. 
We use $\ell_j$ to denote the number of separators in the poset $\cG_j$ from user $j$. 
We let $p_{j,a}$ denote the ranked position of the $a$-th separator in the poset $\cG_j$, 
and we sort the positions such that $p_{j,1} < p_{j,2} < \ldots < p_{j,\ell_j}$.
The set of separators is denoted by $\cP_j = \{p_{j,1},p_{j,2},\cdots,p_{j,\ell_j}\}$. 
For example, since the separator $a$ is ranked at position 1 and $e$ is at the $5$-th position, 
$\ell_j=2$, $p_{j,1}=1$, and $p_{j,2}=5$.   Note that $f$ is not a separator (whereas $a$ is) since 
corresponding $A_{\rm bottom}$ is empty. 

Conveniently, we represent this extracted partial ordering using a set of DAGs, 
which are called {\em rank-breaking graphs}. 
We generate one rank-breaking graph per separator. 
A rank breaking graph $G_{j,a}=(S_j,E_{j,a})$ for user $j$ and the $a$-th separator is defined as  
a directed graph over the set  of offerings $S_j$, where 
we add an edge from a node that is less preferred than the $a$-th separator to the separator, i.e. 
$E_{j,a}=\{(i,i') \,|\, i'\text{ is the $a$-th separator, and } \sigma_j^{-1}(i) > p_{j,a} \}$.
Note that by the definition of the separator, $E_{j,a}$ is a non-empty set. 
An example of rank-breaking graphs are shown in Figure \ref{fig:hasse}.

This rank-breaking graphs were introduced in \cite{ACPX13}, 
where it was shown that 
the pairwise ordinal relations that is represented by edges in the rank-breaking graphs 
are sufficient information for using any estimation based on the idea of rank-breaking.  
Precisely, on the converse side, it was proved in \cite{APX14a} that any pairwise outcomes  that 
is not present in the rank-breaking graphs $G_{j,a}$'s  lead to inconsistency for a general $\theta^*$.  On the achievability side, it was proved that all pairwise outcomes that are present in the rank-breaking graphs give a consistent estimator, 
as long as all the paired comparisons 
in each $G_{j,a}$ are weighted equally. 

It should be noted that rank-breaking graphs are defined slightly differently in \cite{ACPX13}. 
Specifically, \cite{ACPX13} introduced a different notion of rank-breaking graph, where  the vertices represent positions in total ordering. 
An edge between two vertices $i_1$ and $i_2$ denotes that the  pairwise comparison between items ranked at position $i_1$ and $i_2$ is included in the estimator. 
Given such observation from the PL model, 
\cite{ACPX13} and \cite{APX14a} prove that a rank-breaking graph is consistent if and only if it satisfies the following property. If a vertex $i_1$ is connected to any vertex $i_2$, where $i_2 > i_1$, then $i_1$ must be connected to all the vertices $i_3$ such that $i_3 > i_1$.
Although the specific definitions of rank-breaking graphs are different from our setting, 
the mathematical analysis of \cite{ACPX13} still holds when interpreted appropriately. 
Specifically, we consider only those rank-breaking that are consistent under the conditions given in 
\cite{ACPX13}. 
In our rank-breaking graph $G_{j,a}$, a separator node is connected to all the other item nodes that are ranked below it (numerically higher positions).

In the algorithm described in \eqref{eq:likelihood}, we satisfy this sufficient  condition for consistency by restricting to a class of convex optimizations that use 
the same weight $\lambda_{j,a}$ 
for all  $(\kappa-p_{j,a})$ paired comparisons in the objective function, as opposed to allowing more general weights  that defer from a pair to another pair in a rank-breaking graph $G_{j,a}$. 

\bigskip
{\bf Algorithm.}
Consistent rank-breaking first identifies separators in the collected posets $\{\cG_j\}_{j\in[n]}$ 
and transform them into 
rank-breaking graphs $\{ G_{j,a}\}_{j\in[n],a\in[\ell_j]}$ as explained above. 
These rank-breaking graphs are input to the MLE for paired comparisons, 
assuming all directed edges in the rank-breaking graphs 
are independent outcome of pairwise comparisons. 
Precisely, the {\em consistent rank-breaking estimator} solves the convex optimization of maximizing the  paired log likelihoods 
\begin{eqnarray}
	\label{eq:likelihood_0}
	\Lrb(\theta) &=&
	\sum_{j=1}^n \sum_{a = 1}^{\ell_j}
	\,\lambda_{j,a} \, \Big\{ \sum_{(i, \i) \in E_{j,a}}
	 \,	 \Big(  \theta_{\i}  - \log \Big(e^{\theta_i} + e^{\theta_{\i}}\Big) \,\Big)\, \Big\} \;,
\end{eqnarray} 
where $E_{j,a}$'s are defined as above via separators and different choices of the non-negative weights $\lambda_{j,a}$'s are possible and the performance depends on such choices.
Each weight $\lambda_{j,a}$ determine how much we want to weigh the contribution of a 
corresponding rank-breaking graph $G_{j,a}$. 
We define the  {\em consistent rank-breaking estimate} $\widehat\theta$ as  
the optimal solution of the convex program: 
\begin{align} \label{eq:theta_ml}
\widehat{\theta} \;\; \in \;\; \arg\max_{\theta \in \Omega_b} \;\, \Lrb(\theta)\;. 
\end{align}
By changing how we weigh each rank-breaking graph (by choosing the $\lambda_{j,a}$'s), 
the convex program \eqref{eq:theta_ml} spans the entire set of consistent rank-breaking estimators, as characterized   in  \cite{APX14a}. 
However, only asymptotic consistency was known, which holds independent of the choice of the weights $\lambda_{j,a}$'s. 
Naturally, a uniform choice of $\lambda_{j,a}=\lambda$ was proposed in  \cite{APX14a}. 

Note that this can be efficiently solved, since this is a simple convex optimization, in particular a logit regression, with only $O(\sum_{j=1}^n \,\ell_j\, \kappa_j)$ terms. 
For a special case of position-$p$ breaking, the $O(n \, (p-1)!)$ 
complexity of evaluating the objective function  
for the MLE is now significantly reduced to $O(n\,(\kappa-p))$ by rank-breaking. 
Given this potential exponential gain in efficiency,  a natural question of interest is  ``what is the price we pay in the accuracy?''. 
We provide a sharp analysis of the performance of rank-breaking estimators in the finite sample regime, 
that quantifies the price of rank-breaking. 
Similarly, for a practitioner, a core problem of interest is 
how to choose the weights in the optimization in order to achieve  the best accuracy. 
Our analysis provides a data-driven guideline for choosing the optimal weights. 


\bigskip
{\bf Contributions.} 
In this paper, we provide an upper bound on the error achieved by the rank-breaking estimator of \eqref{eq:theta_ml} for any choice of the weights in Theorem \ref{thm:main}.  
This explicitly shows how the error depends on the choice of the weights, 
and  provides a guideline for choosing the optimal weights $\lambda_{j,a}$'s in a data-driven manner.  
We provide the explicit formula for the optimal choice of the weights and provide the 
the error bound in Theorem \ref{thm:main2}. 
The analysis shows the explicit dependence of the error in the problem dimension $d$ and the number of users $n$ that matches the numerical experiments.  

If we are designing surveys and can choose which subset of items to offer to each user and also can decide which type of ordinal data we can collect, then 
we want to design such surveys in a way to maximize the accuracy for a given number of questions asked. 
Our analysis provides how the accuracy depends on the topology of the collected data, and 
 provides a guidance when we do have some control over which questions to ask and which data to collect. 
One  should maximize the spectral gap of corresponding comparison graph. 
Further, for some canonical scenarios, we quantify the price of rank-breaking by comparing the 
error bound of the proposed data-driven rank-breaking  
with the lower bound on the MLE, which can have a  significantly larger computational cost (Theorem \ref{thm:cramer_rao_position_p}).

\bigskip 
\textbf{Notations.} 
{Following is a summary of all the notations defined above. We use $d$ to denote the total number of items and index each item by $i \in \{1,2,\ldots,d\}$. $\theta \in \Omega_b$ denotes vector of utilities associated with each item. $\theta^{*}$ represents true utility and $\widehat{\theta}$ denotes the estimated utility. We use $n$ to denote the number of users/agents and index each user by $j \in \{1,2,\ldots,n\}$. $S_j \subseteq \{1,\ldots,d\}$ refer to the offerings provided to the $j$-th user and we use $\kappa_j = |S_j|$ to denote the size of the offerings. $\cG_j$ denote the DAG (Hasse diagram) representing the partial ordering from user $j$'s preferences.  $\cP_j = \{p_{j,1},p_{j,2},\cdots,p_{j,\ell_j}\}$ denotes the set of separators in the DAG $\cG_j$, where $p_{j,1},\cdots,p_{j,\ell_j}$ are the positions of the separators, and $\ell_j$ is the number of separators. $G_{j,a}=(S_j,E_{j,a})$ denote the rank-breaking graph for the $a$-th separator extracted from the partial ordering $\cG_j$ of user $j$.} 

For any positive integer $N$, let $[N] = \{1,\cdots,N\}$. 
For a ranking $\sigma$ over $S$, i.e., $\sigma$ is a mapping from $[|S|]$ to $S$, let $\sigma^{-1}$ denote the inverse mapping.
 For a vector $x$, let $\norm{x}_2$ denote the standard $l_2$ norm. Let $\vect{1}$ denote the all-ones vector and $\vect{0}$ denote the all-zeros vector with the appropriate dimension. Let $\cS^d$ denote the set of $d \times d$ symmetric matrices with real-valued entries. For $X \in \cS^d$, let ${\lambda_1(X) \leq\lambda_2(X) \leq \cdots \leq \lambda_d(X)}$ denote its eigenvalues sorted in increasing order. Let $\Tr(X) = \sum_{i = 1}^d \lambda_i(X)$ denote its trace and $\norm{X} = \max\{ |\lambda_1(X)|,|\lambda_d(X)| \}$ denote its spectral norm. For two matrices $X,Y \in \cS^d$, we write $X \succeq Y$ if $X-Y$ is positive semi-definite, i.e., $\lambda_1(X-Y) \geq 0$. Let $e_i$ denote a unit vector in $\reals^d$ along the $i$-th direction. 

\section{Comparisons Graph and the Graph Laplacian}

In the analysis of the convex program \eqref{eq:theta_ml}, we show that, with high probability, the 
objective function is strictly concave with $\lambda_2(H(\theta)) \leq - C_b \,\gamma\, \lambda_2(L)< 0$ (Lemma \ref{lem:hessian_positionl}) for all $\theta\in\Omega_b$ and  
the gradient is bounded by $\|\nabla\cL_{\rm RB}(\theta^*)\|_2 \leq C_b' \sqrt{\log d\, \sum_{j\in[n]} \ell_j}$ (Lemma \ref{lem:gradient_topl}). 
Shortly, we will define $\gamma$ and $\lambda_2(L)$, which captures the dependence on the topology of the data, and 
$C_b'$ and $C_b$ are constants that only depend on $b$. 
Putting these together, we will show that there exists a $\theta\in\Omega_b$ such that 
\begin{eqnarray*}
	\|\widehat\theta -\theta^* \|_2 &\leq& \frac{2\|\nabla \cL_{\rm RB}(\theta^*)\|_2 }{-\lambda_2(H(\theta))} \;\, \leq \;\,  C''_b \frac{\sqrt{\log d \, \sum_{j\in[n]} \ell_j}}{\gamma \,\lambda_2(L)} \;. \label{eq:intro_bound}
\end{eqnarray*}
Here $\lambda_2(H(\theta))$ denotes the second largest eigenvalue of a negative semi-definite Hessian matrix $H(\theta)$ of the objective function.  
The reason the second largest eigenvalue shows up is because the top eigenvector is always the all-ones vector which 
by the definition of $\Omega_b$ is infeasible. 
The accuracy depends on the topology of the collected data via the comparison graph of given data. 

\begin{definition} \label{def:comparison_graph1} (Comparison graph $\H$). 
We define a graph $\H([d],E)$ where 
each alternative corresponds to a node, and we put an edge $(i,i')$ 
 if there exists an agent $j$ whose offerings is a set  $S_j$ such that $i, \i \in S_j$. 
Each edge $(i,\i) \in E$ has a weight $A_{i \i}$ defined as 
	\begin{eqnarray*} 
		A_{i\i} &=& \sum_{j\in[n] : i,\i \in S_j} \frac{\ell_j}{\kappa_j(\kappa_j-1)}\;, 
	\end{eqnarray*}
where $\kappa_j = |S_j|$ is the size of each sampled set  and $\ell_j$ is the number of separators in $S_j$ 
 defined by rank-breaking in  Section \ref{sec:intro}. 
 \end{definition}

Define a diagonal matrix $D = {\rm diag}(A\vect{1})$, and 
 the corresponding graph Laplacian $L = D - A$, such that 
	\begin{eqnarray} \label{eq:comparison1_L}
		L &=& \sum_{j = 1}^n \frac{\ell_j}{\kappa_j(\kappa_j-1)}  \sum_{i<\i \in S_j} (e_i - e_{\i})(e_i - e_{\i})^\top.
	\end{eqnarray}
Let $ 0 = \lambda_1(L) \leq \lambda_2(L) \leq \cdots \leq \lambda_d(L)$ denote the (sorted) eigenvalues of $L$. 
Of special interest is $\lambda_2(L)$, also called the spectral gap, which measured how well-connected the graph is. 
Intuitively, one can expect better accuracy when  the spectral gap is larger, as evidenced in previous learning to rank 
results in simpler settings \cite{NOS14,SBB15,HOX14}. 
This is made precise in \eqref{eq:intro_bound}, and in the main result of 
 Theorem \ref{thm:main2}, we appropriately 
rescale the spectral gap and use 
$\alpha\in[0,1]$ defined as  
\begin{eqnarray}
	\label{eq:lambda2_L1}
	\alpha &\equiv& \frac{\lambda_2(L)(d-1)}{\Tr(L)}  \;\;=\;\; \frac{\lambda_2(L)(d-1)}{\sum_{j = 1}^n \ell_j } \;. 
\end{eqnarray}
The accuracy also depends on the topology via the maximum weighted degree defined as 
$D_{\max} \equiv \max_{i \in [d]} D_{ii} = \max_{i \in [d]} \{ \sum_{j: i \in S_j} \ell_j/\kappa_j\}$.  
Note that the average weighted degree is $\sum_i D_{ii}/ d = \Tr(L)/d$, and we rescale it by $D_{\rm max}$ such that  
\begin{eqnarray}  
	\label{eq:lambda2_L1beta}
	\beta &\equiv& \frac{\Tr(L)}{d D_{\max}}  \;\;=\;\; \frac{\sum_{j = 1}^n \ell_j }{d D_{\max}} \;. 
\end{eqnarray}
We will show that the performance of rank breaking estimator depends on the topology of the graph 
through these two parameters. The larger the spectral gap $\alpha$ 
the smaller error we get with the same effective sample size. 
The degree imbalance $\beta\in[0,1]$ determines how many samples are required for the analysis to hold. 
We need smaller number of samples if the weighted degrees are balanced, which happens if $\beta$ is large (close to one). 

The following quantity also determines the convexity of the objective function. 
	\begin{align}
	\label{eq:gamma_def}
	 \gamma \;\equiv\; \min_{j \in [n]} \Bigg\{ \Bigg(1 - \frac{p_{j,\ell_j}}{\kappa_j} \Bigg)^{\ceil{2e^{2b}}-2}\Bigg\}  \;.\; 
	\end{align}
Note that $\gamma$ is between zero and one, and a larger value is desired 
as the objective function becomes more concave and a better accuracy follows. 
When we are collecting data where the size of the offerings $\kappa_j$'s are increasing with $d$ 
but the position of the separators are close to the top, such that $\kappa_j = \omega(d)$ and $p_{j,\ell_j} = O(1)$, 
then for $b=O(1)$ the above quantity $\gamma$ can be made arbitrarily close to one, for large enough problem size $d$. 
On the other hand, when $p_{j,\ell_j}$ is close to $\kappa_j$, the accuracy can degrade significantly as 
stronger alternatives might have small chance of showing up in the rank breaking. 
The value of $\gamma$ is quite sensitive to $b$. 
The reason we have such a inferior dependence on $b$ is because we wanted to give a universal bound 
on the Hessian that is simple. It is not difficult to get a tighter bound with a larger value of $\gamma$, but will inevitably 
depend on the structure of the data in a  complicated fashion. 

To ensure that the (second) largest eigenvalue of the Hessian is small enough, 
 we need enough samples. 
This is captured by $\eta$ defined as 
\begin{align}\label{eq:eta_def}
\eta \;\;\equiv\;\; \max_{j \in [n]} \{\eta_j\} \;,\; \;\;\;\;\;\text{where} \;\; \;\;\;\;\; \eta_j \;\; = \;\;   \frac{\kappa_j}{\max\{\ell_j, \kappa_j - p_{j,\ell_j}\}}\,.
\end{align}
Note that $1 < \eta_j \leq \kappa_j/\ell_j$. 
A smaller value of $\eta$ is desired as we require smaller number of samples, as shown in Theorem \ref{thm:main2}.  
This happens, for instance, when all separators are at the top, such that $p_{j,\ell_j}=\ell_j$ and $\eta_j=\kappa_j/(\kappa_j-\ell_j)$, which is close to one for large $\kappa_j$.  
On the other hand, when all separators are at the bottom of the list, then $\eta$ can be as large as $\kappa_j$. 
 
 We discuss the role of the topology of data captures by these parameters in Section \ref{sec:role}.


%
%

\section{Main Results}
\label{sec:main}
We present the main theoretical results accompanied by corresponding numerical simulations in this section. 

\subsection{Upper Bound on the Achievable Error}

We present the main result that provides an upper bound on the resulting error and explicitly shows the dependence on the topology of the data. 
As explained in Section \ref{sec:intro}, 
we assume that each user provides a partial ranking 
according to his/her position of the separators. Precisely, 
we assume the set of offerings $S_j$, 
the number of separators $\ell_j$, and their respective positions $\cP_j=\{p_{j,1},\ldots,p_{j,\ell_j}\}$ 
are predetermined. 
Each user draws the ranking of items from the PL model, and 
provides the partial ranking according to the separators of the form of 
$\{a>\{b,c,d\}>e>f\}$ in the example in the Figure \ref{fig:hasse}.
\begin{theorem} 
\label{thm:main2} 
Suppose  there are $n$ users, $d$ items parametrized by 
$\theta^*\in\Omega_b$, each user $j$ is presented with a set of offerings $S_j\subseteq [d]$, 
and provides a partial ordering under the PL model. 
	When the effective sample size $\sum_{j=1}^n \ell_j$ is large enough such that 
	\begin{align}
		\label{eq:main21}
		\sum_{j=1}^n \, \ell_j \;\;\geq\;\; \frac{2^{11}e^{18b}\eta\log(\ell_{\max}+2)^2 }{\alpha^2\gamma^2\beta}  d\log d\;, 
	\end{align}
	where  
	$b\equiv \max_{i}|\theta^*_i |$ is the dynamic range, $\ell_{\max} \equiv \max_{j\in[n]} \ell_j$, 
	$\alpha$ is the (rescaled) spectral gap defined in \eqref{eq:lambda2_L1}, 
	$\beta$ is the (rescaled) maximum degree defined in \eqref{eq:lambda2_L1beta}, 
	$\gamma$ and $\eta$ are defined in Eqs. \eqref{eq:gamma_def} and  \eqref{eq:eta_def},  
	then the {\em rank-breaking estimator} in \eqref{eq:theta_ml} 	
	with the choice of 
	\begin{eqnarray} 
		\lambda_{j,a} &=&  \frac{1}{\kappa_j - p_{j,a}} \;, 
		\label{eq:deflambda}
	\end{eqnarray} 
	for all $a\in[\ell_j]$ and $j\in[n]$ achieves  
	\begin{align} 
	\label{eq:main22} 
	\frac{1}{\sqrt{d}}\big\|\widehat{\theta} - \theta^* \big\|_2 \;\; \leq  \;\; \frac{4\sqrt{2}e^{4b}(1+ e^{2b})^2}{\alpha\gamma} \sqrt{\frac{d\, \log d}{\sum_{j=1}^n \ell_j}} \;,
	\end{align}
	with probability at least $ 1- 3e^3d^{-3}$.
\end{theorem}

Consider an ideal case where  
the spectral gap is large such that $\alpha$ is a strictly positive constant and 
the dynamic range $b$ is finite and $\max_{j \in[n]}p_{j,\ell_j}/\kappa_j = C$ for some constant $C <1$ 
such that $\gamma$ is also a constant independent of the problem size  $d$. 
Then the upper bound in \eqref{eq:main22} implies that we need the effective sample size to scale as 
$O(d\log d)$, which is only a logarithmic  factor larger than the number of parameters to be estimated. 
Such a logarithmic gap is also unavoidable and due to the fact that we require high probability bounds, where we want the tail probability to decrease at least polynomially in $d$. 
We discuss the role of the topology of the data in Section \ref{sec:role}. 

The upper bound follows from an analysis of the convex program similar to those in \cite{NOS12,HOX14,SBB15}. 
However, unlike the traditional data collection scenarios, 
the main technical challenge is in analyzing the probability that a particular pair of items appear in the  rank-breaking. 
We provide a proof in Section \ref{sec:proof_main2}.

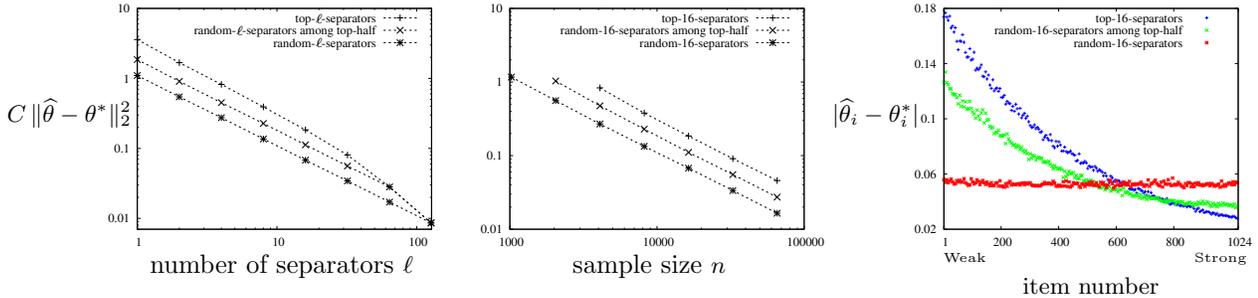
\begin{figure}[h]
 \begin{center}
	\includegraphics[width=.3\textwidth]{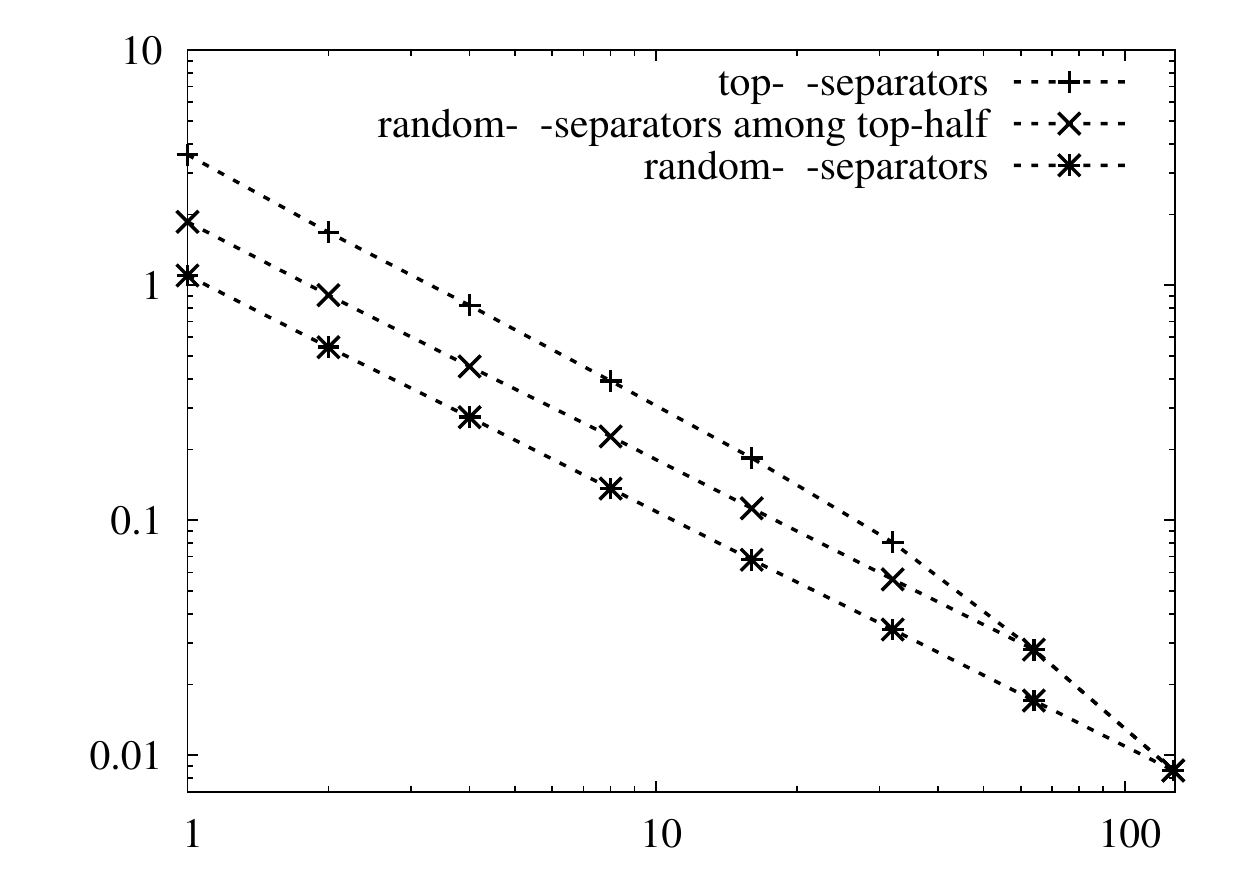} 
	\put(-169,50){\small{$C\,\|\widehat\theta-\theta^*\|_2^2$}}	
	\put(-115,-7){number of separators \small{$\ell$}} 
	\put(-52.5,87.5){\tiny{$\ell$}}	
	\put(-82.5,83){\tiny{$\ell$}}	
	\put(-52.5,78.5){\tiny{$\ell$}}	
	\includegraphics[width=.3\textwidth]{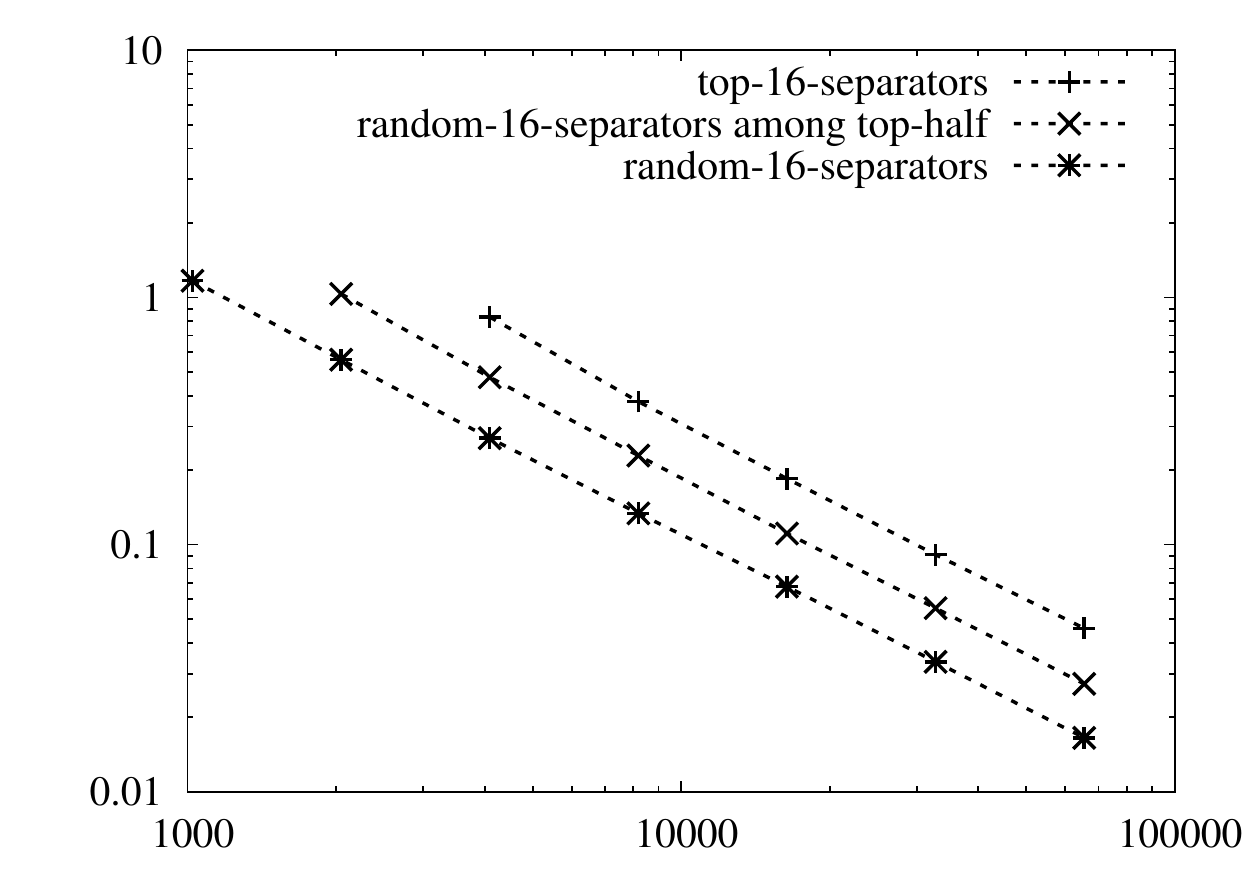}
	\put(-96,-7){sample size \small{$n$}}
	\hspace{.7cm}
	\includegraphics[width=.3\textwidth]{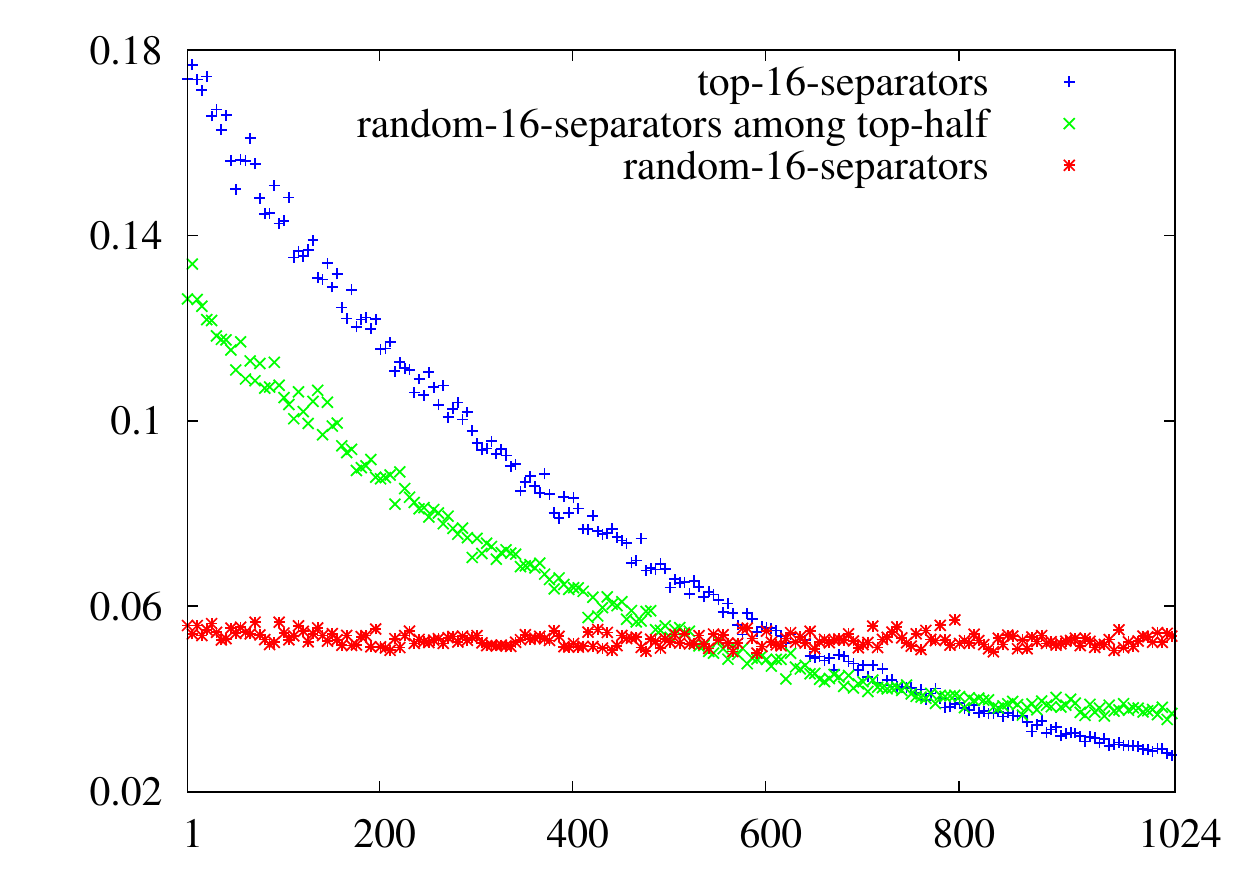}
	\put(-162,50){\small{$|\widehat\theta_i - \theta_i^* |$}}
	\put(-90,-15){\small{item number}}
	\put(-120,-3){\tiny{Weak}}
	\put(-25,-3){\tiny{Strong}}
\end{center}
\caption{Simulation confirms $\|{\theta^* - \widehat{\theta}}\|_2^2 \propto 1/(\ell\,n)$, and 
	smaller error is achieved for separators that are well spread out. }
\label{fig:scaling_l_n}
\end{figure}

In Figure \ref{fig:scaling_l_n} , we verify the scaling of the resulting error via numerical simulations.  
We fix $d=1024$ and $\kappa_j=\kappa = 128$, and vary the number of separators $\ell_j=\ell$ for fixed $n = 128000$ (left), 
and vary the number of samples $n$ for fixed $\ell_j=\ell = 16$ (middle). 
Each point is average over $100$ instances. 
The plot confirms that the mean squared error scales as $1/(\ell \, n)$. 
Each sample is a partial ranking from a set of $\kappa$ alternatives chosen uniformly at random,  
where the partial ranking is from a PL model with weights $\theta^*$ chosen i.i.d. uniformly over $[-b,b]$ with $b=2$.
To investigate the role of the position of the separators, we compare three scenarios.
The {\em top-$\ell$-separators} choose the top $\ell$ positions for separators, 
the {\em  random-$\ell$-separators among top-half} choose $\ell$ positions uniformly random from the top half, 
and the {\em random-$\ell$-separators} choose the positions uniformly at random. 
We observe that when the positions of the separators are well spread out among the $\kappa$ offerings, which happens for 
{\em random-$\ell$-separators}, we get better accuracy.

The figure on the right provides an insight into this  trend for $\ell = 16$ and $n = 16000$. The absolute error $|\theta^*_i - \widehat{\theta_i}|$ is roughly same for each item $i \in [d]$ when breaking positions are chosen uniformly at random between $1$ to $\kappa-1$ whereas it is significantly higher for weak preference score items when breaking positions are restricted between $1$ to $\kappa/2$ or are top-$\ell$. This is due to the fact that the probability of each item being ranked at different positions is different, and in particular probability of the low preference score items being ranked in top-$\ell$ is very small. The third figure is averaged over $1000$ instances. Normalization constant $C$ is $n/d^2$ and $10^{3}\ell/d^2$ for the first and second figures respectively.  For the first figure $n$ is chosen relatively large such that $n\ell$ is large enough even for $\ell = 1$.    

\subsection{The Price of Rank Breaking for the Special Case of Position-$p$ Ranking}

Rank-breaking achieves computational efficiency  at the cost of  estimation accuracy.  
In this section, we quantify this tradeoff for a canonical example of  position-$p$ ranking, where each sample 
provides the following information: an unordered set of $p-1$ items that are ranked high, 
one item that is ranked at the $p$-th position, and the rest of $\kappa_j-p$ items that are ranked on the bottom. 
An example of a sample with position-4 ranking six items $\{a,b,c,d,e,f\}$ might be a partial ranking of $(\{a,b,d\}>\{e\}>\{c,f\})$.  
Since each sample has only one separator for $2<p$, Theorem \ref{thm:main2} simplifies to the following Corollary. 

\begin{corollary}
	Under the hypotheses of Theorem \ref{thm:main2}, there exist positive constants $C$ and $c$ that only depend on $b$ such that if  $n \geq C (\eta d \log d) /(\alpha^2\gamma^2\beta)$ then 
	\begin{align} \label{eq:main3} 
	\frac{1}{\sqrt{d}}\big\|\widehat{\theta} - \theta^* \big\|_2 \;\; \leq  \;\; \frac{c}{\alpha\gamma} \sqrt{\frac{d\, \log d}{n }} \;. 
	\end{align}
	\label{coro:main2}
\end{corollary}
Note that the error only depends on the position $p$ through $\gamma$ and $\eta$, and is not sensitive. 
To quantify the price of rank-breaking, we compare this result to a fundamental lower bound on the minimax rate in Theorem \ref{thm:cramer_rao_position_p}. 
We can compute a sharp lower bound on the minimax rate, using the Cram\'er-Rao bound, and a proof is provided in Section 
\ref{sec:proof_cramer_rao_position_p}.
\begin{theorem} 
	\label{thm:cramer_rao_position_p}
	Let $\mathcal{U}$ denote the set of all unbiased estimators of $\theta^*$ 
	and suppose $b >0$, then
	\begin{eqnarray*}
	\inf_{\widehat{\theta} \in \mathcal{U}} \sup_{\theta^* \in \Omega_b} \E[\norm{\widehat{\theta} - \theta^*}^2] &\geq& 
	 \frac{1}{2p\log(\kappa_{\max})^2} \sum_{i = 2}^d \frac{1}{\lambda_i(L)} 
	 \;\, \geq \;\, \frac{1}{2p\log(\kappa_{\max})^2} \frac{(d-1)^2}{n }\;,
	\end{eqnarray*}
	where $\kappa_{\rm max} = \max_{j\in[n]} |S_j| $ and the second inequality follows from the Jensen's inequality.
\end{theorem}

Note that the second inequality is tight up to a constant factor, 
when the graph is an expander with a large spectral gap. 
For expanders, $\alpha$ in the bound \eqref{eq:main3} is also a strictly positive constant. 
This suggests that rank-breaking gains in computational efficiency by a super-exponential factor of $(p-1)!$, at the price of 
increased error by a factor of $p$, ignoring poly-logarithmic factors.

\subsection{Tighter Analysis for the  Special Case of Top-$\ell$ Separators Scenario}
\label{sec:topl}

The main result in Theorem \ref{thm:main2} is general in the sense that it applies to any partial ranking data 
that is represented by positions of the separators. 
However, the bound can be quite loose, especially when $\gamma$ is small, i.e. 
$p_{j,\ell_j}$ is close to $\kappa_j$. 
For some special cases,  we can tighten the analysis to get a sharper bound. 
One caveat is that we use a slightly sub-optimal choice of parameters $\lambda_{j,a} = 1/\kappa_j$ instead of $1/(\kappa_j - a)$, to simplify the analysis and still get the order optimal error bound we want.   
Concretely, we consider a special case of top-$\ell$ separators scenario, where each agent gives 
a ranked list of her most preferred $\ell_j$ alternatives among $\kappa_j$ offered set of items. 
Precisely, the locations of the separators are $(p_{j,1},p_{j,2},\ldots,p_{j,\ell_j})=(1,2,\ldots,\ell_j)$. 


\begin{theorem} 
\label{thm:topl_upperbound}
Under the PL model, $n$ partial orderings are sampled over $d$ items parametrized by $\theta^* \in \Omega_b$, 
where the $j$-th sample is a ranked list of the top-$\ell_j$ items among the $\kappa_j$ items offered to the agent. 
If 
\begin{align} \label{eq:topl1}
\sum_{j = 1}^n \ell_j \;\; \geq \;\; \frac{2^{12}e^{6b}}{\beta\alpha^2} d\log d\,,
\end{align}
where $b \equiv \max_{i,\i} |\theta^*_i - \theta^*_{\i}|$ and $\alpha,\beta$ are defined in \eqref{eq:lambda2_L1} and \eqref{eq:lambda2_L1beta}, then the {\em rank-breaking estimator} in \eqref{eq:theta_ml} 	
	with the choice of $\lambda_{j,a} =  1/{\kappa_j}$
	for all $a\in[\ell_j]$ and $j\in[n]$ achieves  
	\begin{align} 
	\label{eq:main_topl} 
	\frac{1}{\sqrt{d}}\big\|\widehat{\theta} - \theta^* \big\|_2 \;\; \leq  \;\; \frac{16(1+ e^{2b})^2}{\alpha} \sqrt{\frac{d\, \log d}{\sum_{j=1}^n \ell_j}} \;,
	\end{align}
	with probability at least $ 1- 3e^3 d^{-3}$.
\end{theorem}

A proof is provided in Section \ref{sec:proof_topl_upperbound}. 
In comparison to the general bound in Theorem \ref{thm:main2}, 
this is tighter since there is no dependence in $\gamma$ or $\eta$. 
This gain is significant when, for example, $p_{j,\ell_j}$ is close to $\kappa_j$. 
As an extreme example, if all agents are offered the entire set of alternatives and 
are asked to rank all of them, such that $\kappa_j=d$ and $\ell_j=d-1$ for all $j\in[n]$, 
then the generic bound in \eqref{eq:main22} is loose by a factor of $(e^{4b}/2\sqrt{2}) d^{\lceil2e^{2b}\rceil-2}$, compared to the above bound. 

\begin{figure}[h]
 \begin{center}
	\includegraphics[width=.3\textwidth]{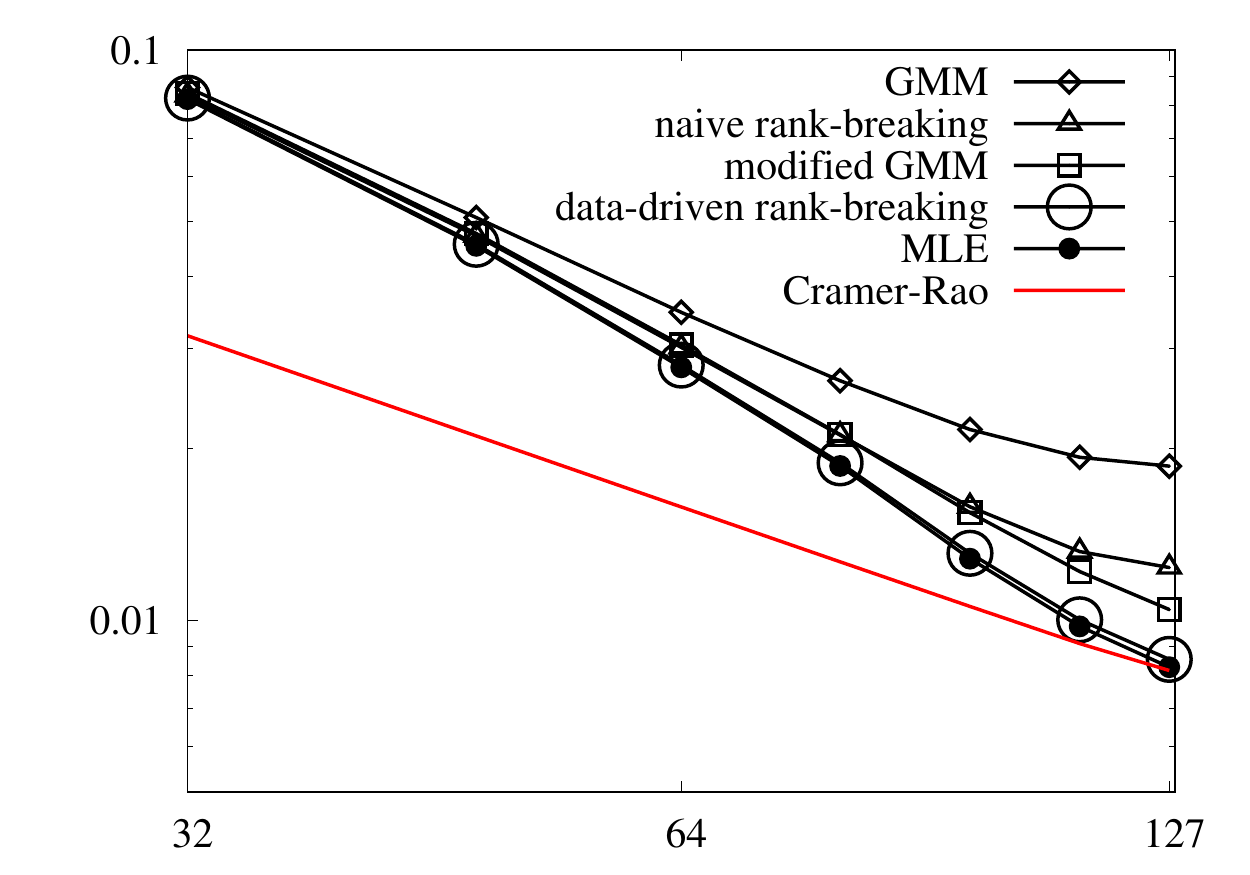}
		\put(-171,50){\small{$C\,\|\widehat\theta-\theta^*\|_2^2$}}	
	\put(-115,-7 ){number of separators \small{$\ell$}} 
	\put(-100,100){\small{Top-$\ell$ separators}}
\end{center}
\caption{The proposed data-driven rank-breaking achieves performance identical to the MLE, and improves over naive rank-breaking with uniform weights. }
\label{fig:top_l}
\end{figure}

In the top-$\ell$ separators scenario, 
the data set consists of the ranking among top-$\ell_j$ items of the set $S_j$, i.e., ${[\sigma_j(1), \sigma_j(2),\cdots, \sigma_j(\ell_j)]}$. 
The corresponding log-likelihood of the PL model is   
	\begin{align}\label{eq:PL_likelihood}
	\L(\theta) = \sum_{j = 1}^n \sum_{m = 1}^{\ell_j} \Big[ \theta_{\sigma_j(m)} - \log \Big( \exp(\theta_{\sigma_j(m)})+\exp(\theta_{\sigma_j(m+1)})+ \cdots + \exp(\theta_{\sigma_j(\kappa_j)})\Big) \Big]\;,
	\end{align}
where $\sigma_j(a)$ is the alternative ranked at the $a$-th position by agent $j$. 
The Maximum Likelihood Estimator (MLE) for this {\em traditional} data set is efficient. 
Hence, there is no computational gain in rank-breaking. 
Consequently, there is no loss in accuracy either, when we use the optimal weights proposed in the above theorem. Figure \ref{fig:top_l} illustrates that the MLE and the data-driven rank-breaking estimator achieve performance that is identical, 
and improve over naive rank-breaking that uses uniform weights. 
We also compare performance of Generalized Method-of-Moments (GMM) proposed by \cite{ACPX13} with our algorithm. In addition, we show that performance of GMM can be improved by optimally weighing pairwise comparisons with $\lambda_{j,a}$. MSE of GMM in both the cases, uniform weights and optimal weights, is larger than our rank-breaking estimator. However, GMM is on average about four times faster than our algorithm. 
We choose $\lambda_{j,a} = 1/(\kappa_j-a)$ in the simulations, as opposed to the $1/\kappa_j$ assumed in the above theorem. 
This settles the question raised in \cite{HOX14} on whether it is possible to achieve optimal accuracy using rank-breaking 
under the top-$\ell$ separators scenario. 
Analytically, it was proved in \cite{HOX14} that under the top-$\ell$ separators scenario, 
naive rank-breaking with uniform weights achieves the same error bound as the MLE, up to a constant factor. 
However, we show that this constant factor gap is not a weakness of the analyses, but the choice of the weights.  
Theorem \ref{thm:topl_upperbound} provides a guideline for choosing the optimal weights, and  
the numerical simulation results in Figure \ref{fig:top_l} show that there is in fact no gap in practice, if we use the optimal weights. 
We use the same settings as that of the first figure of Figure \ref{fig:scaling_l_n} for the figure below.

To prove the order-optimality of the rank-breaking approach up to a constant factor, we can compare the upper bound to a 
 Cram\'er-Rao lower bound on any unbiased estimators, in the following theorem. A proof is provided in Section \ref{sec:proof_cramer_rao_topl}.
\begin{theorem} \label{thm:cramer_rao_topl}
	Consider ranking $\{\sigma_j(i)\}_{i \in [\ell_j]}$ revealed for the set of items $S_j$, for $j \in [n]$. 
	Let $\mathcal{U}$ denote the set of all unbiased estimators of $\theta^*\in\Omega_b$. 
	If $b >0$, then
	\begin{eqnarray}
	\inf_{\widehat{\theta} \in \mathcal{U}} \sup_{\theta^* \in \Omega_b} \E[\norm{\widehat{\theta} - \theta^*}^2] 
	\;\; \geq \;\; \Bigg(1 - \frac{1}{\ell_{\max}}\sum_{i= 1}^{\ell_{\max}} \frac{1}{\kappa_{\max} - i +1}\Bigg)^{-1} \sum_{i = 2}^d \frac{1}{\lambda_i(L)} 
	\;\; \geq \;\;  \frac{(d-1)^2}{\sum_{j = 1}^n \ell_j}\;,
	\label{eq:cramer_rao_topl}
	\end{eqnarray}
	where $\ell_{\max} = \max_{j \in [n]} \ell_j$  and $\kappa_{\max} = \max_{j \in [n]} \kappa_j$. 
	The second inequality follows from the Jensen's inequality.
	\end{theorem}
Consider a case when the comparison graph is an expander such that $\alpha$ is a strictly positive constant, 
and $b=O(1)$ is also finite. 
Then, the Cram\'er-Rao lower bound show that the 
 upper bound in \eqref{eq:main_topl} is optimal up to a logarithmic factor.

\subsection{Optimality of the Choice of the Weights}

We propose the optimal choice of the weights $\lambda_{j,a}$'s in Theorem \ref{thm:main2}. 
In this section, we show numerical simulations results  comparing the proposed approach to 
other naive choices of the weights under various scenarios. 
We fix $d = 1024$ items and the underlying preference vector $\theta^*$ is uniformly distributed over $[-b,b]$ for $b = 2$. 
We generate $n$ rankings over sets $S_j$ of size $\kappa$ for $j \in [n]$ according to the PL model with parameter $\theta^*$. The comparison sets $S_j$'s are chosen independently and uniformly at random from $[d]$. 

\begin{figure}[h]
 \begin{center}
	\includegraphics[width=.3\textwidth]{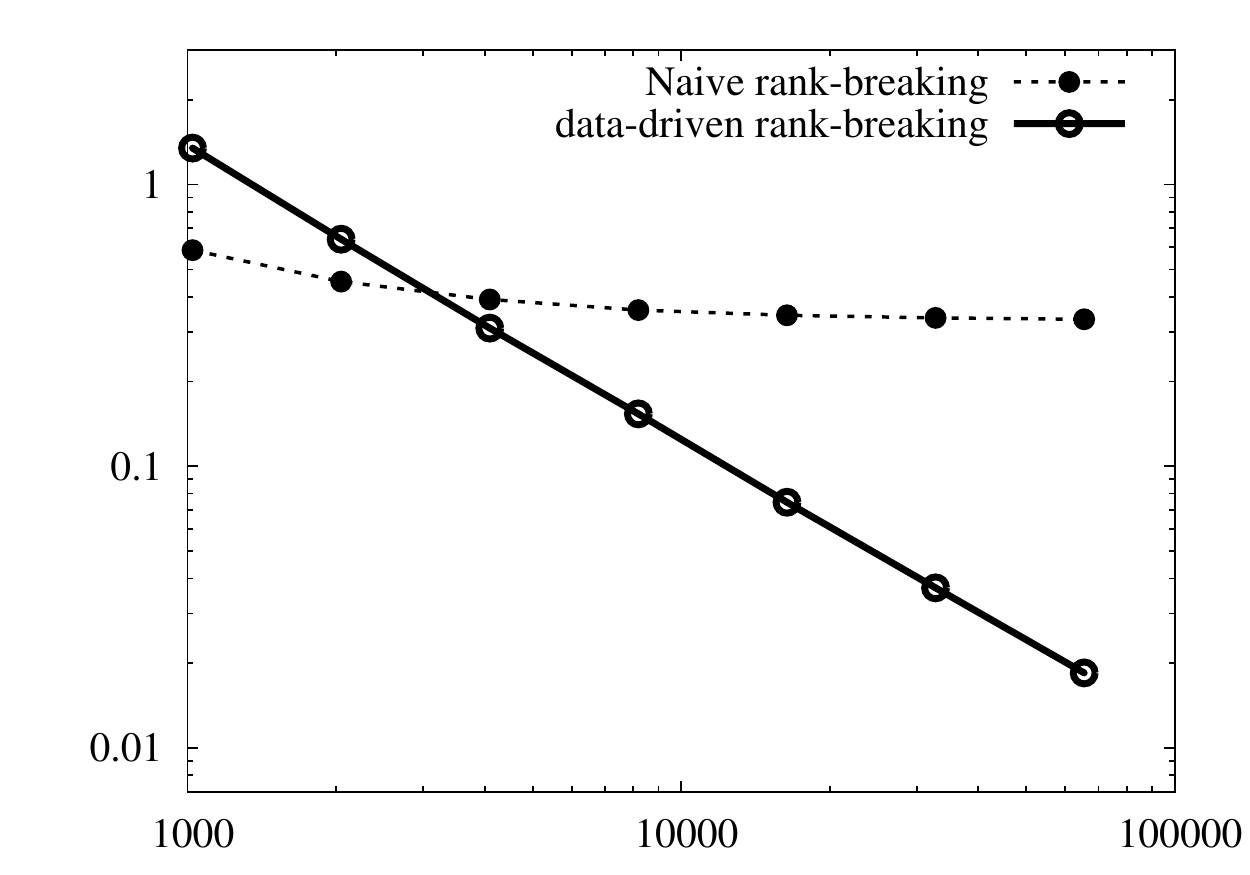}
	\put(-92,-5){\small{sample size $n$}}
	\put(-180,50){\small{$C\,\|\widehat\theta-\theta^*\|_2^2$}}
\end{center}
\caption{Data-driven rank-breaking is consistent, while a random rank-breaking results in inconsistency.}
\label{fig:consistent}
\end{figure}

Figure \ref{fig:consistent} illustrates that a naive choice of rank-breakings can result in inconsistency. 
We create partial orderings data set by 
fixing $\kappa = 128$ and select $\ell=8$ random positions in $\{1,\ldots,127\}$. 
Each data set consists of partial orderings with separators at those $8$ random positions, over $128$ randomly chosen subset of items.  
We vary the sample size $n$ and plot the resulting mean squared error for the two approaches. 
The data-driven rank-breaking, which uses the optimal choice of the weights, achieves error scaling as $1/n$ as predicted by Theorem \ref{thm:main2}, which implies consistency. 
For fair comparisons, we feed the same number of pairwise orderings to a naive rank-breaking estimator. 
This estimator uses randomly chosen pairwise orderings with uniform weights, and is generally inconsistent. 
However, when sample size is small, inconsistent estimators can 
achieve smaller variance leading to smaller error. 
 Normalization constant $C$ is $10^{3}\ell/d^2$, and 
each point is  averaged over $100$ trials. 
We use the minorization-maximization algorithm from \cite{Hun04} for computing the estimates from the rank-breakings. 


Even if we use the consistent rank-breakings first proposed in \cite{APX14a}, there is ambiguity in the choice of the weights. We next study how much we gain by using the proposed optimal choice of the weights. 
The optimal choice, $\lambda_{j,a}=1/(\kappa_j-p_{j,a})$, depends on two parameters: the size of the offerings  $\kappa_j$ and the position of the separators $p_{j,a}$. 
To distinguish the effect of these  two parameters, 
 we first experiment with fixed $\kappa_j=\kappa$ and illustrate the gain of the optimal choice of 
 $\lambda_{j,a}$'s.

\begin{figure}[h]
 \begin{center}
	\includegraphics[width=.3\textwidth]{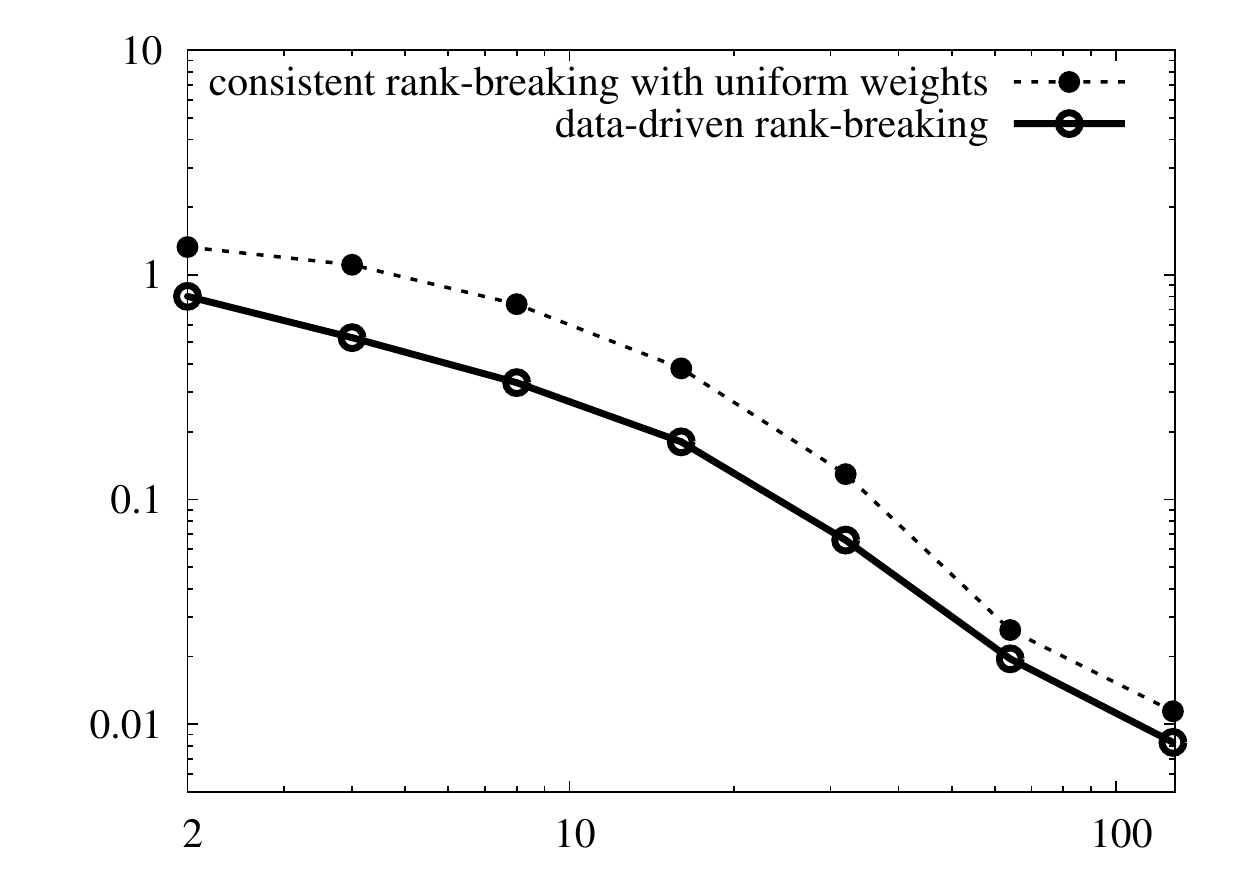}
   	\put(-85,102){Top-$1$ and bottom-$(\ell-1)$ separators}
	\put(-180,50){\small{$C\,\|\widehat\theta-\theta^*\|_2^2$}}	
	\put(-115,-5){number of separators \small{$\ell$}}
	\hspace{.05\textwidth}
	\includegraphics[width=.22\textwidth]{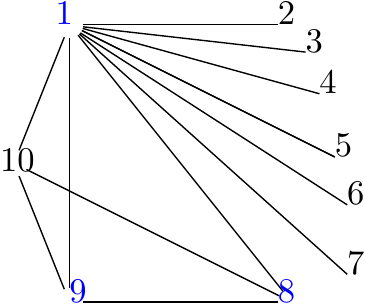}

\end{center}
\caption{There is a constant factor gain of choosing optimal $\lambda_{j,a}$'s when the size of offerings are fixed, i.e. $\kappa_j = \kappa$ (left). We choose a particular set of separators where one separators is at position one and the rest are at the bottom. An example for $\ell=3$ and $\kappa=10$ is shown, where the separators are indicated by blue (right).}
\label{fig:lambda_impact1}
\end{figure}


Figure \ref{fig:lambda_impact1} illustrates that the optimal choice of the weights improves over 
consistent rank-breaking with uniform weights  by a constant factor. 
We fix  $\kappa = 128$ and $n=128000$. 
As illustrated by a figure on the right, the position of the separators are chosen such that 
there is one separator at position one, and the rest of $\ell-1$ separators are at the bottom.
Precisely, $(p_{j,1},p_{j,2},p_{j,3},\ldots,p_{j,\ell})=(1,128-\ell+1,128-\ell+2,\ldots,127)$. 
We consider this scenario to emphasize the gain of optimal weights. 
 Observe that the MSE does not decrease at a rate of $1/\ell$ in this case. The parameter $\gamma$ which appears in the bound of Theorem \ref{thm:main2} is very small when the breaking positions $p_{j,a}$ are of the order $\kappa_j$ as is the case here, when $\ell$ is small. Normalization constant $C$ is $n/d^2$.

\begin{figure}[h]
 \begin{center}
	\includegraphics[width=.3\textwidth]{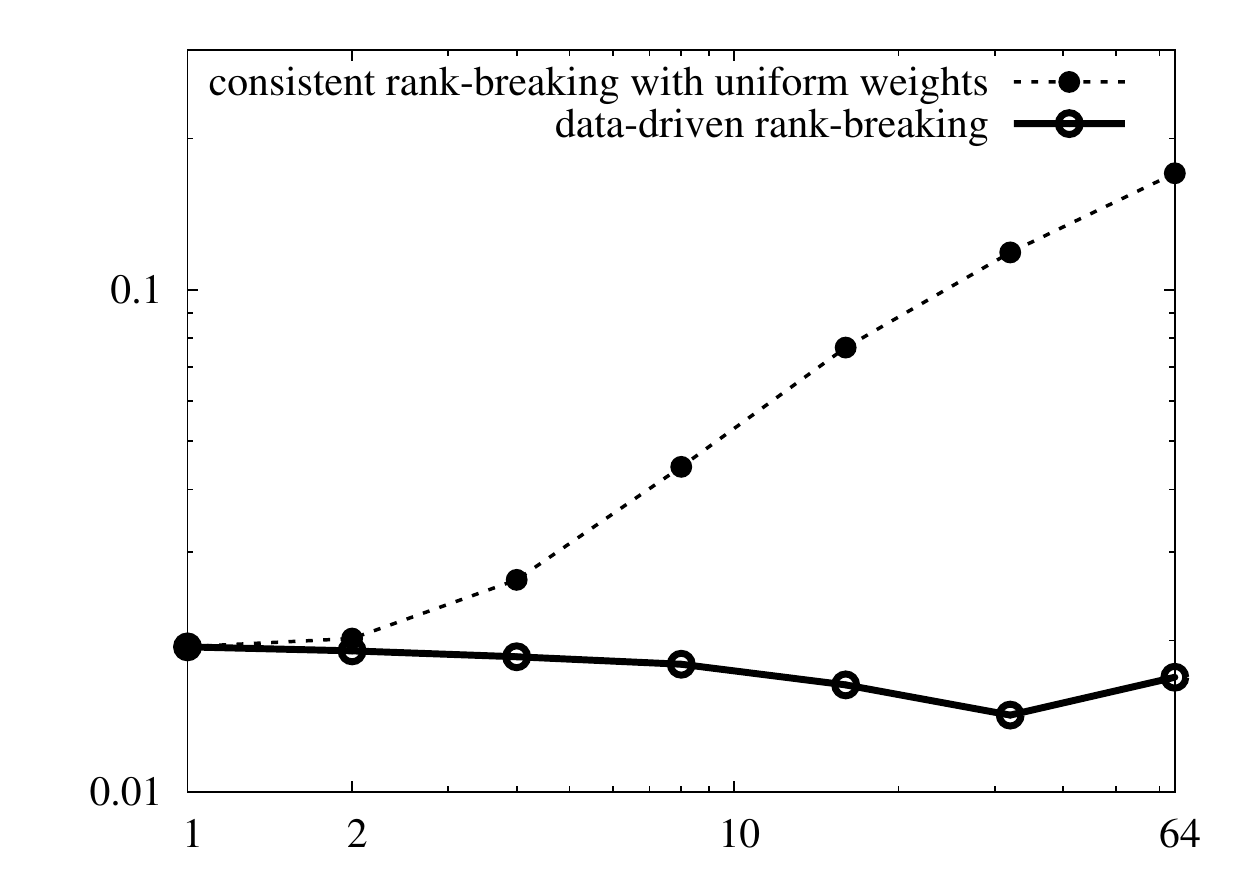}
	\put(-180,50){\small{$C\,\|\widehat\theta-\theta^*\|_2^2$}}	
	\put(-104,-7){\small{heterogeneity $\kappa_1/\kappa_2$}}
\end{center}
\caption{The gain of choosing optimal $\lambda_{j,a}$'s is significant when $\kappa_j$'s are highly heterogeneous. }
\label{fig:lambda_impact2}
\end{figure}

The gain of optimal weights is significant when the size of $S_j$'s are highly heterogeneous. 
Figure \ref{fig:lambda_impact2} compares performance of the proposed algorithm, for the optimal choice and uniform choice of weights $\lambda_{j,a}$ when the comparison sets $S_j$'s are of different sizes. We consider the case when $n_1$ agents provide their top-$\ell_1$ choices over the sets of size $\kappa_1$, and $n_2$ agents provide their top-$1$ choice over the sets of size $\kappa_2$. We take $n_1 = 1024$, $\ell_1 = 8$, and $n_2 = 10n_1\ell_1$. Figure \ref{fig:lambda_impact2} shows MSE for the two choice of weights, when we fix $\kappa_1 = 128$, and vary $\kappa_2$ from $2$ to $128$. As predicted from our bounds, when optimal choice of $\lambda_{j,a}$ is used MSE is not sensitive to sample set sizes $\kappa_2$. The error decays at the rate proportional to the inverse of the effective sample size, which is $n_1\ell_1 + n_2\ell_2 = 11n_1\ell_1$. However, with $\lambda_{j,a} = 1$ when $\kappa_2 = 2$, the MSE is roughly $10$ times worse. Which  reflects that the effective sample size is approximately $n_1\ell_1$, i.e. pairwise comparisons coming from small set size do not contribute without proper normalization. This gap in MSE corroborates bounds of Theorem \ref{thm:main}. Normalization constant $C$ is $10^{3}/d^2$.


\section{The Role of the Topology of the Data}
\label{sec:role} 

We study the role of  topology of the data that provides a guideline for 
designing the collection of data when we do have some control, 
as in recommendation systems, designing surveys, and crowdsourcing. 
The core optimization problem of interest to the designer of such a system is 
to achieve the best accuracy while minimizing the number of questions.

\subsection{The Role of the Graph Laplacian}

Using the same number of samples, comparison graphs with larger spectral gap achieve better accuracy, 
compared to those with smaller spectral gaps. To illustrate how graph topology effects the accuracy, 
we reproduce known spectral properties of canonical graphs, and numerically compare the performance of data-driven rank-breaking for several graph topologies. 
We follow the examples and experimental setup from \cite{SBB15} for a similar result with pairwise comparisons. 
Spectral properties of graphs have been a topic of wide interest for decades. 
We consider a scenario where we fix the size of offerings as $\kappa_j=\kappa=O(1)$ and 
each agent provides partial ranking with  
$\ell$ separators, positions of which are chosen uniformly at random.  
The resulting spectral gap $\alpha$ of different choices of the set $S_j$'s are provided below. 
	The total number edges in the comparisons graph (counting hyper-edges as multiple edges) is defined as  $|E|\equiv{\kappa \choose 2}\,n$.  

\begin{itemize}
	\item Complete graph: when $|E|$ is larger than ${d \choose 2}$, we can design the comparison graph to be a complete graph over $d$ nodes. The weight $A_{ii'}$ on each edge is 
	$n\,\ell/(d(d-1))$, which is the effective number of samples divided by twice the number of edges. 
	Resulting spectral gap is one, which is the maximum possible value. 
	Hence, complete graph is optimal for rank aggregation.
	\item Sparse random graph: when we have limited resources we might not be able to afford a dense graph. When  $|E|$ is of order $o(d^2)$, we have a sparse graph. Consider a scenario where each set $S_j$ is chosen uniformly at random. To ensure connectivity, we need $n=\Omega(\log d)$. Following standard spectral analysis of random graphs, we have $\alpha=\Theta(1)$. Hence, sparse random graphs are near-optimal for rank-aggregation. 
	\item Chain graph: we consider a chain of sets of size $\kappa$ overlapping only by one item. 
	For example, $S_1=\{1,\ldots,\kappa\}$ and $S_2=\{\kappa,\kappa+1,\ldots,2\kappa-1\}$, etc. 
	We choose $n$ to be a multiple of $\tau \equiv (d-1)/(\kappa-1)$ and offer each set $n/\tau$ times.	
	The resulting graph is a chain of size $\kappa$ cliques, and standard spectral analysis shows that 
	$\alpha=\Theta(1/d^2)$. Hence, a chain graph is strictly sub-optimal for rank aggregation. 
	\item Star-like graph: We choose one item to be the center, and every offer set consists of this center node and a set of $\kappa-1$ other nodes chosen uniformly at random without replacement. For example, center node = $\{1\}$, $S_1=\{1,2,\ldots,\kappa\}$ and $S_2=\{1,\kappa+1,\kappa+2,\ldots,2\kappa-1\}$, etc. $n$ is chosen in the way similar to that of the Chain graph. Standard spectral analysis shows that $\alpha=\Theta(1)$ and star-like graphs are near-optimal for rank-aggregation. 
	\item Barbell-like graph: We select an offering $S = \{S',i,j\}$, $|S'| = \kappa-2$ uniformly at random and divide rest of the items into two groups $V_1$ and $V_2$. We offer set $S$ $n\kappa/d$ times. For each offering of set $S$, we offer $d/\kappa -1$ sets chosen uniformly at random from the two groups $\{V_1,i\}$ and $\{V_2,j\}$.
	The resulting graph is a barbell-like graph, and standard spectral analysis shows that 
	$\alpha=\Theta(1/d^2)$. Hence, a chain graph is strictly sub-optimal for rank aggregation. 
\end{itemize}

\begin{figure}[H]
 \begin{center}
	\includegraphics[width=.3\textwidth]{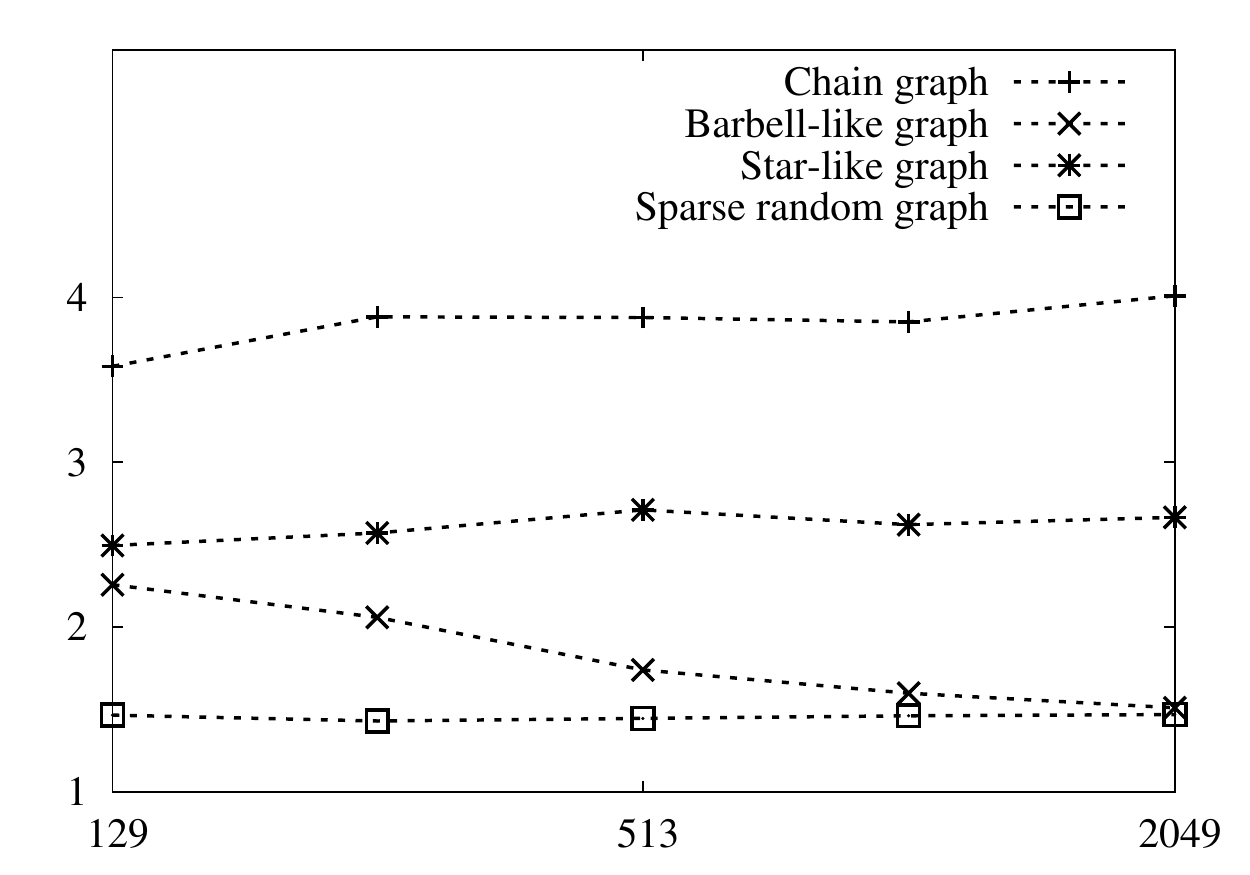} 
	\put(-180,50){\small{$C\,\|\widehat\theta-\theta^*\|_2^2$}}	
	\put(-95,-5){\small{graph size $d$}} 
	\put(-90,100){Random $\theta^*$}
	\includegraphics[width=.3\textwidth]{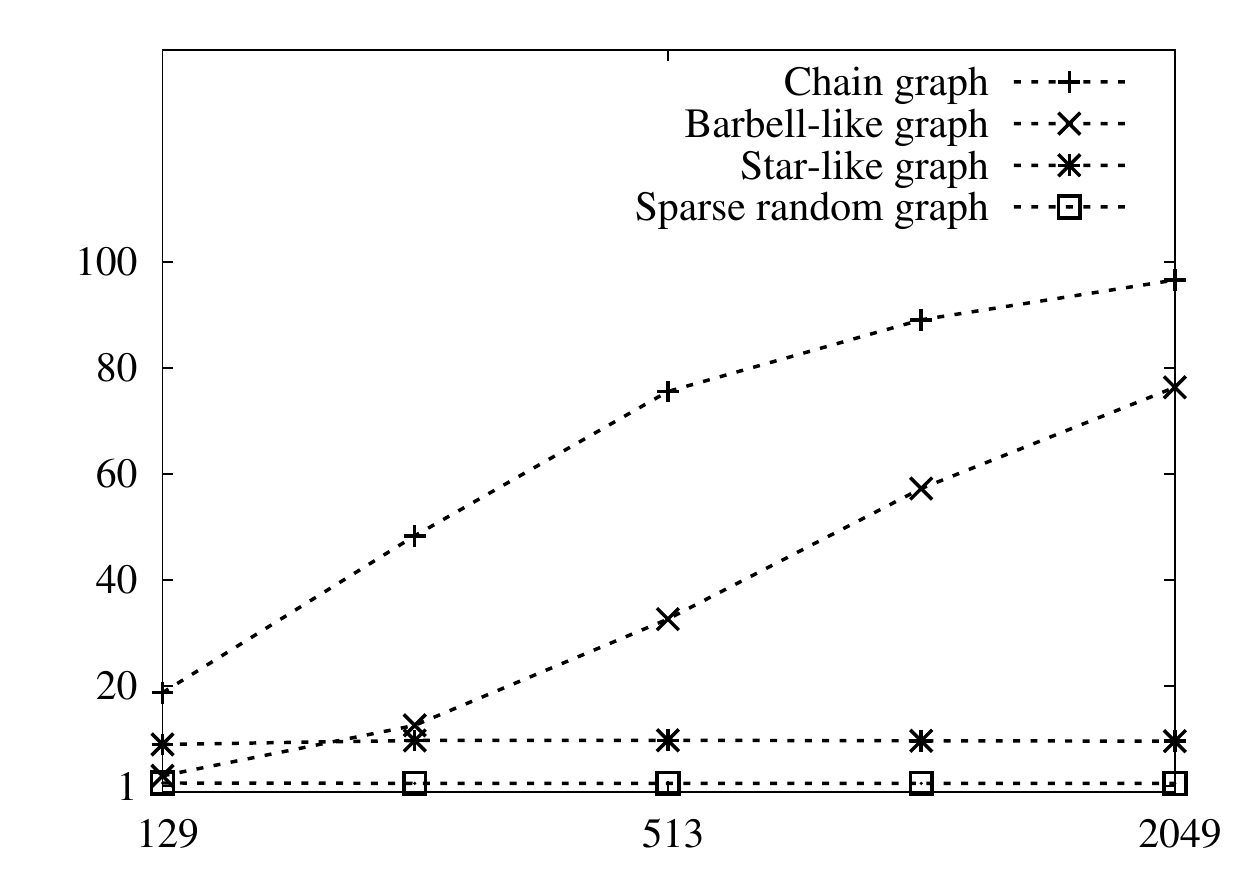} 
	\put(-95,-5){graph size  \small{$d$}} 
	\put(-95,100){Worst-case $\theta^*$}
\end{center}
\caption{For randomly chosen $\theta^*$ the error does not change with $d$ (left). However, for particular worst-case $\theta^*$ the error increases with $d$ for the Chain graph and the Barbell-like graph as predicted by the analysis of the spectral gap (right).}
\label{fig:topology}
\end{figure}
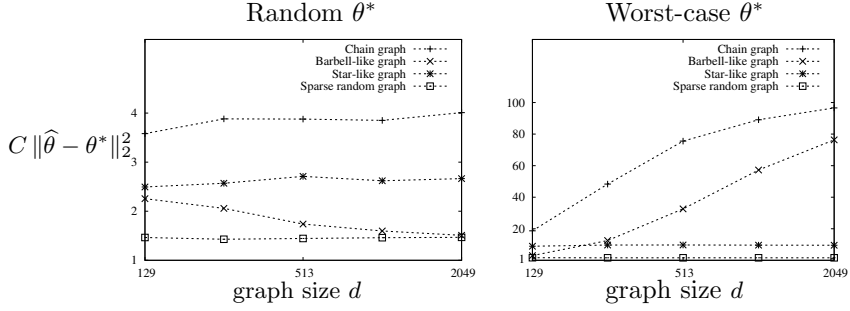

Figure \ref{fig:topology} illustrates how graph topology effects the accuracy. 
When $\theta^*$ is chosen uniformly at random, the accuracy does not change with $d$ (left), and the accuracy is better for those graphs with larger spectral gap. 
However, for a certain worst-case $\theta^*$, the error increases with $d$ for the chain graph and the barbell-like graph, as predicted by the above analysis of the spectral gap. 
We use $\ell = 4$, $\kappa = 17$ and vary $d$ from $129$ to $2049$. $\kappa$ is kept small to make the resulting graphs more like the above discussed graphs. Figure on left shows accuracy when $\theta^*$ is chosen i.i.d. uniformly over $[-b,b]$ with $b=2$. Error in this case is roughly same for each of the graph topologies with chain graph being the worst. 
	However, when $\theta^*$ is chosen carefully error for chain graph and barbell-like graph increases with $d$  as shown in the figure right. 
	We chose $\theta^*$ such that all the items of a set have same weight, either $\theta_i = 0$ or $\theta_i = b$ for chain graph and  barbell-like graph. 
	We divide all the sets equally between the two types for chain graph. For barbell-like graph, we keep the two types of sets on the two different sides of the connector set and equally divide items of the connector set into two types. 
	 Number of samples $n$ is $100(d-1)/(\kappa-1)$ and each point is averaged over $100$ instances. Normalization constant $C$ is $n\ell/d^2$. 

\subsection{The Role of the Position of the Separators}

As predicted by theorem \ref{thm:main2}, rank-breaking fails when 
$\gamma$ is small, i.e. the position of the separators are very close to the bottom. 
An extreme example is the bottom-$\ell$ separators scenario, where 
each person is offered $\kappa$ randomly chosen alternatives, 
and is asked to give a ranked list of bottom $\ell$ alternatives. 
In other words, the $\ell$ separators are placed at $(p_{j,1},\ldots,p_{j,\ell})=(\kappa_j-\ell, \ldots,\kappa-1)$. In this case, $\gamma\simeq 0$ and the error bound is large. 
This is not a weakness of the analysis. In fact we observe large errors under this scenario. 
The reason is that many alternatives that have large weights $\theta_i$'s will rarely be even compared once, making any reasonable estimation infeasible. 

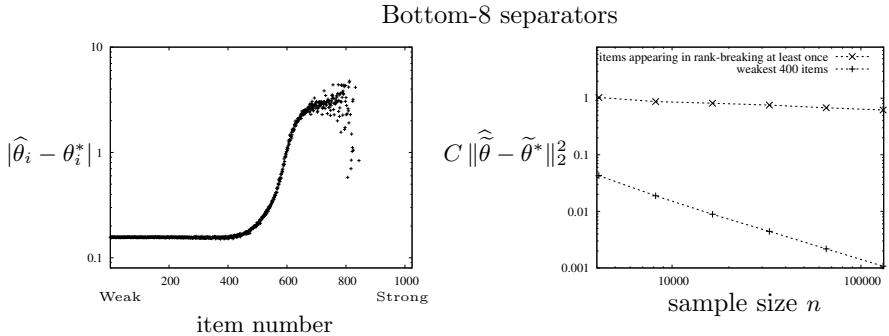
\begin{figure}[H]
 \begin{center}
	\includegraphics[width=.3\textwidth]{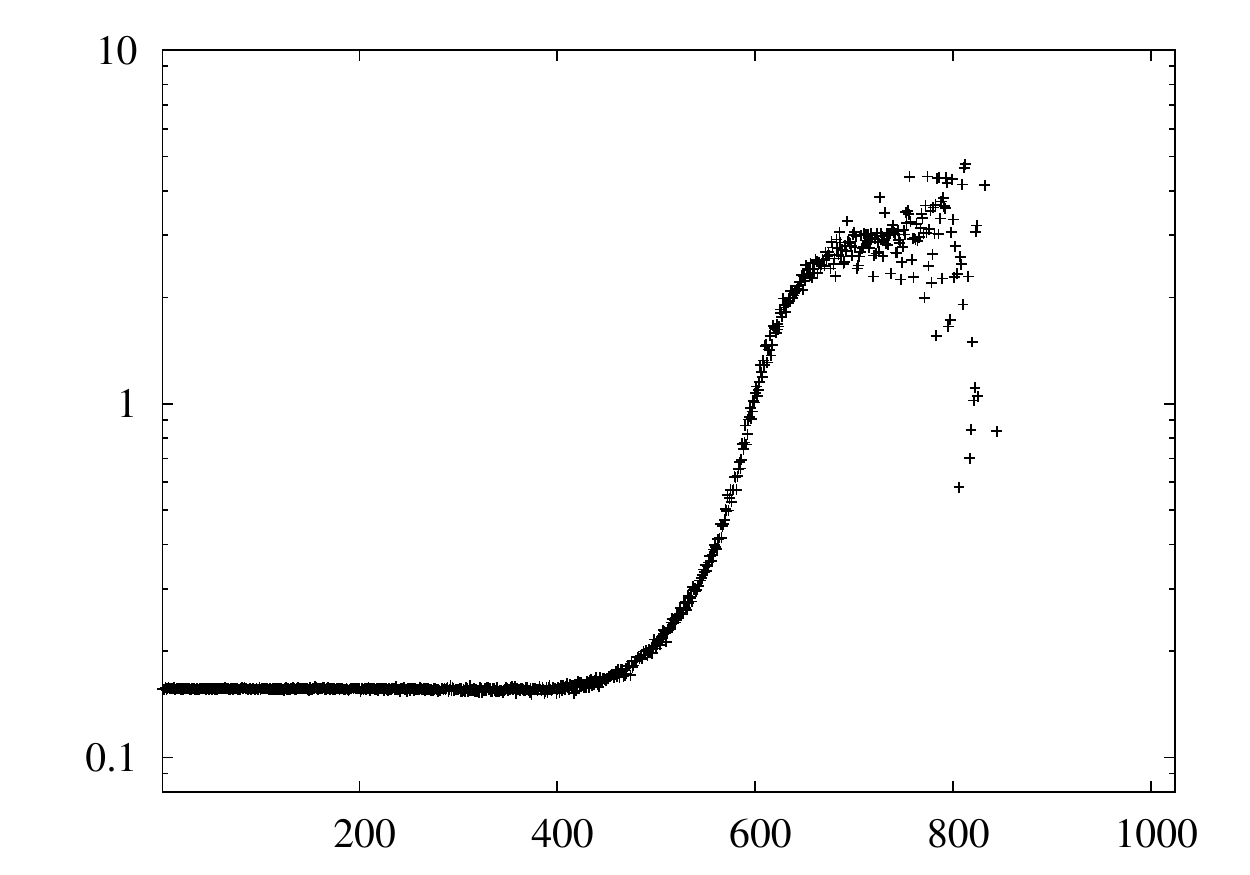}
	\put(-162,50){\small{$|\widehat\theta_i - \theta_i^* |$}}
	   	\put(-20,102){Bottom-$8$ separators}
	\put(-90,-15){\small{item number}}
	\put(-127,-3){\tiny{Weak}}
	\put(-22,-3){\tiny{Strong}}
	\hspace{1.2cm}
	\includegraphics[width=.3\textwidth]{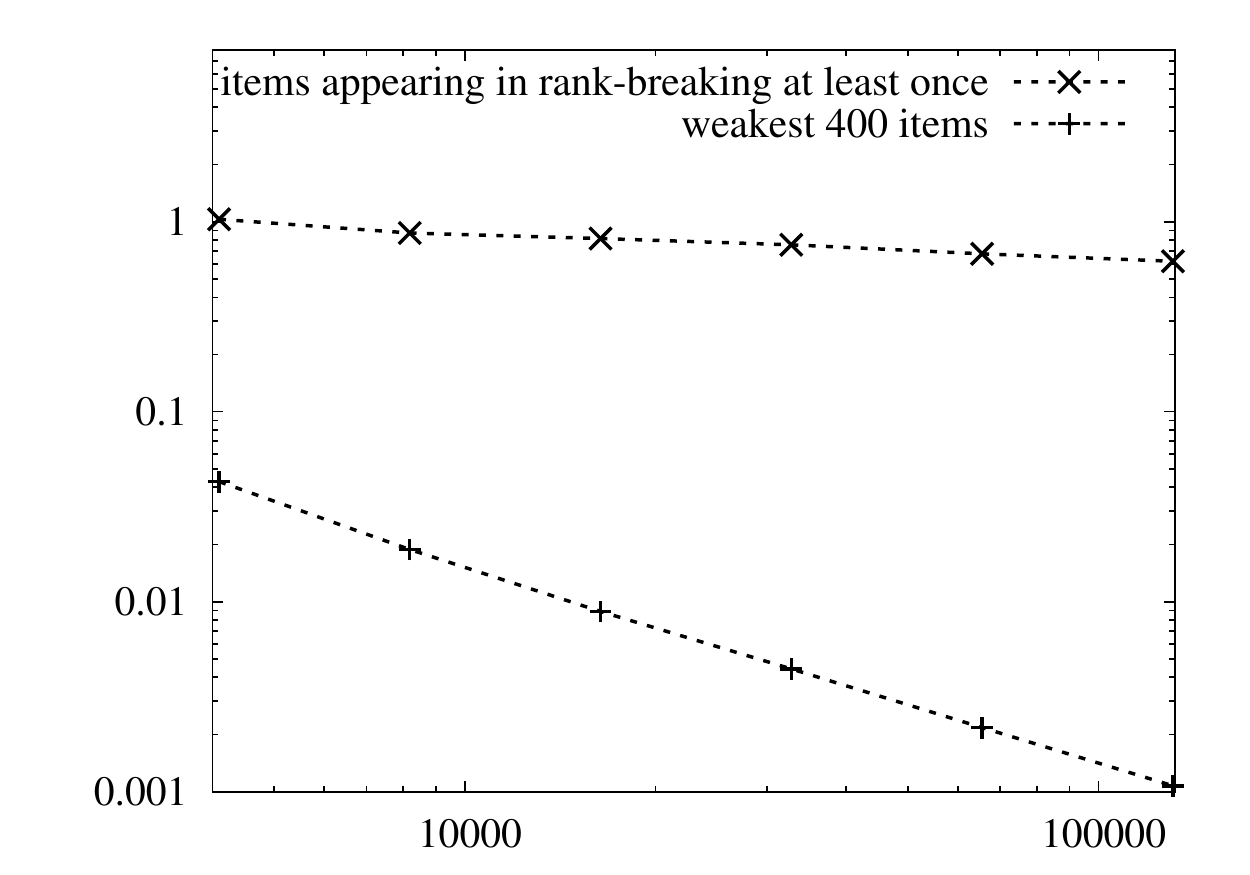}
    \put(-175,50){\small{$C\,\|\widehat\ltheta-\ltheta^*\|_2^2$}}	
	\put(-90,-7){sample size \small{$n$}}
\end{center}
\caption{Under the bottom-$\ell$ separators scenario, accuracy is good only for the bottom 400 items (left). As predicted by Theorem \ref{thm:bottoml_upperbound},  the mean squared error on the bottom 400 items scale as $1/n$, where as the overall mean squared error does not decay (right). 
}
\label{fig:bottom_l_1}
\end{figure}

Figure \ref{fig:bottom_l_1} illustrates this scenario. 
We choose $\ell=8$, $\kappa=128$, and $d=1024$. The other settings are same as that of the first figure of Figure \ref{fig:scaling_l_n}.
The left figure plots the magnitude of the estimation error for each item. 
For about 200 strong items among 1024, we do not even get a single comparison, hence we omit any estimation error.
It clearly shows the trend:  we get good estimates  for about 400 items in the bottom, 
and we get large errors for the rest. 
Consequently, even if we only take those items that have at least one comparison into account, we still get large errors. This is shown in the figure right.
The error barely decays with the sample size. 
However, if we focus on the error for the bottom 400 items, we get good error rate decaying inversely with the sample size. Normalization constant $C$ in the second figure is $10^2 \,x\,d/\ell$ and $10^{2}(400)d/\ell$ for the first and second lines respectively, where $x$ is the number of items that appeared in rank-breaking at least once. We solve convex program \eqref{eq:theta_ml} for $\theta$ restricted to the items that appear in rank-breaking at least once. The second figure of Figure \ref{fig:bottom_l_1} is averaged over $1000$ instances.

We make this observation precise in the following theorem. 
Applying rank-breaking to only 
to those weakest $\ld$ items, we prove an upper bound on the achieved error rate 
that depends on the choice of  the $\ld$. 
Without loss of generality, we suppose the items are sorted such that 
$\theta^*_1 \leq \theta_2^* \leq \cdots \leq \theta_d^*$. 
For a choice of  $\ld = \ell d/ (2 \kappa) $, 
we denote the weakest $\ld$ items by $\ltheta^* \in \reals^{\ld}$ such that $\ltheta_i^* = \theta^*_i - (1/\ld)\sum_{\i = 1}^{\ld} \theta^*_{\i}$, for $i \in [\ld]$. Since $\theta^* \in \Omega_b$, $\ltheta^* \in [-2b,2b]^{\ld}$. The space of all possible preference vectors for $[\ld]$ items is given by $\lOmega = \{ \ltheta \in \reals^{\ld} : \sum_{i =1}^{\ld} \ltheta_i = 0\}$ and $\lOmega_{2b} = \lOmega \cap [-2b,2b]^{\ld}$. 

Although the analysis can be easily generalized, to simplify notations, we fix 
$\kappa_j = \kappa$ and $\ell_j = \ell$ and assume that the comparison sets $S_j$, $|S_j| = \kappa$, are chosen uniformly at random from the set of $d$ items for all $j \in [n]$. 
The rank-breaking log likelihood function $\Lrb(\ltheta)$ for the set of items $[\ld]$ is given by
\begin{eqnarray}
	\label{eq:likelihood_bl_0}
	\Lrb(\ltheta) &=&
	\sum_{j=1}^n \sum_{a = 1}^{\ell_j}
	\,\lambda_{j,a} \, \Big\{ \sum_{(i, \i) \in E_{j,a}}
	 \,	 \I_{\big\{i, \i \in [\ld] \big\}} \Big(  \theta_{\i}  - \log \Big(e^{\theta_i} + e^{\theta_{\i}}\Big) \,\Big)\, \Big\} \;.
\end{eqnarray} 
We analyze the rank-breaking estimator 
\begin{align} \label{eq:theta_ml_bl}
\widehat{\ltheta}  \;\; \equiv  \;\; \max_{\ltheta \in \lOmega_{2b}} \Lrb(\ltheta)\;.
\end{align}
We further simplify notations by fixing $\lambda_{j,a} = 1$, since  
from Equation \eqref{eq:cauchy-schwartz}, we know that 
the error increases by at most a factor of $4$ due to this sub-optimal choice of the weights, under the special scenario studied in this theorem. 

\begin{theorem} 
	\label{thm:bottoml_upperbound}
	Under the bottom-$\ell$ separators scenario and the  PL model, $S_j$'s are chosen uniformly at random of size $\kappa$ and  
	$n$ partial orderings are sampled over $d$ items parametrized by $\theta^* \in \Omega_b$. 
	For $\ld=\ell d / (2 \kappa)$ and any $\ell\geq 4$, if the effective sample size is large enough such that 
	\begin{align} \label{eq:bottoml_1}
	n\ell \;\; \geq \;\; \bigg(\frac{2^{14}e^{8b}}{\chi^2 }\frac{\kappa^3}{\ell^3}\bigg) d\log d\;, 
	\end{align}
	where 
	\begin{eqnarray}
	 \chi & \equiv & \frac14 \Bigg(1 - \exp\bigg(-\frac{ 2}{9(\kappa-2)}  \,\bigg)\,\Bigg), 
	\end{eqnarray}
	then the {\em rank-breaking} estimator in \eqref{eq:theta_ml_bl} achieves
	\begin{align} \label{eq:bottoml_3}
	\frac{1}{\sqrt{\ld}}\big\|\widehat{\ltheta} - \ltheta^*\big\|_2 \; \leq  \; \frac{128(1+ e^{4b})^2}{\chi}\frac{\kappa^{3/2}}{{\ell}^{3/2}}\sqrt{\frac{d\log d}{n\ell} }\;,
	\end{align}
	with probability at least $1 - 3e^3 d^{-3}$.
\end{theorem}

Consider a scenario where $\kappa=O(1)$ and $\ell=\Theta(\kappa)$. Then, $\chi$ 
is a strictly positive constant, and also $\kappa/\ell$ is s finite constant. 
It follows that rank-breaking requires  the effective sample size $n\ell=O(d\log d / \varepsilon^2 )$ in order to achieve 
arbitrarily small error of $\varepsilon>0$, on the weakest $\ld=\ell\,d/(2 \kappa)$ items.

\section{Real-World Data Sets}
\label{sec:real}
On real-world data sets on sushi preferences \cite{Kam03}, we show that 
the data-driven rank-breaking improves over Generalized Method-of-Moments (GMM) proposed by \cite{ACPX13}. 
This is a widely used data set for rank aggregation, for instance in \cite{ACPX13, APX12, MG15a, LLN15, LB11, LB11b}. 
The data set consists of complete rankings over $10$ types of sushi from $n=5000$ individuals. 
Below, we follow the experimental scenarios of the GMM approach in \cite{ACPX13} for fair comparisons. 

\begin{figure}[h]
 \begin{center}
	\includegraphics[width=.3\textwidth]{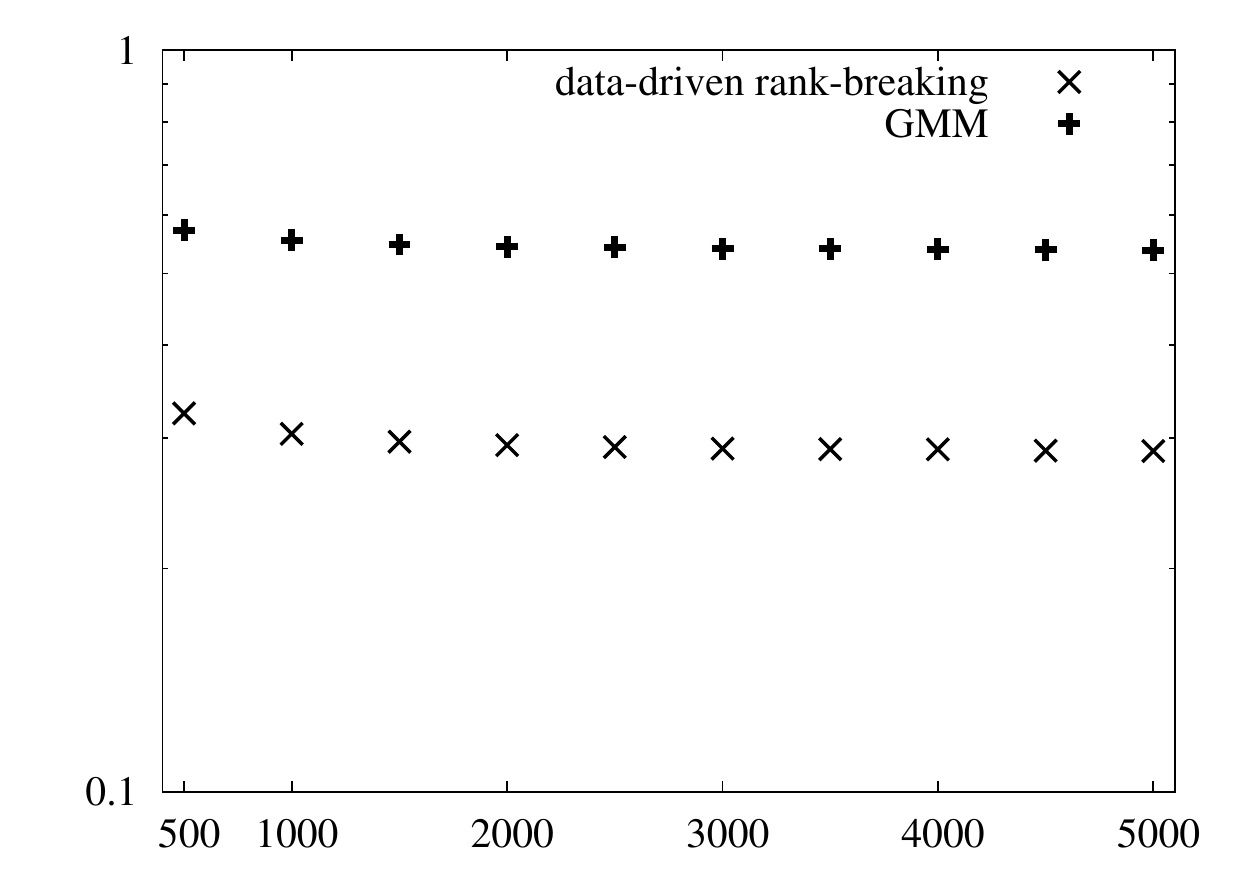}
	\put(-170,50){\small{$\|\widehat\theta-\theta^*\|_2^2$}}	
	\put(-90,-7){sample size \small{$n$}}
	\put(-100,100){\small{Top-$6$ separators}}
	\includegraphics[width=.3\textwidth]{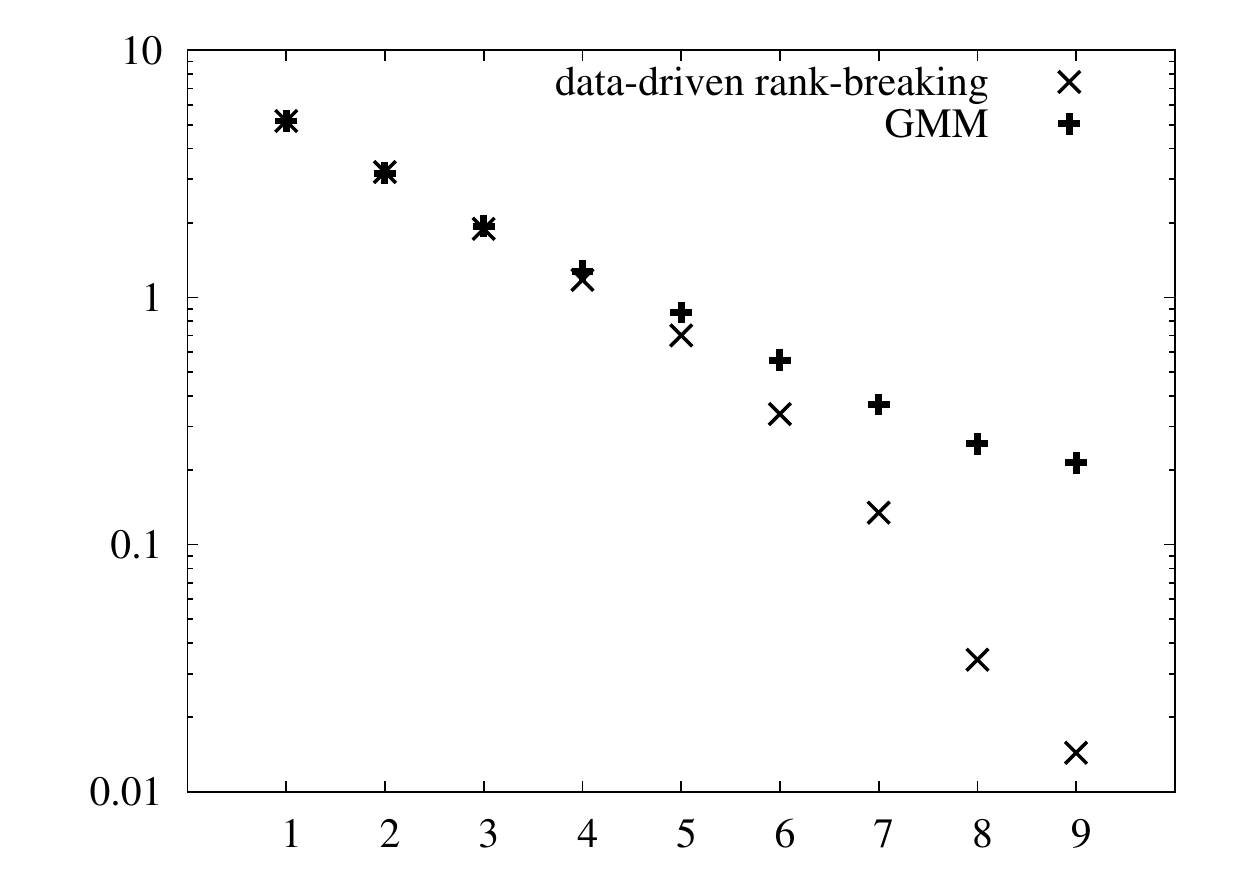}
    \put(-100,100){\small{Top-$\ell$ separators}}
	\put(-115,-7 ){number of separators \small{$\ell$}}  
\end{center}
\caption{The data-driven rank-breaking achieves smaller error compared to the state-of-the-art GMM approach. }
\label{fig:sushi_10_mse}
\end{figure}
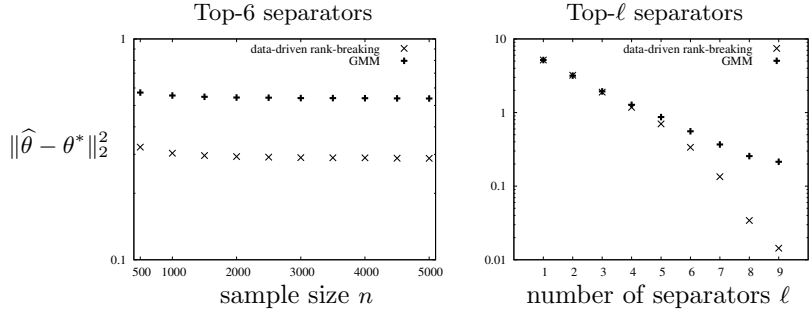

To validate our approach, we first take the estimated PL weights of the 10 types of sushi, using \cite{Hun04} implementation  of the ML estimator,  
over the entire input data of $5000$ complete rankings. 
We take thus created output as the ground truth $\theta^*$. 
To create partial rankings 
and compare the performance of the data-driven rank-breaking to the state-of-the-art GMM approach in Figure \ref{fig:sushi_10_mse}, 
we first fix $\ell=6$ and vary $n$ to simulate top-$\ell$-separators scenario by 
removing the known ordering among bottom $10-\ell$ alternatives for each sample in the data set (left). 
We next fix $n=1000$ and vary $\ell$ and simulate top-$\ell$-separators scenarios  (right). Each point is averaged over $1000$ instances. 
The mean squared error is plotted for both algorithms. 

\begin{figure}[h]
 \begin{center}
	\includegraphics[width=.3\textwidth]{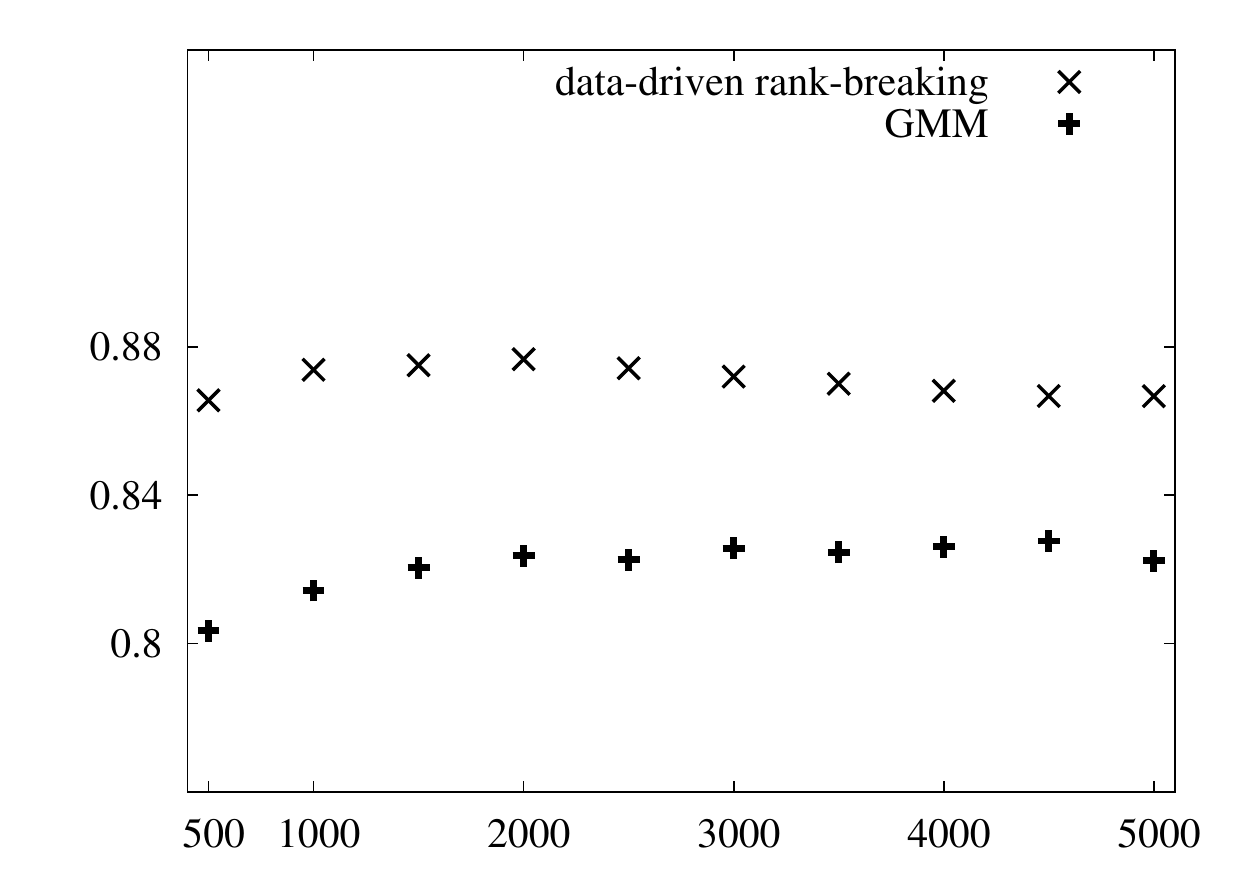}
	\put(-100,100){\small{Top-$6$ separators}}
	\put(-220,50){\small{Kendall Correlation}}	
	\put(-90,-7){sample size \small{$n$}} 
	\includegraphics[width=.3\textwidth]{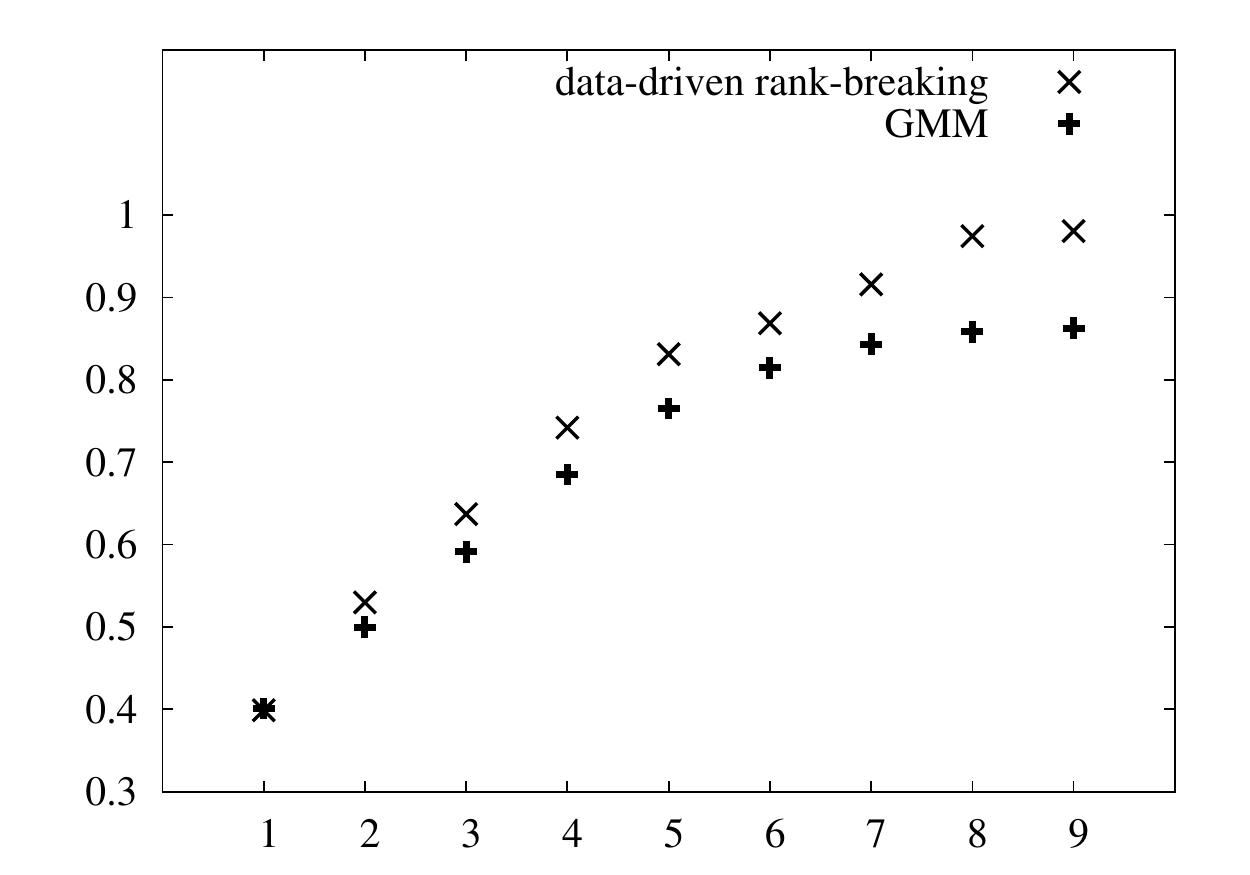}
	\put(-100,100){\small{Top-$\ell$ separators}}
	\put(-115,-7 ){number of separators \small{$\ell$}} 
\end{center}
\caption{The data-driven rank-breaking achieves larger Kendall rank correlation compared to the state-of-the-art GMM approach. }
\label{fig:sushi_10_ken}
\end{figure}
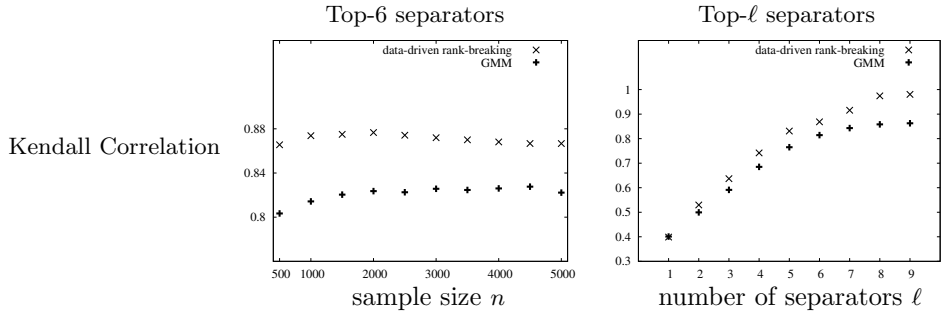

Figure \ref{fig:sushi_10_ken} illustrates the Kendall rank correlation of the rankings estimated by the two algorithms and the ground truth. 
Larger value indicates that the estimate is closer to the ground truth, and the data-driven rank-breaking outperforms the state-of-the-art GMM approach. 

To validate whether PL model is the right model to explain the sushi data set, 
we compare the data-driven rank-breaking, MLE for the PL model, GMM for the PL model, Borda count and Spearman's footrule optimal aggregation. 
We measure the Kendall rank correlation between the estimates and the samples and show the result in Table \ref{tab:sushi_10_all}.
In particular, if $\sigma_1,\sigma_2,\cdots, \sigma_n$ denote sample rankings and $\widehat{\sigma}$ denote the aggregated ranking then the correlation value is $(1/n)\sum_{i = 1}^n \big(1-\frac{4\mathcal{K}(\widehat{\sigma},\sigma_i)}{\kappa(\kappa-1)}\big)$, where $\mathcal{K}(\sigma_1,\sigma_2) = \sum_{i < j \in [\kappa]} \mathbb{I}_{\{(\sigma_1^{-1}(i) - \sigma_1^{-1}(j))(\sigma_2^{-1}(i) - \sigma_2^{-1}(j)) < 0 \}}$. The results are reported for different  number of samples $n$ and different values of $\ell$ under the top-$\ell$ separators scenarios. 
When $\ell=9$, we are using all the complete rankings, and all algorithms are efficient. 
When $\ell < 9$, we have  partial orderings, and Spearman's  footrule optimal aggregation is NP-hard. 
We instead use scaled footrule aggregation (SFO) given in \cite{DKNS01}. 
Most approaches achieve similar performance, except for the Spearman's footrule. 
The proposed data-driven rank-breaking achieves a slightly worse correlation compared to other approaches. 
However, note that none of the algorithms are 
necessarily maximizing the Kendall correlation, and are not expected to be particularly good in this metric. 

\begin{table}[h]
\begin{center}
  \begin{tabular}{  C{2.5cm} |  C{1.5cm}  C{2.cm} C{1.5cm}   C{1.5cm}   C{2cm} }
    & MLE under PL & data-driven RB & GMM & Borda count & Spearman's footrule  \\ \hline
    $n = 500$, $\ell = 9$ & 0.306 & 0.291 & 0.315 & 0.315 & 0.159   \\ 
    $n = 5000$, $\ell = 9$  & 0.309 & 0.309& 0.315 &0.315 & 0.079 \\ 
    $n = 5000$, $\ell = 2$  & 0.199 &0.199 & 0.201&0.200 & 0.113 \\ 
    $n = 5000$, $\ell = 5$  & 0.217& 0.200& 0.217& 0.295& 0.152 \\     
  \end{tabular}
  \caption{Kendall rank correlation on sushi data set.}
\label{tab:sushi_10_all}
\end{center}
\end{table}

We compare our algorithm with the GMM algorithm on two other real-world data-sets as well. We use jester data set \cite{GRG01} that consists of over $4.1$ million continuous ratings between $-10$ to $+10$ of $100$ jokes from $48,483$ users. The average number of jokes rated by an user is $72.6$ with minimum and maximum being $36$ and $100$ respectively. We convert continuous ratings into ordinal rankings. This data-set has been used by \cite{MP00, PD05, CMR07, LM07} for rank aggregation and collaborative filtering.

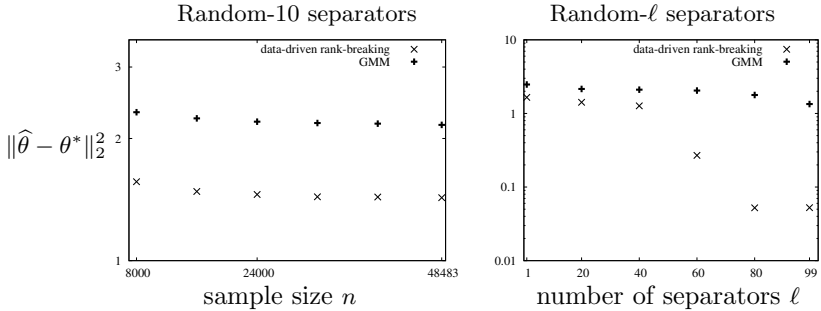
\begin{figure}[h]
 \begin{center}
	\includegraphics[width=.3\textwidth]{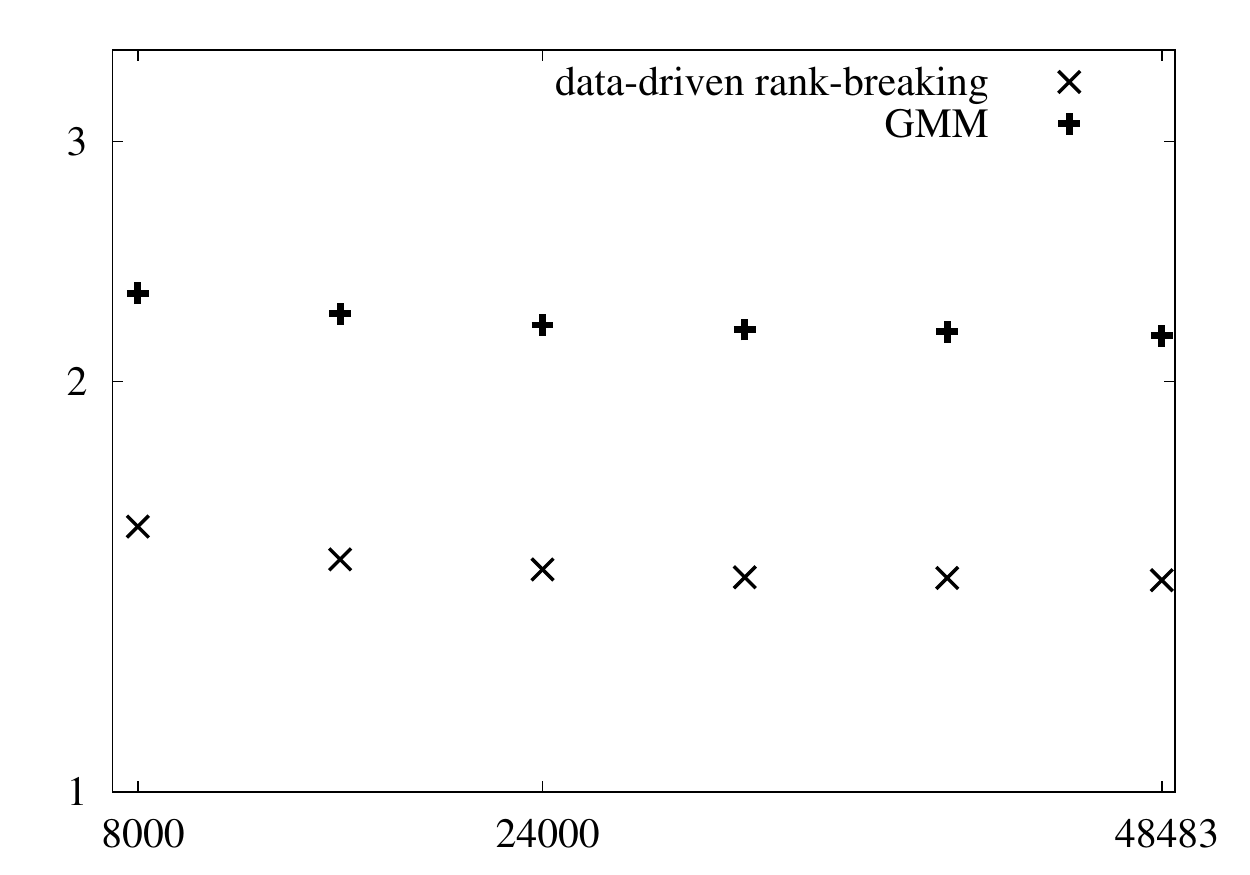}
	\put(-175,50){\small{$\|\widehat\theta-\theta^*\|_2^2$}}	
	\put(-100,-7){sample size \small{$n$}}
	\put(-110,100){\small{Random-$10$ separators}}
	\includegraphics[width=.3\textwidth]{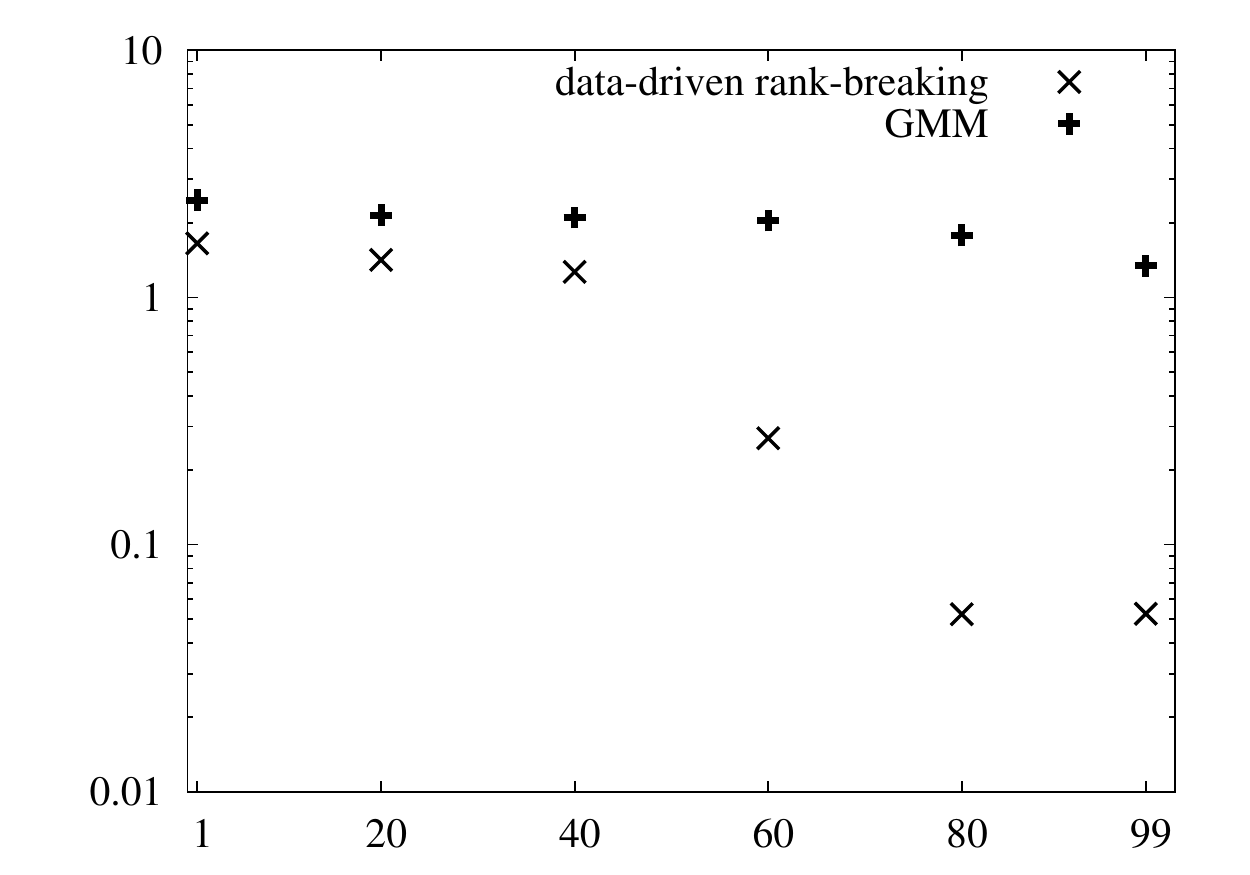}
    \put(-110,100){\small{Random-$\ell$ separators}}
	\put(-115,-7 ){number of separators \small{$\ell$}} 
\end{center}
\caption{jester data set: The data-driven rank-breaking achieves smaller error compared to the state-of-the-art GMM approach. }
\label{fig:jest}
\end{figure}

Similar to the settings of sushi data experiments, we take the estimated PL weights of the 100 jokes over all the rankings as ground truth. Figure \ref{fig:jest} shows comparative performance of the data-driven rank-breaking and the GMM for the two scenarios. We first fix $\ell = 10$ and vary $n$ to simulate random-$10$ separators scenario (left). We next take all the rankings $n = 73421$ and vary $\ell$ to simulate random-$\ell$ separators scenario (rights). Since sets have different sizes, while varying $\ell$ we use full breaking if the setsize is smaller than $\ell$. Each point is averaged over $100$ instances. The mean squared error is plotted for both algorithms. 

We perform similar experiments on American Psychological Association (APA) data-set \cite{Dia89}. The APA elects a president each year by asking each member to rank order a slate of five candidates. The data-set represents full rankings given by 5738 members of the association in 1980's election. The mean squared error is plotted for both algorithms under the settings similar to that of jester data-set.

\begin{figure}[h]
 \begin{center}
	\includegraphics[width=.3\textwidth]{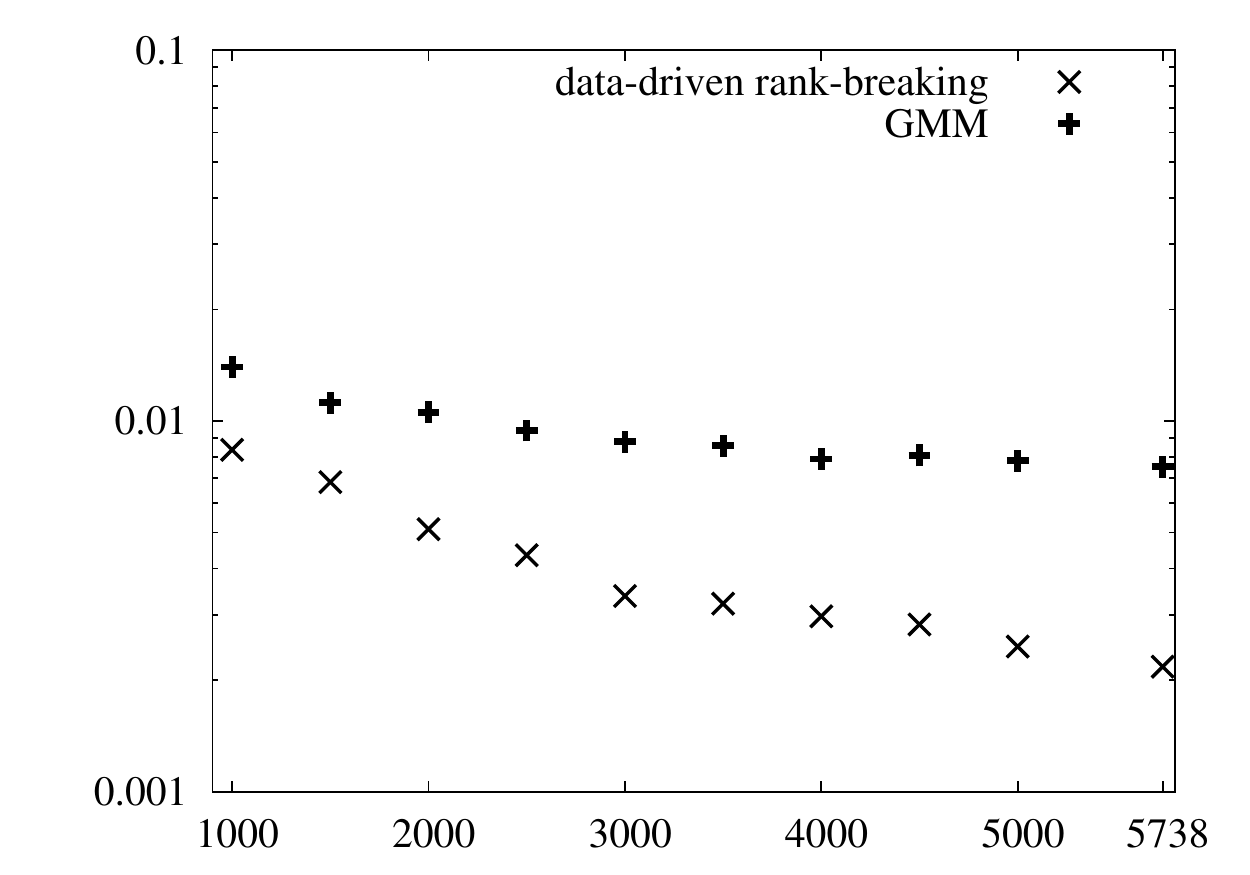}
	\put(-175,50){\small{$\|\widehat\theta-\theta^*\|_2^2$}}	
	\put(-90,-7){sample size \small{$n$}} 
	\put(-110,100){\small{Random-$3$ separators}}
	\includegraphics[width=.3\textwidth]{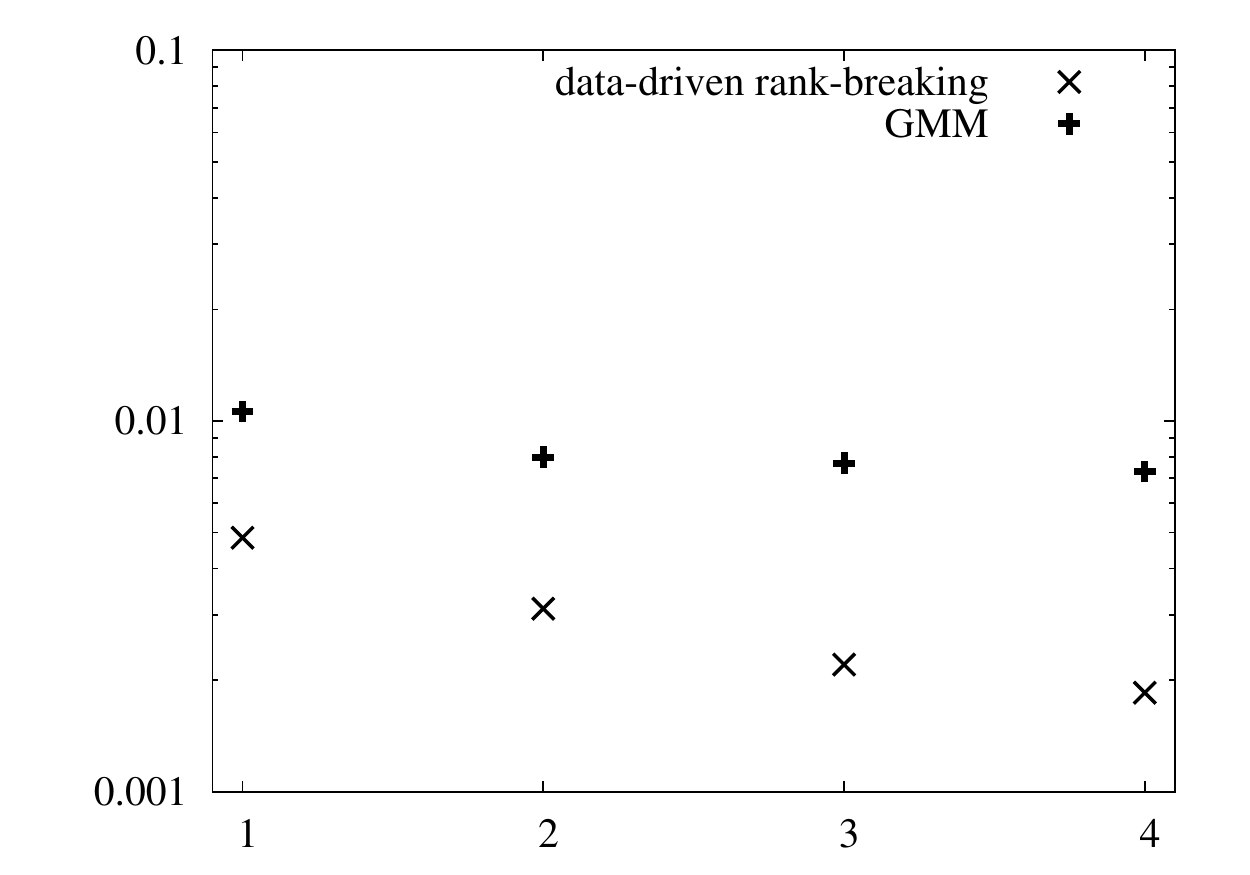}
    \put(-110,100){\small{Random-$\ell$ separators}}
	\put(-115,-7 ){number of separators \small{$\ell$}} 
\end{center}
\caption{APA data set: The data-driven rank-breaking achieves smaller error compared to the state-of-the-art GMM approach. }
\label{fig:apa}
\end{figure}
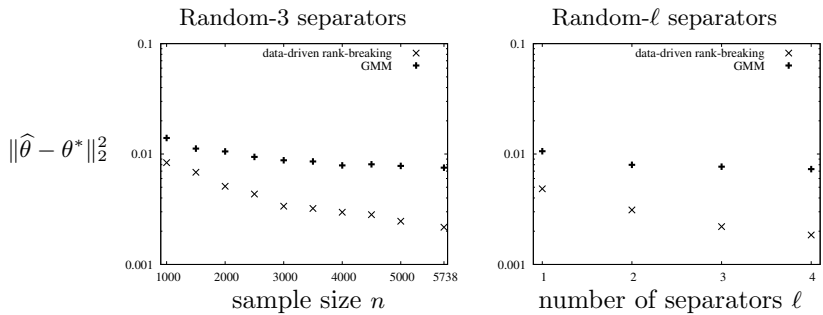

\section{Related Work} 

Initially motivated by elections and voting, rank aggregation has been a topic of mathematical interest 
dating back to Condorcet and Borda \cite{Con1785,Borda1781}. 
Using probabilistic models to infer preferences has been popularized 
in operations research community for applications such as assortment optimization and revenue management. The PL model studied in this paper is a special case of MultiNomial Logit (MNL) models commonly used in 
discrete choice modeling, which has a long history in operations research \cite{McF80}. 
Efficient inference algorithms has been proposed to either find the MLE efficiently or approximately, 
such as the iterative approaches in \cite{Ford57,Dyk60}, 
minorization-maximization approach in \cite{Hun04}, and 
Markov chain approaches in \cite{NOS12,MG15}.
These approaches are shown to achieve minimax optimal error rate 
in the traditional comparisons scenarios. 
Under the pairwise comparisons scenario, 
Negahban et al. \cite{NOS12} provided Rank Centrality that provably achieves minimax optimal error rate 
for randomly chosen pairs, which was later generalized to arbitrary pairwise comparisons in 
\cite{NOS14}. The analysis shows the explicit dependence on the topology of data shows that 
the spectral gap of comparisons graph similar to the one presented in this paper.
This analysis was generalized to $k$-way comparisons in \cite{HOX14} and 
generalized to best-out-of-$k$ comparisons with sharper bounds in \cite{SBB15}. 
In an effort to give a guarantee for exact recovery of the top-$\ell$ items in the ranking, 
Chen et al. in \cite{CS15} proposed a new algorithm based on Rank Centrality that provides a tighter error bound for $L_\infty$ norm, as opposed to the existing $L_2$ error bounds. 
Another interesting direction in learning to rank is non-parametric learning from paired comparisons, initiated in several recent papers such as \cite{DMJ10,RA14,SBGW15,SW15}.

More recently, a more general problem of learning {\em personal} preferences from ordinal data has been studied \cite{YJJ13,LB11b,DIS15}. The MNL model provides a natural generalization of the PL model to this problem. 
When users are classified into a small number of groups with same preferences, mixed MNL model can be learned from data as studied in \cite{AOSV14,OS14,WXSMLH15}. A more general scenario is 
when each user has his/her individual preferences, but inherently represented by  a lower dimensional feature. This problem was first posed as an inference problem in \cite{LN14} where convex relaxation of nuclear norm minimization was proposed with provably optimal guarantees. This was later generalized to $k$-way comparisons in \cite{OTX15}. A similar approach was studied with a different guarantees and assumptions in \cite{PNZSD15}. 
Our algorithm and ideas of rank-breaking can be directly applied to this collaborative ranking under MNL, with the same guarantees for consistency in the asymptotic regime where sample size grows to infinity. However, the analysis techniques for MNL rely on stronger assumptions on how the data is collected, and especially on the independence of the samples. It is not immediate how the analysis techniques developed in this paper can be applied to learn MNL. 

In an orthogonal direction,  new discrete choice models with sparse structures has been proposed recently in \cite{FJS09} and optimization algorithms for revenue management has been proposed \cite{FJS13}. 
In a similar direction, new discrete choice models based on Markov chains has been introduced in \cite{BGG13}, and corresponding revenue management algorithms has been studied in \cite{FT14}. 
However, typically these models are analyzed in the asymptotic regime with infinite samples, 
with the exception of \cite{AS11}. 
A non-parametric choice models for pairwise comparisons also have been studied in \cite{RA14,SBGW15}. 
This provides an interesting opportunities to studying learning to rank for these new choice models. 

We consider a fixed design setting, where inference is separate from data collection. 
There is a parallel line of research which focuses on adaptive ranking, mainly based on pairwise comparisons. When performing sorting from noisy pairwise comparisons,  
Braverman et al. in \cite{BM09} proposed efficient approaches and provided performance guarantees. 
Following this work, there has been recent advances in adaptive ranking \cite{Ail11,JN11,MG15a}.

\section{Discussion}

We study the problem of learning the PL model from ordinal data. 
Under the traditional data collection scenarios, 
several efficient algorithms find the maximum likelihood estimates and at the same time provably 
achieve minimax optimal performance. 
However, for some non-traditional scenarios, computational complexity of finding the maximum likelihood estimate can scale super-exponentially in the problem size. 
We provide the first finite-sample analysis of computationally efficient estimators known as rank-breaking estimators. This provides guidelines for choosing the weights in the estimator to achieve optimal performance, and also explicitly shows how the accuracy depends on the topology of the data. 

This paper provides the first analytical result in the sample complexity of rank-breaking estimators, and quantifies the price we pay in accuracy for the computational gain. 
In general, more complex higher-order rank-breaking can also be considered, 
where instead of breaking a partial ordering into a collection of paired comparisons, 
we break it into a collection of higher-order comparisons.   
The resulting higher-order rank-breakings will enable us to traverse the whole spectrum of 
computational complexity between the pairwise rank-breaking and the MLE. 
We believe this paper opens an interesting new direction towards understanding the whole spectrum of such approaches. 
However, analyzing the Hessian of the corresponding objective function is significantly more involved and requires new technical innovations.

\section{Proofs}

\subsection{Proof of Theorem \ref{thm:main2}}
\label{sec:proof_main2}
We prove a more general result  
for an arbitrary choice of the parameter $\lambda_{j,a}>0$ for all $j\in[n]$ and $a\in[\ell_j]$. 
The following theorem  proves the (near)-optimality of the choice of $\lambda_{j,a}$'s proposed in 
 \eqref{eq:deflambda}, and implies the corresponding error bound as a corollary.  

\begin{theorem} \label{thm:main}
	Under the hypotheses of Theorem \ref{thm:main2} and any $\lambda_{j,a}$'s, 
	the rank-breaking estimator achieves 
	\begin{align}
	\label{eq:main1}
	\frac{1}{\sqrt{d}} \big\|\,\widehat{\theta} - \theta^* \,\big\|_2 \; \,\leq\,  \; \frac{4\sqrt{2}e^{4b}(1+ e^{2b})^2 \sqrt{d \log d}  }{\alpha\, 	\gamma} \frac{\sqrt{\sum_{j=1}^n  \sum_{a=1}^{\ell_j}  \big(\lambda_{j,a}\big)^2 \big(\kappa_j - p_{j,a}\big)\big(\kappa_j- p_{j,a}+1\big)}}{ \sum_{j = 1}^n \sum_{a = 1}^{\ell_j} \lambda_{j,a}(\kappa_j - p_{j,a})}\;,
	\end{align}
	with probability at least $ 1- 3e^{3}d^{-3}$, if 
	\begin{align} \label{eq:main2}
	\sum_{j = 1}^n \sum_{a = 1}^{\ell_j} \lambda_{j,a}(\kappa_j - p_{j,a}) \;\;\geq\;\; 2^{6}e^{18b} \frac{\eta\delta}{\alpha^2 \beta \gamma^2\tau} d\log d \;,
	\end{align}
	where $\gamma$, $\eta$, $\tau$, $\delta$, $\alpha$, $\beta$,    are now functions of $\lambda_{j,a}$'s and defined in 
	\eqref{eq:gamma_def}, \eqref{eq:eta_def}, \eqref{eq:tau_def}, \eqref{eq:delta_def} and \eqref{eq:lambda2_L2}.
\end{theorem}

We first claim that $\lambda_{j,a} = 1/(\kappa_j-p_{j,a}+1)$ is the optimal choice for minimizing the above upper bound on the error. 
From Cauchy-Schwartz inequality and the fact that all terms are non-negative, we have that 
\begin{align} \label{eq:cauchy-schwartz}
	\frac{\sqrt{\sum_{j = 1}^n \sum_{a = 1}^{\ell_j} \big(\lambda_{j,a}\big)^2(\kappa_j - p_{j,a})(\kappa_j - p_{j,a}+1)}}
	{\sum_{j = 1}^n \sum_{a = 1}^{\ell_j} \lambda_{j,a}(\kappa_j - p_{j,a})} 
	\;\;\geq\;\; \frac{1}{\sqrt{ \sum_{j = 1}^n \sum_{a = 1}^{\ell_j} \frac{(\kappa_j - p_{j,a})}{(\kappa_j - p_{j,a}+1)}}}\,,
\end{align}
where $\lambda_{j,a} = 1/(\kappa_j-p_{j,a}+1)$ achieves the universal lower bound on the right-hand side with an equality. 
Since $\sum_{j = 1}^n \sum_{a = 1}^{\ell_j} \frac{(\kappa_j - p_{j,a})}{(\kappa_j - p_{j,a}+1)} \geq \sum_{j=1}^n \ell_j$, 
substituting this into \eqref{eq:main1} gives the desired error bound in \eqref{eq:main22}.
Although we have identified the optimal choice of $\lambda_{j,a}$'s, 
we choose a slightly different value of $\lambda=1/(\kappa_j-p_{j,a})$ for the analysis. 
This achieves the same desired error bound in \eqref{eq:main22}, and significantly simplifies the notations of the sufficient conditions. 

We first define all the parameters in the above theorem for general $\lambda_{j,a}$. 
With a slight abuse of notations, we use the same notations for $\H$, $L$, $\alpha$ and $\beta$ for both the general $\lambda_{j,a}$'s and also the specific choice of $\lambda_{j,a}=1/(\kappa_j-p_{j,a})$. It should be clear from the context what we mean in each case. 
Define 
\begin{eqnarray}
	\tau &\equiv & \min_{j \in [n]} \tau_j  \;, \; \;\;\;\;\text{where}\;\;  \tau_{j} \equiv  \frac{\sum_{a = 1}^{\ell_j} \lambda_{j,a}(\kappa_j - p_{j,a})}{\ell_j}\label{eq:tau_def}\\
	\delta_{j,1} & \equiv & \bigg\{ \max_{a \in [\ell_j]} \Big\{\lambda_{j,a}(\kappa_j - p_{j,a})\Big\} + \sum_{a = 1}^{\ell_j} \lambda_{j,a} \bigg\} \;\;, \;\text{and}\;\;\;\;\;\; \delta_{j,2} \equiv  \sum_{a = 1}^{\ell_j} \lambda_{j,a} \label{eq:delta12_def} \\
	\delta & \equiv & \max_{j \in [n]} \bigg\{ 4 \delta_{j,1}^2 + \frac{2\big(\delta_{j,1}\delta_{j,2} + \delta_{j,2}^2\big)\kappa_j}{\eta_{j}\ell_j} \bigg\} \;\;\,. \label{eq:delta_def} 
\end{eqnarray}
Note that $\delta \geq \delta_{j,1}^2  \geq \max_a \lambda_{j,a}^2 (\kappa_j-p_{j,a})^2 \geq \tau^2$, 
and for the choice of $\lambda_{j,a}=1/(\kappa_j-p_{j,a})$ it simplifies as  $\tau=\tau_j=1$.  
We next define a comparison graph $\H$ for general $\lambda_{j,a}$, which recovers the 
proposed comparison graph  for the optimal choice of $\lambda_{j,a}$'s

\begin{definition} \label{def:comparison_graph2} (Comparison graph $\H$). 
Each item $i \in [d]$ corresponds to a vertex $i$. For any pair of vertices $i,\i$, there is a weighted edge between them if there exists a set $S_j$ such that $i, \i \in S_j$; the weight equals $\sum_{j: i,\i \in S_j} \frac{\tau_{j}\ell_j}{\kappa_j(\kappa_j-1)}$.  
\end{definition}

Let $A$ denote the weighted adjacency matrix, and 
let $D =  {\rm diag}(A \vect{1})$. Define, 
\begin{align}\label{eq:posl_Dlmax}
 	D_{\max} \;\;\equiv \;\; \max_{i \in [d]} D_{ii} \;= \;\max_{i \in [d]} \bigg\{ \sum_{j: i \in S_j} \frac{\tau_{j}\ell_j}{\kappa_j} \bigg\} \;\;\geq\;\; \tau_{\rm min} \max_{i \in [d]} \bigg\{ \sum_{j: i \in S_j} \frac{\ell_j}{\kappa_j} \bigg\}  \,. 
\end{align}
Define graph Laplacian $L $ as $L  = D - A$, i.e., 
\begin{align} \label{eq:comparison2_L}
	L \; =\; \sum_{j = 1}^n \frac{\tau_{j}\ell_j}{\kappa_j(\kappa_j-1)}  \sum_{i<\i \in S_j} (e_i - e_{\i})(e_i - e_{\i})^\top.
\end{align}
Let $ 0 = \lambda_1(L) \leq \lambda_2(L) \leq \cdots \leq \lambda_d(L)$ denote the sorted eigenvalues of 
$L$. 
Note that $\Tr(L ) = \sum_{i =1}^d \sum_{j: i \in S_j}\tau_{j}\ell_j/\kappa_j = \sum_{j = 1}^n \tau_{j}\ell_j$.  Define $\alpha $ and $\beta $ such that
\begin{align}\label{eq:lambda2_L2}
	\alpha  \equiv \frac{\lambda_2(L )(d-1)}{\Tr(L)}  = \frac{\lambda_2(L)(d-1)}{ \sum_{j = 1}^n \tau_{j}\ell_j} \;\; \text{and} \;\; \beta  \equiv \frac{\Tr(L)}{d D_{\max}} = \frac{ \sum_{j = 1}^n \tau_{j}\ell_j}{d D_{\max}} \;. 
	\end{align}

For the proposed choice of  $\lambda_{j,a} = 1/(\kappa_j-p_{j,a})$, we have $\tau_j = 1$ and 
the definitions  of $\H$, $L$, $\alpha$, and $\beta$ reduce to those defined in Definition \ref{def:comparison_graph1}. 
We are left to prove an upper bound, $\delta\leq 32 (\log(\ell_{\max}+2))^2$, which 
implies the sufficient condition   in \eqref{eq:main21} and  finishes the proof of Theorem \ref{thm:main2}. 
We have, 
\begin{eqnarray}\label{eq:main4}
\delta_{j,1} =  \max_{a \in [\ell_j]} \Big\{\lambda_{j,a}(\kappa_j - p_{j,a})\Big\} + \sum_{a = 1}^{\ell_j} \lambda_{j,a}
&=& 1 + \sum_{a = 1}^{\ell_j} \frac{1}{\kappa_j - p_{j,a}} \nonumber\\
&\leq& 1 + \sum_{a=1}^{\ell_j} \frac{1}{a}\nonumber\\
&\leq& 2\log(\ell_j+2) \,, 
\end{eqnarray}  
where in the first inequality follows from taking the worst case for the positions, i.e. $p_{j,a}= \kappa_j-\ell_j+a-1$ 
Using the fact that 
 for any integer $x$, $\sum_{a=0}^{\ell-1} 1/(x+a) \leq \log((x +\ell -1)/(x -1))$, we also have
\begin{eqnarray}\label{eq:main5}
\frac{\delta_{j,2}\kappa_j}{\eta_j\ell_j} &\leq& \sum_{a = 1}^{\ell_j} \frac{1}{\kappa_j - p_{j,a}} \frac{\max{\{\ell_j,\kappa_j - p_{j,\ell_j}\}}}{\ell_j}\nonumber\\
 &\leq& \min\Big\{\,\log(\ell_j+2) \,,\, \log\Big(\frac{\kappa_j-p_{j,\ell_j}+\ell_j -1}{\kappa_j-p_{j,\ell_j} -1 }\Big)\,\Big\} \frac{\max{\{\ell_j,\kappa_j - p_{j,\ell_j}\}}}{\ell_j}
 \nonumber\\
 &\leq& \frac{\log(\ell_j+2)\ell_j}{\max{\{\ell_j,\kappa_j - p_{j,\ell_j} -1}\}} \frac{\max{\{\ell_j,\kappa_j - p_{j,\ell_j}\}}}{\ell_j}\nonumber\\
 &\leq& 2\log(\ell_j+2)\,,
\end{eqnarray} 
where the first inequality follows from the definition of $\eta_j$, Equation \eqref{eq:eta_def}.  
From \eqref{eq:main4}, \eqref{eq:main5}, and the fact that $\delta_{j,2}\leq\log(\ell_j+2)$,  we have
\begin{eqnarray*} \label{eq:main6}
\delta = \max_{j \in [n]} \bigg\{ 4 \delta_{j,1}^2 + \frac{2\big(\delta_{j,1}\delta_{j,2} +\delta_{j,2}^2\big)\kappa_j}{\eta_{j}\ell_j} \bigg\} \;\;\leq\;\; 28 (\log(\ell_{\max} +2))^2\,.
\end{eqnarray*}

\subsection{Proof of Theorem \ref{thm:main}}
We first introduce two key technical lemmas. 
In the following lemma we show that $\E_{\theta^*}[\nabla \Lrb(\theta^*)] = 0$ and provide a bound on the deviation of $\nabla \Lrb(\theta^*)$ from its mean. The expectation $\E_{\theta^*}[\cdot]$ is with respect to  the randomness in the samples drawn according to $\theta^*$.  
The log likelihood Equation \eqref{eq:likelihood_0} can be rewritten as 
\begin{align}
	\label{eq:likelihood}
	\Lrb(\theta) = 
	\sum_{j=1}^n \sum_{a = 1}^{\ell_j}\sum_{i < \i \in S_j}\I_{\big\{(i,\i) \in \Gja\big\}} 
	\lambda_{j,a} \Big(\theta_i\I_{\big\{\sigma_j^{-1}(i) < \sigma_j^{-1}(\i)\big\}} + \theta_{\i}\I_{\big\{\sigma_j^{-1}(i) > \sigma_j^{-1}(\i)\big\}} - \log \Big(e^{\theta_i} 
	+ e^{\theta_{\i}}\Big) \Big)\;.
\end{align}
We use $(i,\i) \in G_{j,a}$ to mean either $(i,\i)$ or $(\i,i)$ belong to $E_{j,a}$. 
Taking the first-order partial derivative of $\Lrb(\theta)$, we get 
\begin{align}\label{eq:liklihood_grad}
\nabla_i\Lrb(\theta^*) \;\, =\;\, \sum_{j:i\in S_j} \sum_{a=1}^{\ell_j}  \sum_{\substack{\i \in S_j \\ \i \neq i}} \,\lambda_{j,a}\,\I_ {\big\{(i,\i) \in G_{j,a}\big\}}  \,\Bigg(\I_{\big\{\sigma_j^{-1}(i) < \sigma_j^{-1}(\i)\big\}}  - \frac{\exp(\theta_i^*)}{\exp(\theta_i^*) + \exp(\theta_{\i}^*)} \Bigg)\;.
\end{align}

\begin{lemma}\label{lem:gradient_topl}
	Under the hypotheses of Theorem \ref{thm:main2}, with probability at least $1 - 2e^{3}d^{-3}$,
	\begin{align*}
	\big\|\nabla\Lrb(\theta^*)\big\|_2 \;\;\leq\;\; \sqrt{ 6\log d \, \sum_{j=1}^n  \sum_{a=1}^{\ell_j}  \big(\lambda_{j,a}\big)^2 \big(\kappa_j - p_{j,a}\big)\big(\kappa_j- p_{j,a}+1\big)} \,.  
	\end{align*}
\end{lemma}

The Hessian matrix $H(\theta) \in \cS^d$ with $H_{i\i}(\theta) = \frac{\partial^2\Lrb(\theta)}{\partial\theta_i \partial\theta_{\i}}$ is given by
\begin{align} \label{eq:hessian}
H(\theta) = -\sum_{j=1}^n \sum_{a=1}^{\ell_j} \sum_{i<\i \in S_j} \I_{\big\{(i,\i) \in G_{j,a}\big\}} \lambda_{j,a} \Bigg( (e_i - e_{\i})(e_i - e_{\i})^\top \frac{\exp(\theta_i + \theta_{\i})}{[\exp(\theta_i) + \exp(\theta_{\i})]^2}\Bigg).
\end{align}
It follows from the definition that $-H(\theta)$ is positive semi-definite for any $\theta \in \reals^d$. The smallest eigenvalue of $-H(\theta)$ is equal to zero and the corresponding eigenvector is all-ones vector. The following lemma lower bounds its second smallest eigenvalue $\lambda_2(-H(\theta))$. 

\begin{lemma}\label{lem:hessian_positionl}
	Under the hypotheses of Theorem \ref{thm:main2}, 
	if  
	\begin{align}\label{eq:posl_lam_cond}
	\sum_{j = 1}^n \sum_{a = 1}^{\ell_j} \lambda_{j,a}(\kappa_j - p_{j,a}) \geq 2^{6}e^{18b} \frac{\eta\delta}{\alpha^2\beta\gamma^2\tau} d\log d
	\end{align}
	then with probability at least $ 1- d^{-3}$, the following holds for any $\theta\in\Omega_b$: 
	\begin{align} \label{eq:lambda2_bound_positionl}
	\lambda_2(-H(\theta)) \;\geq\; \frac{e^{-4b}}{(1+e^{2b})^2}\frac{\alpha \gamma}{d-1}  \sum_{j = 1}^n \sum_{a = 1}^{\ell_j} \lambda_{j,a}(\kappa_j - p_{j,a})\,. 
	\end{align}
\end{lemma}

Define $\Delta = \widehat{\theta} - \theta^*$. It follows from the definition that $\Delta$ is orthogonal to the all-ones vector. By the definition of $\hat{\theta}$ as the optimal solution of the optimization  \eqref{eq:theta_ml}, we know that 
$\Lrb(\widehat{\theta}) \geq \Lrb(\theta^*)$ and thus 
\begin{eqnarray}
\Lrb(\widehat{\theta}) - \Lrb(\theta^*) - \langle\nabla\Lrb(\theta^*),\Delta\rangle \;\geq\; -\langle\nabla\Lrb(\theta^*),\Delta\rangle \;\geq\; -\norm{\nabla\Lrb(\theta^*)}_2\norm{\Delta}_2, \label{eq:thm_ml_1}
\end{eqnarray} 
where the last inequality follows from the Cauchy-Schwartz inequality. By the mean value theorem, 
there exists a $\theta = a\widehat{\theta} + (1-a)\theta^*$ for some $a \in [0,1]$ such that $\theta \in \Omega_b$ and 
\begin{eqnarray}\label{eq:thm_ml_2}
\Lrb(\widehat{\theta}) - \Lrb(\theta^*) - \langle\nabla\Lrb(\theta^*),\Delta\rangle \; =\; \frac{1}{2}\Delta^\top H(\theta)\Delta \leq -\frac{1}{2}\lambda_2(-H(\theta))\norm{\Delta}_2^2,
\end{eqnarray}  
where the last inequality holds because the Hessian matrix $-H(\theta)$ is positive semi-definite with $H(\theta)\vect{1} = \vect{0}$ and $\Delta^\top\vect{1} = 0$. Combining \eqref{eq:thm_ml_1} and \eqref{eq:thm_ml_2},
\begin{eqnarray} \label{eq:thm_ml_3}
\norm{\Delta}_2 \;\;\leq\;\; \frac{2\norm{\nabla\Lrb(\theta^*)}_2}{\lambda_2(-H(\theta))}.
\end{eqnarray}  
Note that $\theta \in \Omega_b$ by definition. Theorem \ref{thm:main} follows by combining Equation \eqref{eq:thm_ml_3} with Lemma \ref{lem:gradient_topl} and Lemma \ref{lem:hessian_positionl}.

\subsubsection{Proof of Lemma \ref{lem:gradient_topl}}
The idea of the proof is to view $\nabla\Lrb(\theta^*)$ as the final value of a discrete time vector-valued martingale with values in $\reals^d$. Define $\nabla\L_{G_{j,a}}(\theta^*)$ as the gradient vector arising out of each rank-breaking graph $\{G_{j,a}\}_{j\in [n], a \in [\ell_j]}$ that is
\begin{align}
	\label{eq:rev1}
	\nabla_i\L_{G_{j,a}}(\theta^*) \equiv \sum_{\substack{\i \in S_j \\ \i \neq i}} \,\lambda_{j,a}\,\I_ {\big\{(i,\i) \in G_{j,a}\big\}}  \,\Bigg(\I_{\big\{\sigma_j^{-1}(i) < \sigma_j^{-1}(\i)\big\}}  - \frac{\exp(\theta_i^*)}{\exp(\theta_i^*) + \exp(\theta_{\i}^*)} \Bigg)\;.
\end{align}
Consider $\nabla\L_{G_{j,a}}(\theta^*)$ as the incremental random vector in a martingale of $\sum_{j=1} \ell_j$ time steps. Lemma \ref{lem:consistency} shows that the expectation of each incremental vector is zero. Observe that the conditioning event $\{i'' \in S \,:\,\sigma^{-1}(i'')<p_{j,a} \}$ given in  \eqref{eq:grad_eq8} is equivalent to conditioning on the history $\{G_{j,a'}\}_{a'<a}$. Therefore, using the assumption that the rankings $\{\sigma_j\}_{j \in [n]}$ are mutually independent, we have that the conditional expectation of $\nabla\L_{G_{j,a}}(\theta^*)$ conditioned on $\{G_{j',a''}\}_{j' < j, a'' \in [\ell_{j'}]}$ is zero. 
Further, the conditional expectation of $\nabla\L_{G_{j,a}}(\theta^*)$ is zero even when 
conditioned on the rank breaking due to previous separators $\{G_{j,a'}\}_{a'<a}$ that are ranked higher (i.e. $a'<a$), which follows from the next lemma. 

\begin{lemma} \label{lem:consistency}
For a position-$p$ rank breaking graph $G_p$, defined over a set of items $S$, where $p \in [|S|-1]$, 
\begin{align} \label{eq:grad_eq6}
	\P\Big[\sigma^{-1}(i) < \sigma^{-1}(\i) \;\Big|\;  \big(i,\i\big) \in G_p  \Big] \;=\; \frac{\exp(\theta^*_{i})}{\exp(\theta^*_{i})+\exp(\theta^*_{i'})} \;,
\end{align}
for all $i,i'\in S$ and also 
\begin{align} \label{eq:grad_eq8}
	\P\Big[\sigma^{-1}(i) < \sigma^{-1}(\i) \;\Big|\;  \big(i,\i\big) \in G_p \text{ and } \{i'' \in S \,:\,\sigma^{-1}(i'')<p \} \Big] \;=\; \frac{\exp(\theta^*_{i})}{\exp(\theta^*_{i})+\exp(\theta^*_{i'})} \;.
\end{align}
\end{lemma}

This is one of the key technical lemmas since it implies that the proposed rank-breaking is consistent, i.e. 
$\E_{\theta^*}[\nabla \Lrb(\theta^*)] = 0$. 
Throughout the proof of Theorem \ref{thm:main2}, 
this is the only place where the assumption on the proposed (consistent) rank-breaking is used. 
According to a companion theorem in \cite[Theorem 2]{APX14a}, 
it also follows that any rank-breaking that is not union of position-$p$ rank-breakings results in 
inconsistency, i.e. $\E_{\theta^*}[\nabla \Lrb(\theta^*)] \neq 0$. 
We claim that for each rank-breaking graph $G_{j,a}$, $\norm{\nabla\L_{G_{j,a}}(\theta^*)}_2^2 \leq (\lambda_{j,a})^2 (\kappa_j - p_{j,a})(\kappa_j- p_{j,a}+1)$. By Lemma \ref{lem:az_gen} which is a generalization of the vector version of the Azuma-Hoeffding inequality found in \cite[Theorem 1.8]{hayes2005large}, we have 
\begin{eqnarray*}
\P\big[\big\|\nabla\Lrb(\theta^*)\big\|_2 \geq \delta \big] \;\;\leq\;\; 2e^{3}\exp\Bigg(\frac{-\delta^2}{2\sum_{j=1}^n  \sum_{a=1}^{\ell_j}  \big(\lambda_{j,a}\big)^2 \big(\kappa_j - p_{j,a}\big)\big(\kappa_j- p_{j,a}+1\big)}\Bigg)\,,
\end{eqnarray*}
which implies the result.

\begin{lemma}\label{lem:az_gen}
Let $(X_1,X_2,\cdots, X_n)$ be real-valued martingale taking values in $\reals^d$ such that $X_0 = 0$ and for every $1 \leq i \leq n$, $\norm{X_i-X_{i-1}}_2 \leq c_i$, for some non-negative constant $c_i$. Then for every $\delta > 0$,
\begin{eqnarray}
\P[\norm{X_n}_2 \geq \delta] & \leq & 2e^{3}e^{-\frac{\delta^2}{2\sum_{i=1}^n c_i^2}}\,.
\end{eqnarray}  
\end{lemma}

It follows from the upper bound on $\norm{\nabla\L_{G_{j,a}}(\theta^*)}_2^2 \leq c_i^2$ with
$c_i^2=\lambda^2\big( (k_j-p_{j,a})^2 + (k_j-p_{j,a}) \big)$. 
In the expression \eqref{eq:rev1}, $\nabla\L_{G_{j,a}}(\theta^*)$ has one entry at $p_{j,a}$-th position that is compared to $(k_j-p_{j,a})$ other items 
and $(k_j-p_{j,a})$ entries that is compared only once, giving the bound 
\begin{eqnarray*}
	\norm{\nabla\L_{G_{j,a}}(\theta^*)}_2^2 &\leq& 
	\lambda_{j,a}^2(k_j-p_{j,a})^2 + \lambda_{j,a}^2 (k_j-p_{j,a}) \;.
\end{eqnarray*} 

\subsubsection{Proof of Lemma \ref{lem:consistency}}
Define event $E \equiv \{(i,\i) \in G_p \}$. Observe that
\begin{align*}
E = \Big\{\Big(\I_ {\{(\sigma^{-1}(i) = p\}} + \I_{\{\sigma^{-1}(\i)) = p\}} = 1\Big) \wedge  \Big(\sigma^{-1}(i),\sigma^{-1}(\i) \geq p \Big)\Big\} \;.
\end{align*}
Consider any set $\Omega \subset S\setminus\{i,\i\}$ such that $|\Omega| = p-1$. Let $M$ denote an event that items of the set $\Omega$ are ranked in top-$(p-1)$ positions in a particular order. It is easy to verify the following:
\begin{eqnarray*}
\P\Big[\sigma^{-1}(i) < \sigma^{-1}(\i) \Big| E, M\Big] &=& \frac{\P\Big[\big(\sigma^{-1}(i) < \sigma^{-1}(\i)\big),  E, M\Big]}{\P\Big[E, M\Big]}\\
&=& \frac{\P\Big[\big(\sigma^{-1}(i)= p\big), M\Big]}{\P\Big[\big(\sigma^{-1}(i)= p\big), M\Big] + \P\Big[\big(\sigma^{-1}(\i) =p \big), M\Big]} \\
&=& \frac{\exp(\theta^*_i)}{\exp(\theta^*_i) + \exp(\theta^*_{\i})} = \P\Big[\sigma^{-1}(i) < \sigma^{-1}(\i) \Big]\;.
\end{eqnarray*}  
Since $M$ is any particular ordering of the set $\Omega$ and $\Omega$ is any subset of $S\setminus\{i,\i\}$ such that $|\Omega| = p -1$, conditioned on event $E$ probabilities of all the possible events $M$ over all the possible choices of set $\Omega$ sum to $1$. 

\subsubsection{Proof of Lemma \ref{lem:az_gen}}
It follows exactly along the lines of proof of Theorem 1.8 in \cite{hayes2005large}.

\subsubsection{Proof of Lemma \ref{lem:hessian_positionl}}
The Hessian  $H(\theta)$ 
is given in  \eqref{eq:hessian}. For all  $j\in [n]$, define $M^{(j)} \in \cS^d$ as
\begin{eqnarray} \label{eq:posl_M_j_def}
M^{(j)} &\equiv&  \sum_{a=1}^{\ell_j} \lambda_{j,a}  \sum_{i<\i \in S_j} \I_{\big\{(i,\i)\; \in \; G_{j,a}\big\}} (e_i - e_{\i})(e_i - e_{\i})^\top,
\end{eqnarray}
and let $M \equiv \sum_{j=1}^n  M^{(j)}$. Observe that $M$ is positive semi-definite and the smallest eigenvalue of $M$ is zero with the corresponding eigenvector given by the all-ones vector. If $|\theta_i| \leq b$, for all $i \in [d]$, $\frac{\exp(\theta_i + \theta_{\i})}{[\exp(\theta_i) + \exp(\theta_{\i})]^2} \geq \frac{e^{2b}}{(1+ e^{2b})^2}$. Recall the definition of $H(\theta)$ from Equation \eqref{eq:hessian}. It follows that $-H(\theta) \succeq \frac{e^{2b}}{(1+ e^{2b})^2} M$ for $\theta \in \Omega_b$. Since, $-H(\theta)$ and $M$ are symmetric matrices, from Weyl's inequality we have, $\lambda_2(-H(\theta)) \geq \frac{e^{2b}}{(1+ e^{2b})^2} \lambda_2(M)$. 
Again from Weyl's inequality, it follows that 
\begin{eqnarray}
	\lambda_2(M) &\geq& \lambda_2(\E[M]) - \norm{M-\E[M]} \;, 
\end{eqnarray}
	where $\|\cdot\|$ denotes the spectral norm. 
	We will show in \eqref{eq:positionl_expec} that
	 $\lambda_2(\E[M]) \geq 2 \gamma e^{-6b} (\alpha/(d-1)) \sum_{j = 1}^n \tau_j\ell_j$, 
	 and in \eqref{eq:hess_posl_6} that 
	 $\norm{M-\E[M]}\leq 8e^{3b}\sqrt{\frac{\eta\delta\log d}{\beta \tau d}\sum_{j = 1}^n \tau_j \ell_j} $.

\begin{eqnarray} \label{eq:lambda2_M}
\lambda_2(M) &\geq& \frac{2e^{-6b} \alpha \gamma}{d-1}  \sum_{j = 1}^n \tau_j\ell_j - 8e^{3b}\sqrt{\frac{\eta\delta\log d}{\beta \tau d}\sum_{j = 1}^n \tau_j \ell_j} \;\geq\; \frac{e^{-6b} \alpha \gamma}{d-1}  \sum_{j = 1}^n \tau_j\ell_j \;, 
\end{eqnarray}
where the last inequality follows from the assumption that $\sum_{j = 1}^n \tau_j \ell_j \geq 2^{6}e^{18b} \frac{\eta\delta}{\alpha^2\beta \gamma^2\tau} d\log d$. 
This proves the desired claim. 

To prove the lower bound on $\lambda_2(\E[M])$, notice that 
\begin{eqnarray} \label{eq:posl_expec1a} 
\E[M] &=&  \sum_{j = 1}^n \sum_{a =1}^{\ell_j} \lambda_{j,a} \sum_{i<\i \in S_j}   \P\Big[(i,\i) \in G_{j,a} \Big| (i,\i \in S_j) \Big] (e_i - e_{\i})(e_i - e_{\i})^\top\;.
\end{eqnarray}
The following lemma provides a lower bound on $\P[(i,\i) \in G_{j,a} | (i,\i \in S_j)]$.

\begin{lemma} \label{lem:posl_lowerbound}
Consider a ranking $\sigma$ over a set $S \subseteq [d]$ such that $|S| = \kappa$. 
For any two items $i,\i \in S$, $\theta\in\Omega_b$, and  $1 \leq \ell \leq \kappa-1$,
\begin{align} \label{eq:posl_lowerbound_eq} 
\P_{\theta}\Big[\sigma^{-1}(i) = \ell, \sigma^{-1}(\i) > \ell \Big] \;\;\geq\;\; \frac{e^{-6b}(\kappa-\ell)}{\kappa(\kappa-1)} \bigg(1 - \frac{\ell}{\kappa}\bigg)^{\alpha_{i,i',\ell,\theta} -2} \;,
\end{align}
where the probability $\prob_\theta$ is with respect to the sampled ranking resulting from PL weights $\theta\in\Omega_b$, 
and $\alpha_{i,i',\ell,\theta}$ is defined as  
$1 \leq \alpha_{i,i',\ell,\theta} = \ceil{\widetilde{\alpha}_{i,i',\ell,\theta}}$, and $\widetilde{\alpha}_{i,i',\ell,\theta}$ is, 
	\begin{align} \label{eq:posl_alpha}
	\widetilde{\alpha}_{i,i',\ell,\theta} \;\;  \equiv \;\; 
	\max_{\ell'\in[\ell]} \max_{\substack{\Omega \subseteq S\setminus\{i,\i\} \\ : |\Omega| = \kappa-\ell'}} \Bigg\{\frac{\exp(\theta_i)+\exp(\theta_{\i})}{\big(\sum_{j\in \Omega} 	\exp(\theta_j)\big)/|\Omega|} \Bigg \}\;.
	\end{align} 
\end{lemma}

Note that we do not need $\max_{\ell' \in[\ell]}$ in the above equation as the expression achieves its maxima at $\ell' = \ell$, 
but we keep the definition to avoid any confusion.
In the worst case, $2e^{-2b} \leq \widetilde{\alpha}_{i,i',\ell,\theta} \leq 2e^{2b}$. 
Therefore, using definition of rank breaking graph $G_{j,a}$, and Equations \eqref{eq:posl_expec1a} and \eqref{eq:posl_lowerbound_eq} we have,
\begin{eqnarray}
\E[M] &\succeq&  \gamma e^{-6b} \sum_{j = 1}^n \sum_{a =1}^{\ell_j} \lambda_{j,a} \frac{2(\kappa_j-p_{j,a})}{\kappa_j(\kappa_j-1)} \sum_{i<\i \in S_j} (e_i - e_{\i})(e_i - e_{\i})^\top \nonumber\\
&\succeq& 2\gamma  e^{-6b} \sum_{j = 1}^n  \frac{1}{\kappa_j(\kappa_j-1)} \sum_{a =1}^{\ell_j} \lambda_{j,a}(\kappa_j-p_{j,a}) \sum_{i<\i \in S_j} (e_i - e_{\i})(e_i - e_{\i})^\top \nonumber\\
&=&  2\gamma e^{-6b} L, \label{eq:positionl_expec}
\end{eqnarray}
where we used $\gamma\leq (1-p_{j,\ell_j}/\kappa_j )^{\alpha_1-2}$ which follows for the definition in \eqref{eq:gamma_def}. \eqref{eq:positionl_expec} follows from the definition of Laplacian $L $, defined for the comparison graph $\H $ in Definition \ref{def:comparison_graph2}. 
Using $\lambda_2(L) = (\alpha /(d-1)) \sum_{j = 1}^n \tau_j\ell_j$ from \eqref{eq:lambda2_L2}, 
we get the desired bound 
 $\lambda_2(\E[M]) \geq 2 \gamma e^{-6b} (\alpha /(d-1)) \sum_{j = 1}^n \tau_j\ell_j$.

Next we need to upper bound $\|\sum_{j =1}^n\E[(M^{j})^2]\|$ to bound the  deviation of $M$ from its expectation. 
To this end, we prove an upper bound on $\P[\sigma_j^{-1}(i) = p_{j,a} \; | \;i \in S_j ]$ in the following lemma. 
\begin{lemma} \label{lem:posl_upperbound}
Under the hypotheses of Lemma \ref{lem:posl_lowerbound},
	\begin{align} \label{eq:posl_upperbound_eq}
	\P_{\theta}\Big[ \sigma^{-1}(i) = \ell \Big] \;\;\leq\;\;  \frac{e^{6b}}{\kappa} \bigg(1 - \frac{\ell}{\kappa+\alpha_{i,\ell,\theta}} \bigg)^{\alpha_{i,\ell,\theta} -1} 
	 \;\; \leq \;\; \frac{e^{6b}}{\kappa-\ell} \;,
	\end{align}
where 
$0 \leq \alpha_{i,\ell,\theta} = \floor{\widetilde{\alpha}_{i,\ell,\theta}}$, and $\widetilde{\alpha}_{i,\ell,\theta}$ is,
\begin{align} \label{eq:posl_upper1}
\widetilde{\alpha}_{i,\ell,\theta} \;\; \equiv \;\; \min_{\ell' \in [\ell]} \min_{\substack{\Omega \in S\setminus\{i\} \\ : |\Omega| = \kappa-\ell'+1}} \Bigg\{\frac{\exp(\theta_i)}{\big(\sum_{j\in \Omega} \exp(\theta_j)\big)/|\Omega|} \Bigg \}\;.
\end{align} 

In the worst case, $e^{-2b} \leq \widetilde{\alpha}_{i,\ell,\theta} \leq e^{2b}$. Note that $\alpha_{i,\ell,\theta} =0$ gives the worst upper bound.
\end{lemma}
Therefore using Equation \eqref{eq:posl_upperbound_eq}, for all $i \in [d]$, we have,
\begin{align}\label{eq:hess_posl_16}
\P\Big[\sigma_j^{-1}(i) \in \cP_j \Big] \leq \min \Bigg\{1, \frac{e^{6b}\ell_j}{\kappa_j - p_{j,\ell_j}} \Bigg\} \;\leq\;  \frac{e^{6b}\ell_j}{\max\{ \ell_j, \kappa_j - p_{j,\ell_j}\}} \leq  \frac{e^{6b}\eta \ell_j}{ \kappa_j}\,,
\end{align}
where we used $\eta$ defined in Equation \eqref{eq:eta_def}.
Define a diagonal matrix $D^{(j)} \in \cS^{d}$ and a matrix $A^{(j)} \in \cS^d$,  
\begin{eqnarray}
A^{(j)}_{i\i} &\equiv & \I_{\big\{i,\i \in S_j  \big\}} \,\sum_{a=1}^{\ell_j} \lambda_{j,a} \I_{\big\{(i,\i) \in G_{j,a}\big\}}\;,\; \text{for all} \;\; i,\i \in [d] \,, \label{eq:hess_posl_8}
\end{eqnarray}
and $D^{(j)}_{ii} = \sum_{i'\neq i} A^{(j)}_{ii'}$. 
Observe that $M^{(j)} = D^{(j)} - A^{(j)}$. 
For all $i \in [d]$, we have,
\begin{eqnarray}
D^{(j)}_{ii} &=& \I_{\big\{i \in S_j  \big\}} \sum_{\i = 1}^{\kappa_j} \I_{\big\{\sigma_j^{-1}(i) = \i \big\}} \sum_{a=1}^{\ell_j} \lambda_{j,a}  \deg_{G_{j,a}}(\sigma_j^{-1}(\i)) \nonumber\\
&\leq& \I_{\big\{i \in S_j  \big\}}\Bigg\{ \I_{\big\{\sigma^{-1}_j(i) \in \cP_j\big\}}\Bigg( \max_{a \in [\ell_j]} \Big\{\lambda_{j,a}(\kappa_j - p_{j,a})\Big\} + \sum_{a = 1}^{\ell_j} \lambda_{j,a} \Bigg) +  \I_{\big\{\sigma^{-1}_j(i) \notin \cP_j\big\}} \Bigg( \sum_{a = 1}^{\ell_j} \lambda_{j,a} \Bigg)\Bigg\} \nonumber\\
&=& \I_{\big\{i \in S_j  \big\}}\bigg\{ \I_{\big\{\sigma^{-1}_j(i) \in \cP_j\big\}}\delta_{j,1} \; + \; \I_{\big\{\sigma^{-1}_j(i) \notin \cP_j\big\}} \delta_{j,2}\bigg\}, \label{eq:hess_posl_9}
\end{eqnarray}
where the last equality follows from the definition of $\delta_{j,1}$ and $\delta_{j,2}$ in Equation \eqref{eq:delta12_def}. 
Note that $\max_{i \in [d]} \{D_{ii}\} = \delta_{j,1}$. Using \eqref{eq:hess_posl_16} and \eqref{eq:hess_posl_9}, we have,
\begin{eqnarray}\label{eq:hess_posl_14}
\E\Big[D^{(j)}_{ii}\Big] &\leq & \I_{\big\{i \in S_j  \big\}} \Bigg\{ \frac{e^{6b}\eta\ell_j}{\kappa_j} \bigg(\delta_{j,1} + \frac{\delta_{j,2}\kappa_j}{\eta\ell_j} \bigg) \Bigg\} \,.
\end{eqnarray}
Similarly we have,
\begin{eqnarray}\label{eq:hess_posl_10}
\E\Big[\big(D^{(j)}_{ii}\big)^2\Big] &\leq & \I_{\big\{i \in S_j  \big\}} \Bigg\{ \frac{ e^{6b}\eta\ell_j}{\kappa_j} \bigg( \delta_{j,1}^2 + \frac{\delta_{j,2}^2\kappa_j}{\eta\ell_j}   \bigg) \Bigg\} 
\end{eqnarray}
For all $i \in [d]$, we have,
\begin{eqnarray} \label{eq:hess_posl_12}
\E\Bigg[\sum_{\i = 1}^d \big(\big(A^{(j)}\big)^2\big)_{i\i} \Bigg] & \leq & \E\Bigg[ \bigg(\sum_{\i =1 }^d A^{(j)}_{i\i} \bigg) \max_{i \in [d]} \bigg\{ \sum_{\i =1 }^d A^{(j)}_{i\i}\bigg\} \Bigg] \nonumber\\
 &\leq & \E\bigg[ D^{(j)}_{ii} \delta_{j,1} \bigg] \nonumber\\
&\leq& \I_{\big\{i \in S_j  \big\}} \Bigg\{ \frac{e^{6b}\eta\ell_j}{\kappa_j} \bigg(\delta_{j,1}^2 + \frac{\delta_{j,1}\delta_{j,2}\kappa_j}{\eta\ell_j} \bigg)\Bigg\} \,.
 \end{eqnarray}
Using \eqref{eq:hess_posl_10} and \eqref{eq:hess_posl_12}, we have, for all $i \in [d]$,
\begin{eqnarray}
&&\sum_{\i = 1}^d \Big|\E\Big[\big(\big(M^{(j)}\big)^2\big)_{i\i}\Big]\Big| \nonumber\\
& = & \sum_{\i = 1}^d \Bigg|\E\Big[\big(\big(D^{(j)}\big)^2\big)_{i\i}\Big] - \E\Big[\big(D^{(j)} A^{(j)}\big)_{i\i}\Big]
- \E\Big[\big( A^{(j)} D^{(j)} \big)_{i\i}\Big]
+ \E\Big[\big(\big(A^{(j)}\big)^2\big)_{i\i}\Big] \Bigg| \nonumber\\
&\leq& 2\E\Big[\big(D^{(j)}_{ii}\big)^2\Big] + \sum_{\i = 1}^d \bigg( \E\Big[\delta_{j,1}\big(A^{(j)}\big)_{i\i}\Big] +  \E\Big[\big(\big(A^{(j)}\big)^2\big)_{i\i}\Big] \bigg) \nonumber\\
&\leq&  \I_{\big\{i \in S_j  \big\}} \Bigg\{ \frac{e^{6b}\eta\ell_j}{\kappa_j}\bigg( 4 \delta_{j,1}^2 + \frac{2\big(\delta_{j,1}\delta_{j,2} +\delta_{j,2}^2\big)\kappa_j}{\eta\ell_j} \bigg)  \Bigg\}\nonumber\\
&=&  \I_{\big\{i \in S_j  \big\}} \bigg\{ \frac{e^{6b}\delta\eta\ell_j}{\kappa_j}  \bigg\}\,, \label{eq:hess_posl_15}
\end{eqnarray}  
where the last equality follows from the definition of $\delta$, Equation \eqref{eq:delta_def}.

To bound $\|\sum_{j =1}^n \E[(M^{(j)})^2]\|$, we use the fact that for $J \in \reals^{d\times d}, \norm{J} \leq \max_{i \in [d]}\sum_{\i = 1}^d|J_{i\i}|$. Therefore, we have
\begin{eqnarray}
\Bigg\|\sum_{j =1}^n \E\Big[(M^{(j)})^2\Big]\Bigg\|  & \leq & e^{6b}\delta\eta \max_{i \in [d]}  \Bigg\{\sum_{j:i \in S_j} \frac{\ell_j}{\kappa_j}  \Bigg\} \nonumber\\
& = & \frac{e^{6b}\eta\delta}{\tau} D_{\max} \label{eq:hess_posl_4}\\
& = & \frac{e^{6b}\eta\delta}{\beta \tau  d}  \sum_{j = 1}^n \tau_{j}\ell_j\;, \label{eq:hess_posl_5}
\end{eqnarray}
where \eqref{eq:hess_posl_4} follows from the definition of $D _{\max}$ in Equation\eqref{eq:posl_Dlmax} and \eqref{eq:hess_posl_5} follows from the definition of $\beta$ in  \eqref{eq:lambda2_L2}. 
Observe that from Equation \eqref{eq:hess_posl_9}, $\norm{M^{(j)}} \leq 2\delta_{j,1} \leq 2\sqrt{\delta}$. Applying matrix Bernstein inequality, we have,
\begin{eqnarray*}
\mathbb{P}\Big[\big\|M - \E[M]\big\| \geq t\Big] \leq d \,\exp\Bigg(\frac{-t^2/2}{\frac{e^{6b}\eta\delta}{\beta \tau d} \sum_{j = 1}^n \tau_{j}\ell_j + 4\sqrt{\delta}t/3}\Bigg). 
\end{eqnarray*}
Therefore, with probability at least $1 - d^{-3}$, we have, 
\begin{align} \label{eq:hess_posl_6}
\big\|M - \E[M]\big\| \leq 4e^{3b}\sqrt{\frac{\eta\delta\log d}{\beta \tau d}\sum_{j = 1}^n \tau_j \ell_j} +\frac{64 \sqrt{\delta}\log d}{3} \leq 8e^{3b}\sqrt{\frac{\eta\delta\log d}{\beta \tau d}\sum_{j = 1}^n \tau_j \ell_j}  \;,
\end{align}
where the second inequality uses $\sum_{j = 1}^n \tau_j \ell_j \geq 2^{6} (\beta \tau /\eta)d\log d$ which follows from the assumption that $\sum_{j = 1}^n \tau_j \ell_j \geq 2^{6}e^{18b} \frac{\eta\delta}{\tau\gamma^2\alpha^2 \beta } d\log d$ and the fact that $\alpha, \beta \leq 1$, $\gamma \leq 1$, $\eta\geq 1$, and $\delta> \tau^2$.

\subsubsection{Proof of Lemma \ref{lem:posl_lowerbound}}
Since providing a lower bound on $\P_{\theta}\big[\sigma^{-1}(i) = \ell, \sigma^{-1}(\i) > \ell \big] $ 
for arbitrary $\theta$ is challenging, 
we construct a new set of parameters $\{\ltheta_j\}_{j\in[d]}$ from the original $\theta$. 
These new parameters are constructed such that  it is both easy to 
compute the probability and also provides a lower bound on the original distribution.
We denote the sum of the weights by  $W \equiv \sum_{j \in S} \exp(\theta_j)$. 
We define a new set of parameters $\{\ltheta_j\}_{j \in S}$: 
\begin{eqnarray}
	\ltheta_j &=& \left\{ \begin{array}{rl}  
		\log(\widetilde{\alpha}_{i,i',\ell,\theta}/2) &\; \text{for} \; j = i \text{ or }\i\;, \\
		0&\;\text{otherwise}\;. \end{array}\right. 
\end{eqnarray}
Similarly define $\widetilde{W} \equiv \sum_{j \in S} \exp(\ltheta_j) = \kappa-2+\widetilde{\alpha}_{i,i',\ell,\theta}$.  We have,
\begin{eqnarray}\label{eq:posl_3}
&& \P_{\theta}\Big[\sigma^{-1}(i) = \ell, \sigma^{-1}(\i) > \ell \Big] \nonumber\\
&=& \sum_{\substack{j_1 \in S \\ j_1 \neq i,\i}} \Bigg(\frac{\exp(\theta_{j_1})}{W}  \sum_{\substack{j_2 \in S \\ j_2 \neq i,\i,j_1}} \Bigg(\frac{\exp(\theta_{j_2})}{W-\exp(\theta_{j_1})}\cdots \nonumber \\
&& \Bigg( \sum_{\substack{j_{\ell-1} \in S \\ j_{\ell-1} \neq i,\i, \\ j_1,\cdots,j_{\ell-2}}} \frac{\exp(\theta_{j_{\ell-1}})}{W-\sum_{k=j_1}^{j_{\ell-2}}\exp(\theta_{k})}\frac{\exp(\theta_i)}{W-\sum_{k=j_1}^{j_{\ell-1}}\exp(\theta_{k})}\Bigg) \cdots\Bigg)\Bigg) \nonumber\\
&=&\frac{\exp(\theta_i)}{W} \,\sum_{\substack{j_1 \in S \\ j_1 \neq i,\i}} \Bigg( \frac{\exp(\theta_{j_1})}{W-\exp(\theta_{j_1})} \sum_{\substack{j_2 \in S \\ j_2 \neq i,\i,j_1}} \Bigg( \frac{\exp(\theta_{j_2})}{W-\exp(\theta_{j_1})-\exp(\theta_{j_2})}\cdots \nonumber\\
&& \sum_{\substack{j_{\ell-1} \in S \\ j_{\ell-1} \neq i,\i, \\ j_1,\cdots,j_{\ell-2}}} \Bigg( \frac{\exp(\theta_{j_{\ell-1}})}{W-\sum_{k=j_1}^{j_{\ell-1}}\exp(\theta_{k})} \Bigg)\cdots \Bigg)\Bigg) \nonumber\\ 
\end{eqnarray}
Consider the last summation term in the above equation and let
$\Omega_\ell = S\setminus\{i,i',j_1,\ldots,j_{\ell-2}\}$.
Observe that, $|\Omega_\ell| = \kappa-\ell$ and from equation \eqref{eq:posl_alpha}, $\frac{\exp(\theta_{i})+\exp(\theta_{\i})}{\sum_{j \in \Omega_\ell} \exp(\theta_j)} \leq \frac{\widetilde{\alpha}_{i,i',\ell,\theta}}{\kappa-\ell}$. We have, 
\begin{eqnarray}
&&\sum_{j_{\ell-1} \in \Omega_\ell} \frac{\exp(\theta_{j_{\ell-1}})}{W-\sum_{k=j_1}^{j_{\ell-1}}\exp(\theta_{k})}  \nonumber\\
&=& \sum_{j_{\ell-1} \in \Omega_\ell } \frac{\exp(\theta_{j_{\ell-1}})}{W-\sum_{k=j_1}^{j_{\ell-2}}\exp(\theta_{k}) - \exp(\theta_{j_{\ell-1}})} \nonumber\\
&\geq& \frac{\sum_{j_{\ell-1} \in \Omega_\ell}\exp(\theta_{j_{\ell-1}})}{W-\sum_{k=j_1}^{j_{\ell-2}}\exp(\theta_{k})-\big(\sum_{j_{\ell-1} \in \Omega_\ell}\exp(\theta_{j_{\ell-1}})\big)/|\Omega_\ell|} \label{eq:posl_jensen_ineq}\\
&=& \frac{\sum_{j_{\ell-1} \in \Omega_\ell}\exp(\theta_{j_{\ell-1}})}{\exp(\theta_i)+\exp(\theta_{\i}) +\sum_{j_{\ell-1} \in \Omega_\ell}\exp(\theta_{j_{\ell-1}})-\big(\sum_{j_{\ell-1} \in \Omega_\ell}\exp(\theta_{j_{\ell-1}})\big)/|\Omega_\ell|} \nonumber\\
&=&\Bigg({\frac{\exp(\theta_{i})+\exp(\theta_{\i})}{\sum_{j_{\ell-1} \in \Omega_\ell} \exp(\theta_{j_{\ell-1}})} + 1 - \frac{1}{\kappa-\ell}}\Bigg)^{-1} \nonumber\\
&\geq& \Bigg(\frac{\widetilde{\alpha}_1}{\kappa-\ell} + 1 - \frac{1}{\kappa-\ell}\Bigg)^{-1} \label{eq:posl_1}\\
&=& \frac{\kappa-\ell}{\widetilde{\alpha}_1 + \kappa-\ell-1} \nonumber\\
&=& \sum_{j_{\ell-1} \in \Omega_\ell } \frac{\exp(\ltheta_{j_{\ell-1}})}{\widetilde{W}-\sum_{k=j_1}^{j_{\ell-2}}\exp(\ltheta_{k}) - \exp(\ltheta_{j_{\ell-1}})} \label{eq:posl_2}\;,
\end{eqnarray}
where \eqref{eq:posl_jensen_ineq} follows from the Jensen's inequality and the fact that for any $c >0$, $0 < x < c$, $\frac{x}{c-x}$ is convex in $x$. Equation \eqref{eq:posl_1} follows from the definition of $\widetilde{\alpha}_{i,i',\ell,\theta}$, \eqref{eq:posl_alpha}, and the fact that $|\Omega_\ell| = \kappa-\ell$. Equation \eqref{eq:posl_2} uses the definition of $\{\ltheta_j\}_{j \in S}$.

Consider $\{\Omega_{\widetilde{\ell}}\}_{2 \leq \widetilde{\ell} \leq \ell - 1}$, $|\Omega_{\widetilde{\ell}}| = \kappa - \widetilde{\ell}$, corresponding to the subsequent summation terms in \eqref{eq:posl_3}.  Observe that $\frac{\exp(\theta_i)+\exp(\theta_{\i})}{\sum_{j \in \Omega_{\widetilde{\ell}}} \exp(\theta_j)} \leq \widetilde{\alpha}_{i,i',\ell,\theta}/|\Omega_{\widetilde{\ell}}|$. Therefore, each summation term  in equation  \eqref{eq:posl_3} can be lower bounded by the corresponding term where $\{\theta_j\}_{j \in S}$ is replaced by $\{\ltheta_j\}_{j \in S}$. Hence, we have
\begin{eqnarray} \label{eq:posl_4}
&&\P_{\theta}\Big[\sigma^{-1}(i) = \ell, \sigma^{-1}(\i) > \ell \Big] \nonumber\\
&\geq& \frac{\exp(\theta_i)}{W} \sum_{\substack{j_1 \in S \\ j_1 \neq i,\i}} \Bigg( \frac{\exp(\ltheta_{j_1})}{\widetilde{W}-\exp(\ltheta_{j_1})} \sum_{\substack{j_2 \in S \\ j_2 \neq i,\i,j_1}} \Bigg( \frac{\exp(\ltheta_{j_2})}{\widetilde{W}-\exp(\ltheta_{j_1})-\exp(\ltheta_{j_2})}\cdots \nonumber\\
&& \sum_{\substack{j_{\ell-1} \in S \\ j_{\ell-1} \neq i,\i, \\ j_1,\cdots,j_{\ell-2}}} \Bigg( \frac{\exp(\ltheta_{j_{\ell-1}})}{\widetilde{W}-\sum_{k=j_1}^{j_{\ell-1}}\exp(\ltheta_{k})} \Bigg)\Bigg)\Bigg) \nonumber\\
&\geq&  \frac{e^{-4b} \exp(\ltheta_i)}{\widetilde{W}}  \sum_{\substack{j_1 \in S \\ j_1 \neq i,\i}} \Bigg( \frac{\exp(\ltheta_{j_1})}{\widetilde{W}-\exp(\ltheta_{j_1})} \sum_{\substack{j_2 \in S \\ j_2 \neq i,\i,j_1}} \Bigg( \frac{\exp(\ltheta_{j_2})}{\widetilde{W}-\exp(\ltheta_{j_1})-\exp(\ltheta_{j_2})}\cdots \nonumber\\
&& \sum_{\substack{j_{\ell-1} \in S \\ j_{\ell-1} \neq i,\i, \\j_1,\cdots,j_{\ell-2}}} \Bigg( \frac{\exp(\ltheta_{j_{\ell-1}})}{\widetilde{W}-\sum_{k=j_1}^{j_{\ell-1}}\exp(\ltheta_{k})} \Bigg)\Bigg)\Bigg)\nonumber\\
&=& \big(e^{-4b}\big) \P_{\ltheta}\Big[\sigma^{-1}(i) = \ell, \sigma^{-1}(\i) > \ell \Big] \;.
\end{eqnarray}
The second inequality uses $\frac{\exp(\theta_i)}{W} \geq e^{-2b}/\kappa$ and $\frac{\exp(\ltheta_i)}{\widetilde{W}} \leq e^{2b}/\kappa$. 
Observe that $\exp(\ltheta_j) = 1$ for all $j \neq i,\i$ and $\exp(\ltheta_i) + \exp(\ltheta_{\i}) = \widetilde{\alpha}_{i,i',\ell,\theta} \leq \ceil{\widetilde{\alpha}_{i,i',\ell,\theta}} = \alpha_{i,i',\ell,\theta} \geq 1$. Therefore, we have
\begin{eqnarray}
&&\P_{\ltheta}\Big[\sigma^{-1}(i) = \ell, \sigma^{-1}(\i) > \ell \Big] \nonumber\\
&=& {\kappa-2 \choose \ell-1} \frac{(\widetilde{\alpha}_{i,i',\ell,\theta}/2)(\ell-1)!}{(\kappa-2+\widetilde{\alpha}_{i,i',\ell,\theta})(\kappa-2+\widetilde{\alpha}_{i,i',\ell,\theta} - 1)\cdots(\kappa-2+\widetilde{\alpha}_{i,i',\ell,\theta} - (\ell-1))} \nonumber\\
&\geq& \frac{(\kappa-2)!}{(\kappa-\ell-1)!} \frac{e^{-2b}}{(\kappa + \alpha_{i,i',\ell,\theta}-2)(\kappa + \alpha_{i,i',\ell,\theta} - 3)\cdots(\kappa + \alpha_{i,i',\ell,\theta} - (\ell+1))} \label{eq:posl_5} \\
& =& \frac{e^{-2b}(\kappa-\ell + \alpha_{i,i',\ell,\theta}-2)(\kappa -\ell +\alpha_{i,i',\ell,\theta}-3)\cdots (\kappa -\ell)}{(\kappa+\alpha_{i,i',\ell,\theta}-2)(\kappa+\alpha_{i,i',\ell,\theta}-3)\cdots(\kappa-1)} \nonumber\\
&=& \frac{e^{-2b}}{(\kappa-1)} \frac{(\kappa-\ell + \alpha_{i,i',\ell,\theta}-2)(\kappa -\ell +\alpha_{i,i',\ell,\theta}-3)\cdots (\kappa -\ell)}{(\kappa+\alpha_{i,i',\ell,\theta}-2)(\kappa+\alpha_{i,i',\ell,\theta}-3)\cdots(\kappa)}\nonumber\\
&\geq& \frac{e^{-2b}}{(\kappa-1)}  \bigg( 1- \frac{\ell}{\kappa}\bigg)^{\alpha_{i,i',\ell,\theta}-1}\nonumber\\
&=& \frac{e^{-2b}(\kappa-\ell)}{\kappa(\kappa-1)}  \bigg( 1- \frac{\ell}{\kappa}\bigg)^{\alpha_{i,i',\ell,\theta}-2}, \label{eq:posl_6}
\end{eqnarray}
where \eqref{eq:posl_5} follows from the fact that $\widetilde{\alpha}_{i,i',\ell,\theta} \geq 2e^{-2b}$. Claim \eqref{eq:posl_lowerbound_eq} follows by combining Equations \eqref{eq:posl_4} and \eqref{eq:posl_6}.

\subsubsection{Proof of Lemma \ref{lem:posl_upperbound}} 

Analogous to the proof of Lemma \ref{lem:posl_lowerbound}, 
we construct a new set of parameters $\{\ltheta_j\}_{j\in[d]}$ from the original $\theta$. 
We denote the sum of the weights by  $W \equiv \sum_{j \in S} \exp(\theta_j)$. 
We define a new set of parameters $\{\ltheta_j\}_{j \in S}$: 
\begin{eqnarray}
	\ltheta_j &=& \left\{ \begin{array}{rl}  
		\log(\widetilde{\alpha}_{i,\ell,\theta}) &\; \text{for} \; j = i \;, \\
		0&\;\text{otherwise}\;. \end{array}\right. 
\end{eqnarray}
Similarly define $\widetilde{W} \equiv \sum_{j \in S} \exp(\ltheta_j) = \kappa-1+\widetilde{\alpha}_{i,\ell,\theta}$.  We have, 
\begin{align}
& \P_{\theta}\Big[\sigma^{-1}(i) = \ell \Big] \nonumber\\
&= \sum_{\substack{j_1 \in S \\ j_1 \neq i}} \Bigg(\frac{\exp(\theta_{j_1})}{W}  \sum_{\substack{j_2 \in S \\ j_2 \neq i,j_1}} \Bigg(\frac{\exp(\theta_{j_2})}{W-\exp(\theta_{j_1})}\cdots \Bigg( \sum_{\substack{j_{\ell-1} \in S \\ j_{\ell-1} \neq i, \\ j_1,\cdots,j_{\ell-2}}} \frac{\exp(\theta_{j_{\ell-1}})}{W-\sum_{k=j_1}^{j_{\ell-2}}\exp(\theta_{k})}\frac{\exp(\theta_i)}{W-\sum_{k=j_1}^{j_{\ell-1}}\exp(\theta_{k})}\Bigg)\Bigg)\Bigg) \nonumber\\
&\leq  \sum_{\substack{j_1 \in S \\ j_1 \neq i}} \Bigg(\frac{\exp(\theta_{j_1})}{W}  \sum_{\substack{j_2 \in S \\ j_2 \neq i,j_1}} \Bigg(\frac{\exp(\theta_{j_2})}{W-\exp(\theta_{j_1})}\cdots \Bigg( \sum_{\substack{j_{\ell-1} \in S \\ j_{\ell-1} \neq i, \\ j_1,\cdots,j_{\ell-2}}} \frac{\exp(\theta_{j_{\ell-1}})}{W-\sum_{k=j_1}^{j_{\ell-2}}\exp(\theta_{k})}\Bigg)\Bigg)\Bigg) \frac{e^{2b}}{\kappa-\ell+1} \label{eq:posl_upper2}
\end{align}
Consider the last summation term in the equation \eqref{eq:posl_upper2}, and let $\Omega_\ell = S\setminus\{i,j_1,\ldots,j_{\ell-2}\}$,
such that $|\Omega_\ell| = \kappa-\ell+1$. Observe that from equation \eqref{eq:posl_upper1}, $\frac{\exp(\theta_i)}{\sum_{j \in \Omega_\ell} \exp(\theta_j)} \geq \frac{\lalpha_{i,\ell,\theta}}{\kappa-\ell+1}$. We have,
\begin{eqnarray}
\sum_{j_{\ell-1} \in \Omega_\ell} \frac{\exp(\theta_{j_{\ell-1}})}{W-\sum_{k=j_1}^{j_{\ell-2}}\exp(\theta_{k})} &=& \frac{\sum_{j_{\ell-1} \in \Omega_\ell} \exp(\theta_{j_{\ell-1}}) }{ \exp(\theta_i)+ \sum_{j_{\ell-1} \in \Omega_\ell} \exp(\theta_{j_{\ell-1}})} \nonumber\\
&\leq& \bigg( \frac{\lalpha_{i,\ell,\theta}}{\kappa-\ell+1}  + 1\bigg)^{-1} \nonumber\\
&=& \frac{\kappa-\ell+1}{\lalpha_{i,\ell,\theta} + \kappa - \ell +1} \nonumber\\
&=& \sum_{j_{\ell-1} \in \Omega_\ell} \frac{\exp(\ltheta_{j_{\ell-1}})}{\lW-\sum_{k=j_1}^{j_{\ell-2}}\exp(\ltheta_{k})}, \label{eq:posl_upper3}
\end{eqnarray}
where \eqref{eq:posl_upper3} follows from the definition of $\{\ltheta\}_{j \in S}$.

Consider $\{\Omega_{\widetilde{\ell}}\}_{2 \leq \widetilde{\ell} \leq \ell - 1}$, $|\Omega_{\widetilde{\ell}}| = \kappa - \widetilde{\ell} +1$, corresponding to the subsequent summation terms in \eqref{eq:posl_upper2}.  Observe that $\frac{\exp(\theta_i)}{\sum_{j \in \Omega_{\widetilde{\ell}}} \exp(\theta_j)} \geq \lalpha_{i,\ell,\theta}/|\Omega_{\widetilde{\ell}}|$. Therefore, each summation term  in equation  \eqref{eq:posl_3} can be lower bounded by the corresponding term where $\{\theta_j\}_{j \in S}$ is replaced by $\{\ltheta_j\}_{j \in S}$. Hence, we have
\begin{eqnarray} \label{eq:posl_upper4}
&&\P_{\theta}\Big[\sigma^{-1}(i) = \ell\Big] \nonumber\\
&\leq & \sum_{\substack{j_1 \in S \\ j_1 \neq i}} \Bigg(\frac{\exp(\ltheta_{j_1})}{\lW}  \sum_{\substack{j_2 \in S \\ j_2 \neq i,j_1}} \Bigg(\frac{\exp(\ltheta_{j_2})}{\lW-\exp(\ltheta_{j_1})}\cdots \Bigg( \sum_{\substack{j_{\ell-1} \in S \\ j_{\ell-1} \neq i, \\ j_1,\cdots,j_{\ell-2}}} \frac{\exp(\ltheta_{j_{\ell-1}})}{\lW-\sum_{k=j_1}^{j_{\ell-2}}\exp(\ltheta_{k})}\Bigg)\Bigg)\Bigg) \frac{e^{2b}}{\kappa-\ell+1} \nonumber\\
&\leq & e^{4b} \sum_{\substack{j_1 \in S \\ j_1 \neq i}} \Bigg(\frac{\exp(\ltheta_{j_1})}{\lW}  \sum_{\substack{j_2 \in S \\ j_2 \neq i,j_1}} \Bigg(\frac{\exp(\ltheta_{j_2})}{\lW-\exp(\ltheta_{j_1})}\cdots \nonumber\\
&& \Bigg( \sum_{\substack{j_{\ell-1} \in S \\ j_{\ell-1} \neq i, \\ j_1,\cdots,j_{\ell-2}}} \frac{\exp(\ltheta_{j_{\ell-1}})}{\lW-\sum_{k=j_1}^{j_{\ell-2}}\exp(\ltheta_{k})} \frac{\exp(\ltheta_i)}{\lW - \sum_{k = j_1}^{j_{\ell-1}}  \exp(\ltheta_k)} \Bigg)\Bigg)\Bigg) \nonumber\\
&\leq & e^{4b} \P_{\ltheta}\Big[\sigma^{-1}(i) = \ell\Big] \label{eq:posl_upper5}
\end{eqnarray}
The second inequality uses $\lalpha_2/(\kappa- \ell+\lalpha_{i,\ell,\theta}) \geq e^{-2b}/(\kappa - \ell +1)$. Observe that $\exp(\ltheta_j) = 1$ for all $j \neq i$ and $\exp(\ltheta_i) = \widetilde{\alpha}_{i,\ell,\theta} \geq \floor{\widetilde{\alpha}_{i,\ell,\theta}} = \alpha_{i,\ell,\theta} \geq 0$. Therefore, we have
\begin{eqnarray}
\P_{\ltheta}\Big[\sigma^{-1}(i) = \ell \Big] 
&=& {\kappa-1 \choose \ell-1} \frac{\lalpha_{i,\ell,\theta}(\ell-1)!}{(\kappa-1+\widetilde{\alpha}_{i,\ell,\theta})(\kappa-2+\widetilde{\alpha}_{i,\ell,\theta})\cdots(\kappa-\ell+\widetilde{\alpha}_{i,\ell,\theta}) } \nonumber\\
&\leq& \frac{(\kappa-1)!}{(\kappa-\ell)!} \frac{e^{2b}}{(\kappa -1 + \alpha_{i,\ell,\theta})(\kappa -2+ \alpha_{i,\ell,\theta} )\cdots(\kappa -\ell + \alpha_{i,\ell,\theta} )} \nonumber\\
&\leq& \frac{e^{2b}}{\kappa}  \bigg( 1- \frac{\ell}{\kappa+\alpha_{i,\ell,\theta}}\bigg)^{\alpha_{i,\ell,\theta}-1}, \label{eq:posl_upper6}
\end{eqnarray}
Note that equation \eqref{eq:posl_upper6} holds for all values of $\alpha_{i,\ell,\theta} \geq 0$. Claim \ref{eq:posl_upperbound_eq} follows by combining Equations \eqref{eq:posl_upper5} and \eqref{eq:posl_upper6}.

\subsection{Proof of Theorem \ref{thm:cramer_rao_position_p}}
\label{sec:proof_cramer_rao_position_p}
Let $H(\theta) \in \mathcal{S}^d$ be Hessian matrix  such that $H_{i\i}(\theta) = \frac{\partial^2\L(\theta)}{\partial\theta_i \partial \theta_{\i}}$. The Fisher information matrix is defined as $I(\theta)  = -\E_\theta[H(\theta)]$. 
Fix any unbiased estimator $\widehat{\theta}$ of $\theta \in \Omega_b$. Since, $\widehat{\theta} \in \mathcal{U}$, $\widehat{\theta} - \theta$ is orthogonal to $\vect{1}$. The Cram\'er-Rao lower bound then implies that ${\E[\norm{\widehat{\theta} - \theta^*}^2] \geq \sum_{i = 2}^d \frac{1}{\lambda_i(I(\theta))}}$. Taking the supremum over both sides gives
\begin{align*}
	\sup_{\theta}\E[\norm{\widehat{\theta} - \theta}^2]  \;\; \geq \;\;  \sup_{\theta} \sum_{i=2}^d \frac{1}{\lambda_i(I(\theta))} \geq \sum_{i = 2}^d \frac{1}	
	{\lambda_i(I(\vect{0}))}\;.
\end{align*}
The following lemma provides a lower bound on $\E_\theta[H(\vect{0})]$, where $\vect{0}$ indicates the all-zeros vector. 
\begin{lemma}\label{lem:cr_lem}
Under the hypotheses of Theorem \ref{thm:cramer_rao_position_p},
\begin{align}\label{eq:cr0}
\E_\theta[H(\vect{0})] \;\;\succeq\;\;  - \sum_{j=1}^n \frac{2p\log(\kappa_j)^2}{\kappa_j(\kappa_j-1)} \sum_{\i<i \in S_j}(e_i - e_{\i})(e_i - e_{\i})^{\top} \,.
\end{align}
\end{lemma}
Observe that $I(\vect{0})$ is positive semi-definite. Moreover, $\lambda_1(I(\vect{0}))$ is zero and the corresponding eigenvector is the all-ones vector. It follows that
\begin{eqnarray*}
I(0) &\preceq &   \sum_{j=1}^n \frac{2p\log(\kappa_j)^2}{\kappa_j(\kappa_j-1)}  \sum_{\i<i \in S_j}(e_i - e_{\i})(e_i - e_{\i})^{\top} \\
& \preceq &  2p\log(\kappa_{\max})^2 \underbrace{\sum_{j=1}^n \frac{1}{\kappa_j(\kappa_j-1)} \sum_{\i<i \in S_j}(e_i - e_{\i})(e_i - e_{\i})^{\top}}_{=L}\;,
\end{eqnarray*}
where $L$ is the Laplacian defined for the comparison graph $\H$, Definition \ref{def:comparison_graph1}, as 
$\ell_j = 1$ for all $j \in [n]$ in this setting. By Jensen's inequality, we have
\begin{align*}
\sum_{i = 2}^d \frac{1}{\lambda_i(L)} \geq \frac{(d-1)^2}{\sum_{i = 2}^d \lambda_i(L)} = \frac{(d-1)^2}{\Tr(L)} = \frac{(d-1)^2}{n}.
\end{align*}

\subsubsection{Proof of Lemma \ref{lem:cr_lem}}
Define $\L_j(\theta)$ for $j \in [n]$ such that $\L(\theta) = \sum_{j = 1}^n \L_j(\theta)$. Let $H^{(j)}(\theta) \in \mathcal{S}^d$ be the Hessian matrix such that $H^{(j)}_{i\i}(\theta) = \frac{\partial^2\L_j(\theta)}{\partial\theta_i \partial \theta_{\i}}$ for $i,\i \in S_j$. We prove that for all $j \in [n]$,  
\begin{align}\label{eq:cr01}
\E_\theta[H^{(j)}(\vect{0})] \;\; \succeq\;\;  - \frac{2p\log(\kappa_j)^2}{\kappa_j(\kappa_j-1)} \sum_{\i<i \in S_j}(e_i - e_{\i})(e_i - e_{\i})^{\top} \,.
\end{align}
In the following, we omit superscript/subscript $j$ for brevity. With a slight abuse of notation, 
we use $\I_{\{\Omega^{-1}(i) = a\}} = 1$ if item $i$ is ranked at the $a$-th position in all the orderings $\sigma \in \Omega$. Let $\P[\theta]$ be the likelihood of observing $\Omega^{-1}(p) = i^{(p)}$ and the set $\Lambda$ (the set of the items that are ranked before the $p$-th position). We have,     
\begin{align} \label{eq:cr1}
\P(\theta) = \sum_{\sigma \in \Omega} \Bigg(\frac{\exp\big(\sum_{m = 1}^{p} \theta_{\sigma(m)} \big)}{\prod_{a=1}^{p} \Big(\sum_{m'=a}^{\kappa} \exp\big(\theta_{\sigma(m')}\big) \Big)}\Bigg)\,.
\end{align}
For $i,\i \in S_j $, we have 
\begin{align}\label{eq:cr7}
H_{i\i}(\theta) = \frac{1}{\P(\theta)} \frac{\partial^2\P(\theta)}{\partial\theta_i \partial \theta_{\i}} - \frac{\nabla_i \P(\theta) \nabla_{\i} \P(\theta)}{\big(\P(\theta)\big)^2} 
\end{align}
We claim that at $\theta = \vect{0}$, 

\begin{eqnarray}
-H_{i\i}(\vect{0}) = \left\{ \begin{array}{rl}
C_1 & \;\; \text{if} \;  i = \i, \; \big\{\Omega^{-1}(i) \geq p \big\}  \label{eq:cr81} \\
C_2 + A_3^2 - C_3 & \;\; \text{if} \;  i = \i,\; \big\{\Omega^{-1}(i) < p \big\} \label{eq:cr82}\\
-B_1 &  \;\; \text{if} \;  i \neq \i, \; \big\{\Omega^{-1}(i) \geq p, \;\Omega^{-1}(\i) \geq p \big\}  \label{eq:cr83} \\
-B_2 &  \;\; \text{if} \; i \neq \i, \; \big\{\Omega^{-1}(i) \geq p, \; \Omega^{-1}(\i) < p \big\}  \label{eq:cr84} \\
-B_2 &  \;\; \text{if} \; i \neq \i, \; \big\{\Omega^{-1}(i) < p, \; \Omega^{-1}(\i) \geq p\big\}  \label{eq:cr85} \\
-(B_3 + B_4 - A_3^2) & \;\; \text{if} \; i \neq \i, \; \big\{\Omega^{-1}(i) < p,\; \Omega^{-1}(\i) < p\big\} \;. \label{eq:cr86}
\end{array}
\right.
\end{eqnarray}


where constants $A_3, B_1, B_2, B_3, B_4, C_1, C_2$ and $C_3$ are defined in Equations \eqref{eq:crA3}, \eqref{eq:crB1}, \eqref{eq:crB2}, \eqref{eq:crB3}, \eqref{eq:crB4}, \eqref{eq:crC1}, \eqref{eq:crC2} and \eqref{eq:crC3} respectively. 
From this computation of the Hessian, note that we have 
\begin{align} \label{eq:cr11}
H(\vect{0}) = \sum_{\i<i \in S}(e_i - e_{\i})(e_i - e_{\i})^{\top} \Big(H_{i\i}(\vect{0}) \Big) \;. 
\end{align}
which follows directly from the fact that the diagonal entries are summations of the off-diagonals, i.e. 
$C_1 = B_1(\kappa-p) + B_2(p-1)$ and $C_2 + A_3^2 - C_3 = B_2(\kappa-p+1) + (B_3 + B_4 - A_3^2)(p-2)$. 
The second equality follows from the fact that $C_2 = B_2(\kappa-p+1) + B_3(p-2)$ and $A_3^2(p-1) = B_4(p-2) + C_3$. 
Note that since $\theta = \vect{0}$, all items are exchangeable. Hence, 
$\E[H_{i\i}(\vect{0})] = \E[H_{ii}(\vect{0})]/(\kappa-1)$, and substituting this into \eqref{eq:cr11} and using Equations \eqref{eq:cr81}, we get 
\begin{eqnarray} 
&& \E\Big[ H(\vect{0})\Big] \nonumber\\
&=& -\frac{1}{\kappa-1}\bigg(\P\big[\Omega^{-1}(i) \geq p \big]C_1 + \P\big[\Omega^{-1}(i) < p \big](C_2 + A_3^2 - C_3)\bigg) \sum_{\i<i \in S}(e_i - e_{\i})(e_i - e_{\i})^{\top} \nonumber\\
&\succeq & - \frac{1}{\kappa(\kappa-1)} \sum_{\i<i \in S}(e_i - e_{\i})(e_i - e_{\i})^{\top}  \nonumber\\
&& \Bigg((\kappa-p+1)\log\bigg(\frac{\kappa}{\kappa-p}\bigg) + (p-1)\bigg(\log\bigg(\frac{\kappa}{\kappa-p+1}\bigg) + \log\bigg(\frac{\kappa}{\kappa-p+1}\bigg)^2 \bigg)\Bigg)  \nonumber\\\label{eq:cr12}\\
&\succeq & -\frac{2p\log(\kappa)^2}{\kappa(\kappa-1)} \sum_{\i<i \in S}(e_i - e_{\i})(e_i - e_{\i})^{\top} \;, \label{eq:cr13} 
\end{eqnarray}
where \eqref{eq:cr12} uses $\sum_{a = 1}^p \frac{1}{\kappa -a+1} \leq \log\big(\frac{\kappa}{\kappa-p}\big)$ and $C_3 \geq 0$. Equation \eqref{eq:cr13} follows from the fact that for any $x>0$, $\log(1+x) \leq x$. 
To prove \eqref{eq:cr81}, we have the first order partial derivative of $\P(\theta)$ given by
\begin{eqnarray} \label{eq:cr2}
\nabla_i \P(\theta) &=&  \I_{\{\Omega^{-1}(i) \leq p \}}\P(\theta)  -  \sum_{\sigma \in \Omega} \Bigg(\frac{\exp\big(\sum_{m = 1}^{p} \theta_{\sigma(m)} \big)}{\prod_{a=1}^{p} \Big(\sum_{m'=a}^{\kappa} \exp\big(\theta_{\sigma(m')}\big) \Big)} \Bigg( \sum_{a = 1}^p \frac{\I_{\{\sigma^{-1}(i) \geq a \}}\exp(\theta_i)}{\sum_{m'=a}^{\kappa} \exp\big(\theta_{\sigma(m')}\big)} \Bigg)   \Bigg) \,.
\end{eqnarray}
Define constants $A_1$, $A_2$ and $A_3$ such that
\begin{eqnarray} 
A_1 & \equiv & \P(\theta) \big|_{\{\theta = \vect{0}\}} = \frac{(p-1)!}{\kappa(\kappa-1)\cdots(\kappa-p+1)}, \label{eq:crA1}\\
A_2 & \equiv & \Bigg( \sum_{a = 1}^p \frac{\exp(\theta_i)}{\sum_{m'=a}^{\kappa} \exp\big(\theta_{\sigma(m')}\big)} \Bigg)\Bigg|_{\{\theta = \vect{0}\}} = \Bigg(\frac{1}{\kappa} + \frac{1}{\kappa-1}+\cdots + \frac{1}{\kappa-p+1}\Bigg), \label{eq:crA2} \\
A_3 & \equiv & \Bigg(\frac{(p-1)(p-2)!}{(p-1)!(\kappa)}  + \frac{(p-2)(p-2)!}{(p-1)!(\kappa-1)} + \cdots + \frac{(p-2)!}{(p-1)!(\kappa-p+2)} \Bigg) \,. \label{eq:crA3}
\end{eqnarray}
Observe that,  for all $i \in [d]$, 
\begin{align} \label{eq:cr4}
\nabla_i \P(\theta) \big|_{\{\theta = \vect{0}\}}  = A_1 \Big( \I_{\{\Omega_j^{-1}(i) = p\}}(1 - A_2) + \I_{\{\Omega_j^{-1}(i) < p\}}(1 - A_3) - \I_{\{\Omega_j^{-1}(i) > p\}}A_2 \Big) \;\; \,.
\end{align}
Further define constants $B_1$, $B_2$, $B_3$ and $B_4$ such that
\begin{eqnarray}
B_1 &\equiv & \Bigg(\frac{1}{\kappa^2} + \frac{1}{(\kappa-1)^2} + \cdots + \frac{1}{(\kappa - p+1)^2}\Bigg), \label{eq:crB1}\\
B_2 & \equiv & \Bigg(\frac{p-1}{(p-1)\kappa^2}  + \frac{p-2}{(p-1)(\kappa-1)^2} + \cdots + \frac{1}{(p-1)(\kappa-p+2)^2}  \Bigg), \label{eq:crB2} \\
B_3 & \equiv & \Bigg( \frac{(p-1)(p-2)(p-3)!}{(p-1)!\kappa^2} + \frac{(p-2)(p-3)(p-3)!}{(p-1)!(\kappa-1)^2} + \cdots + \frac{2(p-3)!}{(p-1)!(\kappa-p+3)^2}  \Bigg), \label{eq:crB3} \\
B_4 & \equiv & \frac{(p-3)!}{(p-1)!} \Bigg(\sum_{a,b \in [p-1], b \neq a} \bigg(\frac{1}{\kappa} + \frac{1}{\kappa -1} + \cdots + \frac{1}{\kappa - a+1} \bigg) \bigg(\frac{1}{\kappa} + \frac{1}{\kappa -1} + \cdots + \frac{1}{\kappa - b+1} \bigg)    \Bigg) \,. \label{eq:crB4}
\end{eqnarray}
Observe that,
\begin{eqnarray} \label{eq:cr6}
&&\frac{\partial^2\P(\theta)}{\partial\theta_i \partial \theta_{\i}}\bigg|_{\theta = \vect{0}} \nonumber\\
&=& \I_{\big\{\Omega^{-1}(i),\Omega^{-1}(\i) > p\big\}} A_1 \Big((-A_2)(-A_2) + B_1 \Big) \nonumber\\
&&+ \;\Big(\I_{\big\{\Omega^{-1}(i) > p, \Omega^{-1}(\i) = p\big\}} + \I_{\big\{\Omega^{-1}(i) = p, \Omega^{-1}(\i) > p\big\}} \Big) A_1 \Big((-A_2)(1-A_2) + B_1 \Big) \nonumber\\
&&+ \; \Big( \I_{\big\{\Omega^{-1}(i) = p, \Omega^{-1}(\i) < p\big\}} + \I_{\big\{\Omega^{-1}(i) < p, \Omega^{-1}(\i) = p\big\}}\Big) A_1 \Big((1-A_3) + (-A_2)(1-A_3) + B_2 \Big) \nonumber\\
&& + \;  \Big(\I_{\big\{\Omega^{-1}(i) > p, \Omega^{-1}(\i) < p\big\}} + \I_{\big\{\Omega^{-1}(i) < p, \Omega^{-1}(\i) > p\big\}} \Big) A_1 \Big((-A_2)(1-A_3) + B_2 \Big) \nonumber\\
&& + \;\I_{\big\{\Omega^{-1}(i) < p, \Omega^{-1}(\i) < p\big\}} A_1 \Big((1-A_3) + (-A_3) + B_4 + B_3 \Big)\,.
\end{eqnarray}
The claims in \eqref{eq:cr81} are easy to verify by combining Equations \eqref{eq:cr4} and \eqref{eq:cr6} with  \eqref{eq:cr7}. 
Also, define constants $C_1$, $C_2$ and $C_3$ such that,
\begin{eqnarray}
C_1 &\equiv &\Bigg( \frac{\kappa-1}{(\kappa)^2} + \frac{\kappa - 2}{(\kappa-1)^2} + \cdots + \frac{\kappa - p}{(\kappa-p+1)^2} \Bigg)\,,\label{eq:crC1}\\
C_2 & \equiv & \Bigg(\frac{(p-1)(p-2)!(\kappa-1)}{(p-1)!(\kappa)^2}  + \frac{(p-2)(p-2)!(\kappa-2)}{(p-1)!(\kappa-1)^2} + \cdots + \frac{(p-2)!(\kappa-p+1)}{(p-1)!(\kappa-p+2)^2} \Bigg) \,, \label{eq:crC2}\\
C_3 &\equiv & \frac{(p-2)!}{(p-1)!} \Bigg(\sum_{a,b \in [p-1], b=a} \bigg(\frac{1}{\kappa} + \frac{1}{\kappa-1} + \cdots + \frac{1}{\kappa-a+1}\bigg) \bigg(\frac{1}{\kappa} + \frac{1}{\kappa-1} + \cdots + \frac{1}{\kappa-b+1}\bigg) \Bigg) \,, \label{eq:crC3}
\end{eqnarray}
such that,
\begin{eqnarray} \label{eq:cr10}
\frac{\partial^2\P(\theta)}{\partial\theta_i^2}\bigg|_{\theta = \vect{0}} &=& \I_{\{ \Omega^{-1}(i) > p \}}A_1\Big((-A_2)(-A_2) - C_1 \Big) + \I_{\{ \Omega^{-1}(i) = p \}}A_1\Big((1-A_2) - A_2(1-A_2) - C_1 \Big) \nonumber\\
&& + \, \I_{\{ \Omega^{-1}(i) < p \}}A_1 \Big((1-A_3) - A_3 - C_2 + C_3 \Big)\,.
\end{eqnarray}
The claims \eqref{eq:cr81}   is easy to verify by combining Equations \eqref{eq:cr4} and \eqref{eq:cr10} with \eqref{eq:cr7}.

\subsection{Proof of Theorem \ref{thm:topl_upperbound}}
\label{sec:proof_topl_upperbound}

The proof is analogous to the proof of Theorem \ref{thm:main}. It differs primarily in the lower bound that is achieved for the second smallest eigenvalue of the Hessian matrix $H(\theta)$, \eqref{eq:hessian}. 
\begin{lemma}\label{lem:hessian_topl}
Under the hypotheses of Theorem \ref{thm:topl_upperbound}, if $\sum_{j = 1}^n \ell_j \geq (2^{12}e^{6b}/\beta\alpha^2) d\log d$ then with probability at least $ 1- d^{-3}$,
\begin{align} \label{eq:lambda2_bound_topl}
\lambda_2(-H(\theta)) \;\geq\; \frac{\alpha}{2(1+ e^{2b})^2} \frac{1}{d-1} \sum_{j = 1}^n \ell_j\,. 
\end{align}
\end{lemma}
Using Lemma \ref{lem:gradient_topl} that is derived for the general value of $\lambda_{j,a}$ and $p_{j,a}$, and by substituting $\lambda_{j,a} = 1/(\kappa_j-1)$ and $p_{j,a} = a$ for each $j \in [n]$, we get that with probability at least $1 - 2e^3d^{-3}$,
\begin{align} \label{eq:gradient_bound_topl}
\|\nabla\Lrb(\theta^*)\|_2  \;\leq\;  \sqrt{16\log d\sum_{j=1}^n \ell_j} \;. 
\end{align}
Theorem \ref{thm:topl_upperbound} follows from Equations \eqref{eq:gradient_bound_topl}, \eqref{eq:lambda2_bound_topl}  and \eqref{eq:thm_ml_3}.

\subsubsection{Proof of Lemma \ref{lem:hessian_topl}}
Define $M^{(j)} \in \cS^d$ as
\begin{eqnarray} \label{eq:M_j_def_topl}
M^{(j)} &=& \frac{1}{\kappa_j -1} \sum_{i<\i \in S_j} \sum_{a = 1}^{\ell_j} \I_{\{(i,\i)\; \in \; G_{j,a}\}}  (e_i - e_{\i})(e_i - e_{\i})^\top,
\end{eqnarray}
and let $M = \sum_{j=1}^n M^{(j)}$. Similar to the analysis carried out in the proof of Lemma \ref{lem:hessian_positionl}, we have $\lambda_2(-H(\theta)) \geq \frac{e^{2b}}{(1 + e^{2b})^2} \lambda_2(M)$, when $\lambda_{j,a} = 1/(\kappa_j-1)$ is substituted in the Hessian matrix $H(\theta)$, Equation \eqref{eq:hessian}. From Weyl's inequality we have that
\begin{align}\label{eq:topl3}
\lambda_2(M) \;\; \geq \lambda_2(\E[M]) - \norm{M - \E[M]}\,.
\end{align} 
We will show in \eqref{eq:topl_expec} that $\lambda_2(\E[M]) \geq e^{-2b}(\alpha/(d-1))\sum_{j = 1}^n \ell_j$ and in \eqref{eq:topl_error} that $\norm{M - \E[M]} \leq 32e^{b}\sqrt{\frac{\log d}{\beta d}\sum_{j=1}^n \ell_j}$. 
\begin{align} \label{eq:topl4}
\lambda_2(M) \; \geq \; \frac{\alpha e^{-2b}}{d-1} \sum_{j=1}^n \ell_j \;-\; 32e^{b} \sqrt{\frac{\log d}{\beta d}\sum_{j=1}^n \ell_j} \;\geq \;  \frac{\alpha e^{-2b}}{2(d-1)} \sum_{j=1}^n \ell_j\;, 
\end{align}
where the last inequality follows from the assumption that $\sum_{j=1}^n \ell_j \geq (2^{12}e^{6b}/\beta\alpha^2) d\log d$. This proves the desired claim.

To prove the lower bound on $\lambda_2(\E[M])$, notice that
\begin{eqnarray} \label{eq:topl5}
\E[M] &=&  \sum_{j = 1}^n \frac{1}{\kappa_j -1} \sum_{i<\i \in S_j}  \E\Bigg[ \sum_{a = 1}^{\ell_j} \I_{\{(i,\i) \in G_{j,a}\}} \Big| (i,\i \in S_j) \Bigg] (e_i - e_{\i})(e_i - e_{\i})^\top  \;.
\end{eqnarray} 
Using the fact that $p_{j,a} = a$ for each $j \in [n]$, and the definition of rank-breaking graph $G_{j,a}$, we have that
\begin{eqnarray}\label{eq:topl6}
\E\Bigg[ \sum_{a = 1}^{\ell_j} \I_{\{(i,\i) \in G_{j,a}\}} \Big| (i,\i \in S_j) \Bigg] &=& \P\Big[\I_{\{\sigma_j^{-1}(i) \leq \ell_j\}} + \I_{\{\sigma_j^{-1}(\i) \leq \ell_j\}} \geq 1 \Big| (i,\i \in S_j) \Big] \nonumber\\
& \geq &  \P\Big[(\sigma^{-1}(i) \leq \ell_j \Big| (i,\i \in S_j)\Big]\,.
\end{eqnarray}
The following lemma provides a lower bound on $\P[(\sigma^{-1}(i) \leq \ell_j | (i,\i \in S_j)]$. 
\begin{lemma}\label{lem:prob_toplbound}
Consider a ranking $\sigma$ over a set of items $S$ of size $\kappa$. For any item $i \in S$, 
\begin{align} \label{eq:prob_toplbound_eq}
\P[(\sigma^{-1}(i) \leq \ell] \geq e^{-2b}\frac{\ell}{\kappa}\;.
\end{align}
\end{lemma}
Therefore, using the fact that $(e_i - e_{\i})(e_i - e_{\i})^\top$ is positive semi-definite, and Equations \eqref{eq:topl5}, \eqref{eq:topl6} and \eqref{eq:prob_toplbound_eq} we have
\begin{eqnarray} \label{eq:topl_expec} 
\E[M] &\succeq& e^{-2b}  \sum_{j = 1}^n \frac{\ell_j}{\kappa_j(\kappa_j-1)} \sum_{i<\i \in S_j} (e_i - e_{\i})(e_i - e_{\i})^\top = e^{-2b} L,
\end{eqnarray}
where $L$ is the Laplacian defined for the comparison graph $\H$, Definition \ref{def:comparison_graph1}. Using $\lambda_2(L) = (\alpha/(d-1))\sum_{j = 1}^n \ell_j$ from \eqref{eq:lambda2_L1}, we get the desired bound $\lambda_2(\E[M]) \geq e^{-2b}(\alpha/(d-1))\sum_{j = 1}^n \ell_j$.

For top-$\ell_j$ rank breaking, $M^{(j)}$ is also given by
\begin{align} \label{eq:topl7}
M^{(j)} = \frac{1}{\kappa_j -1}\Big((\kappa_j - \ell_j)\diag(e_{\{I_j\}}) +\ell_j \diag(e_{\{S_j\}}) - e_{\{I_j\}}e_{\{S_j\}}^\top - e_{\{S_j\}}e_{\{I_j\}}^\top + e_{\{I_j\}}e_{\{I_j\}}^\top \Big),
\end{align}  
where $e_{\{S_j\}},e_{\{I_j\}} \in \reals^d$ are zero-one vectors, $e_{\{S_j\}}$ has support corresponding to the set of items $S_j$ and $e_{\{I_j\}}$ has support corresponding to the random top-$\ell_j$ items in the ranking $\sigma_j$. $I_j = \{\sigma_j(1), \sigma_j(2),\cdots, \sigma_j(\ell_j)\}$ for $j \in [n]$. $(M^{(j)})^2$ is given by
\begin{eqnarray*}
(M^{(j)})^2 &=& \frac{1}{(\kappa_j -1)^2}\Big((\kappa_j^2 - \ell_j^2)\diag(e_{\{I_j\}}) + {\ell_j}^2\diag(e_{\{S_j\}}) - \nonumber\\
&& \hspace{5em}(\kappa_j +\ell_j)(e_{\{I_j\}}e_{\{S_j\}}^\top + e_{\{S_j\}}e_{\{I_j\}}^\top -e_{\{I_j\}}e_{\{I_j\}}^\top ) + \ell_j e_{\{S_j\}}e_{\{S_j\}}^\top \Big).
\end{eqnarray*}
Note that $\P[i \in I_j| i \in S_j] \leq \ell_j e^{2b}/\kappa_j$ for all $i \in S_j$. Its proof is similar to the proof of Lemma \ref{lem:prob_toplbound}. Therefore, we have $\E[\diag(e_{\{I_j\}})] \preceq \ell_j e^{2b}/\kappa_j \diag(e_{\{\vect{1}\}})$. To bound $\|\sum_{j =1}^n\E[(M^{(j)})^2]\|$, we use the fact that for $J \in \reals^{d\times d}, \norm{J} \leq \max_{i \in [d]}\sum_{\i = 1}^d|J_{i\i}|$. Maximum of row sums of $\E[e_{\{I_j\}}e_{\{I_j\}}^\top]$ is upper bounded by $\max_{i \in [d]}\big\{\ell_j\P[i \in I_j| i \in S_j]\big\} \leq {\ell_j}^2 e^{2b}/\kappa_j$. Therefore using triangle inequality, we have, 
\begin{eqnarray}
&&\Bigg\|\sum_{j =1}^n\E\big[(M^{(j)})^2\big]\Bigg\|  \nonumber\\
& \leq &  \max_{i \in [d]} \Bigg\{\sum_{j:i \in S_j} \frac{1}{(\kappa_j-1)^2} \Bigg(\frac{(\kappa_j^2 - {\ell_j}^2)\ell_j e^{2b}}{\kappa_j} + {\ell_j}^2 + e^{2b}(\kappa_j+\ell_j)(2\ell_j + {\ell_j}^2/\kappa_j) + \ell_j\kappa_j \Bigg)\Bigg\} \nonumber\\
&\leq &  \max_{i \in [d]} \Bigg\{\sum_{j:i \in S_j} \frac{\ell_j e^{2b}}{\kappa_j}\Bigg(\frac{(\kappa_j^2 -{\ell_j}^2)}{(\kappa_j-1)^2} + \frac{\ell_j\kappa_j}{{(\kappa_j-1)^2}} + \frac{2(\kappa_j+\ell_j)\kappa_j}{(\kappa_j-1)^2} + \frac{(\kappa_j+\ell_j)\ell_j}{(\kappa_j-1)^2} + \frac{\kappa_j^2}{(\kappa_j-1)^2} \Bigg)\Bigg\} \nonumber\\
& \leq &  \max_{i \in [d]} \Bigg\{\sum_{j:i \in S_j} \frac{\ell_j e^{2b}}{\kappa_j}\Bigg(\frac{(\kappa_j^2 -1)}{(\kappa_j-1)^2} + \frac{\kappa_j(\kappa_j-1)}{{(\kappa_j-1)^2}} + \frac{4\kappa_j^2}{(\kappa_j-1)^2} + \frac{2\kappa_j(\kappa_j-1)}{(\kappa_j-1)^2} + \frac{\kappa_j^2}{(\kappa_j-1)^2} \Bigg)\Bigg\} \nonumber\\
& \leq &  \max_{i \in [d]} \Bigg\{\sum_{j:i \in S_j} \frac{\ell_j e^{2b}}{\kappa_j}\Bigg(3 + 2 + 16 + 4 + 4 \Bigg)\Bigg\} \label{eq:topl_grad1}\\
& \leq & 29 e^{2b}  \max_{i \in [d]}\bigg \{\sum_{j:i \in S_j}\frac{\ell_j}{\kappa_j}\bigg\} \nonumber\\
& = & 29  e^{2b}D_{\max} \label{eq:topl_grad2}\\
& = &  \frac{29  e^{2b}}{\beta d} \sum_{j = 1}^n \ell_j\;, \label{eq:topl_grad3}
\end{eqnarray}
where \eqref{eq:topl_grad1} uses the fact that $\kappa_j \geq 2$ and $1 \leq \ell_j \leq \kappa_j -1$ for all $j \in [n]$. \eqref{eq:topl_grad2} follows from the definition of $D_{\max}$, Definition \ref{def:comparison_graph1} and \eqref{eq:topl_grad3} follows from the Equation \eqref{eq:lambda2_L1beta}. Also, note that $\norm{M_j} \leq 2$ for all $j \in [n]$. Applying matrix Bernstien inequality, we have,
\begin{eqnarray*}
\mathbb{P}\Big[\norm{M - \E[M]} \geq t\Big] \leq d \,\exp\Bigg(\frac{-t^2/2}{\frac{29e^{2b}}{\beta d}\sum_{j=1}^n\ell_j + 4t/3}\Bigg). 
\end{eqnarray*}
Therefore, with probability at least $1 - d^{-3}$, we have, 
\begin{align} \label{eq:topl_error}
\norm{M - \E[M]} \leq 22e^{b}\sqrt{\frac{\log d}{\beta d} \sum_{j=1}^n \ell_j} +\frac{64 \log d}{3} \leq 32e^{b}\sqrt{\frac{\log d}{\beta d}\sum_{j=1}^n \ell_j} \;,
\end{align}
where the second inequality follows from the assumption that $\sum_{j = 1}^n \ell_j \geq 2^{12} d\log d$ and $\beta \leq 1$. 

\subsubsection{Proof of Lemma \ref{lem:prob_toplbound}}
Define $i_{\min} \equiv \arg \min_{i \in S} \theta_i$. We claim the following. 
For all $i \in S$ and any $1 \leq \ell \leq |S|-1$, 
\begin{align} \label{eq:prob_topl_eq}
\mathbb{P}[\sigma^{-1}(i) > \ell] \;\leq\; \mathbb{P}[\sigma^{-1}(i_{\min}) > \ell] \;\; \text{and} \;\; \mathbb{P}[\sigma^{-1}(i_{\min}) = \ell]  \; \geq \; \mathbb{P}[\sigma^{-1}(i_{\min}) = 1]\,.
\end{align}
Therefore $\mathbb{P}[\sigma^{-1}(i) \leq \ell] \;\geq\; \mathbb{P}[\sigma^{-1}(i_{\min}) \leq \ell]$. Using $\mathbb{P}[\sigma^{-1}(i_{\min}) = 1] > e^{-2b}/\kappa$, we get the desired bound $\mathbb{P}[\sigma^{-1}(i) \leq \ell] >  e^{-2b} \ell/\kappa$. 

To prove the claim \eqref{eq:prob_topl_eq}, let $\widehat{\sigma}_1^\ell$ denote a ranking of top-$\ell$ items of the set $S$ and $\P[\widehat{\sigma}_1^\ell]$ be the probability of observing $\widehat{\sigma}_1^\ell$. Let ${i \in (\widehat{\sigma}_1^\ell)^{-1}}$ denote that $i = (\widehat{\sigma}_1^\ell)^{-1}(j)$ for some $1 \leq j \leq \ell$. Let
\begin{align*}
\Omega_1 = \Big\{ \widehat{\sigma}_1^\ell : {i \notin (\widehat{\sigma}_1^\ell)^{-1}}, {i_{\min} \in (\widehat{\sigma}_1^\ell)^{-1}} \Big\} \;\; \text{and} \;\; \Omega_2 = \Big\{ \widehat{\sigma}_1^\ell : {i \in (\widehat{\sigma}_1^\ell)^{-1}}, {i_{\min} \notin (\widehat{\sigma}_1^\ell)^{-1}} \Big\}.
\end{align*}
We have $\mathbb{P}[\sigma^{-1}(i) > \ell]  - \mathbb{P}[\sigma^{-1}(i_{\min}) > \ell] =  \sum_{\widehat{\sigma}_1^\ell \in \Omega_1}\P[\widehat{\sigma}_1^\ell] -   \sum_{\widehat{\sigma}_1^\ell \in \Omega_2} \P[\widehat{\sigma}_1^\ell].$
Now, take any ranking $\widehat{\sigma}_1^\ell \in \Omega_1$ and construct another ranking $\widetilde{\sigma}_1^\ell$ from $\widehat{\sigma}_1^\ell$ by replacing $i_{\min}$ with $i$-th item. Observe that $ \P[\widehat{\sigma}_1^\ell] \leq \P[\widetilde{\sigma}_1^\ell]$ and $\widetilde{\sigma}_1^\ell \in \Omega_2$. Moreover, such a construction gives a bijective mapping between $\Omega_1$ and $\Omega_2$. Hence, the first claim is proved. For the second claim, let
\begin{align*}
\widehat{\Omega}_1 = \Big\{ \widehat{\sigma}_1^\ell : {(\widehat{\sigma}_1^\ell)^{-1}(i_{\min}) = 1} \Big\} \;\; \text{and} \;\; \widehat{\Omega}_2 = \Big\{ \widehat{\sigma}_1^\ell : {(\widehat{\sigma}_1^\ell)^{-1}(i_{\min}) = \ell} \Big\}.
\end{align*}
We have $\mathbb{P}[\sigma^{-1}(i_{\min}) =1]  - \mathbb{P}[\sigma^{-1}(i_{\min}) = \ell] =\sum_{\widehat{\sigma}_1^\ell \in \widehat{\Omega}_1}\P[\widehat{\sigma}_1^\ell]-   \sum_{\widehat{\sigma}_1^\ell \in \widehat{\Omega}_2} \P[\widehat{\sigma}_1^\ell].$
Now, take any ranking $\widehat{\sigma}_1^\ell \in \widehat{\Omega}_1$ and construct another ranking $\widetilde{\sigma}_1^\ell$ from $\widehat{\sigma}_1^\ell$ by swapping items at $1$st position and $\ell$-th position. Observe that  $ \P[\widehat{\sigma}_1^\ell] \leq \P[\widetilde{\sigma}_1^\ell]$ and $\widetilde{\sigma}_1^\ell \in \widehat{\Omega}_2$. Moreover, such a construction gives a bijective mapping between $\widehat{\Omega}_1$ and $\widehat{\Omega}_2$. Hence, the claim is proved.

\subsection{Proof of Theorem \ref{thm:cramer_rao_topl}}
\label{sec:proof_cramer_rao_topl}
The first order partial derivative of $\L(\theta)$, Equation \eqref{eq:PL_likelihood}, is given by
\begin{eqnarray*}
&&\nabla_i\L(\theta) \nonumber\\
&=&\sum_{j:i\in S_j} \sum_{m =1}^{\ell_j} \I_{\{\sigma_j^{-1}(i) \geq m \}} \Big[ \I_{\{\sigma_j(m) = i\}} - \frac{\exp(\theta_i)}{\exp(\theta_{\sigma_j(m)})+\exp(\theta_{\sigma_j(m+1)})+ \cdots + \exp(\theta_{\sigma_j(\kappa_j)})} \Big], \; \forall i \in [d]
\end{eqnarray*}
and the Hessian matrix $H(\theta) \in \mathcal{S}^d$ with $H_{i\i}(\theta) = \frac{\partial^2\L(\theta)}{\partial\theta_i \partial \theta_{\i}}$ is given by
\begin{align}
H(\theta) = - \sum_{j = 1}^n \sum_{i<\i \in S_j} (e_i - e_{\i})(e_i - e_{\i})^\top \sum_{m = 1}^{\ell_j} \frac{\exp(\theta_i+ \theta_{\i})\I_{\{\sigma_j^{-1}(i),\sigma_j^{-1}(\i) \geq m\}}}{[\exp(\theta_{\sigma_j(m)})+\exp(\theta_{\sigma_j(m+1)})+ \cdots + \exp(\theta_{\sigma_j(\kappa_j)})]^2}.
\end{align}
It follows from the definition that $-H(\theta)$ is positive semi-definite for any $\theta \in \reals^n$. 

The Fisher information matrix is defined as $I(\theta)  = -\E_\theta[H(\theta)]$ and given by
\begin{align*}
I(\theta) = \sum_{j = 1}^n \sum_{i<\i \in S_j} (e_i - e_{\i})(e_i - e_{\i})^\top  \sum_{m = 1}^{\ell_j} \E\Bigg[ \frac{\I_{\{\sigma_j^{-1}(i),\sigma_j^{-1}(\i) \geq m\}}}{[\exp(\theta_{\sigma_j(m)})+ \cdots + \exp(\theta_{\sigma_j(\kappa_j)})]^2}\Bigg]\exp(\theta_i+ \theta_{\i}).
\end{align*}
Since $-H(\theta)$ is positive semi-definite, it follows that $I(\theta)$ is positive semi-definite. Moreover, $\lambda_1(I(\theta))$ is zero and the corresponding eigenvector is the all-ones vector. Fix any unbiased estimator $\widehat{\theta}$ of $\theta \in \Omega_b$. Since, $\widehat{\theta} \in \mathcal{U}$, $\widehat{\theta} - \theta$ is orthogonal to $\vect{1}$. The Cram\'er-Rao lower bound then implies that ${\E[\norm{\widehat{\theta} - \theta^*}^2] \geq \sum_{i = 2}^d \frac{1}{\lambda_i(I(\theta))}}$. Taking the supremum over both sides gives
\begin{align*}
\sup_{\theta}\E[\norm{\widehat{\theta} - \theta}^2] \geq  \sup_{\theta} \sum_{i=2}^d \frac{1}{\lambda_i(I(\theta))} \geq \sum_{i = 2}^d \frac{1}{\lambda_i(I(\vect{0}))}\;.
\end{align*}
If $\theta$ equals the all-zero vector, then
\begin{align*}
\P_\theta[\sigma_j^{-1}(i),\sigma_j^{-1}(\i) \geq m] = \frac{{\kappa_j-m+1 \choose 2}}{{\kappa_j \choose 2}}  = \frac{(\kappa_j - m +1)(\kappa_j - m)}{\kappa_j(\kappa_j - 1)}.
\end{align*}
It follows from the definition that 
\begin{eqnarray*}
I(0) &=& \sum_{j = 1}^n \sum_{i<\i \in S_j} (e_i - e_{\i})(e_i - e_{\i})^\top \sum_{m = 1}^{\ell_j} \frac{(\kappa_j - m)}{\kappa_j(\kappa_j - 1)(\kappa_j - m +1)} \\
&\preceq& \ell\Big(1 - \frac{1}{\ell_j}\sum_{m= 1}^{\ell_j} \frac{1}{\kappa_{\max} - m +1}\Big) \underbrace{\sum_{j = 1}^n  \frac{1}{\kappa_j(\kappa_j - 1)} \sum_{i<\i \in S_j} (e_i - e_{\i})(e_i - e_{\i})^\top}_{=L}\;,
\end{eqnarray*}
where $L$ is the Laplacian defined for the comparison graph $\H$, Definition \ref{def:comparison_graph1}. By Jensen's inequality, we have
\begin{align*}
\sum_{i = 2}^d \frac{1}{\lambda_i(L)} \geq \frac{(d-1)^2}{\sum_{i = 2}^d \lambda_i(L)} = \frac{(d-1)^2}{\Tr(L)} = \frac{(d-1)^2}{n}.
\end{align*}

\subsection{Proof of Theorem \ref{thm:bottoml_upperbound}}

We prove a slightly more general result that implies the desired theorem. 
For $\ell\geq 4$, we can choose $\beta_1=1/2$. Then, the condition that 
$\gamma_{\beta_1}\leq1$ implies $\ld\leq (\ell/2+1)(d-2)/(\kappa-2)$, which implies $\ld \leq \ell d / (2 \kappa)$. With the choice of 
$\ld  = \ell d / (2 \kappa) $, this implies Theorem \ref{thm:bottoml_upperbound}.
\begin{theorem} 
	\label{thm:bottoml_upperbound_general}
	Under the bottom-$\ell$ separators scenario and the  PL model, $n$ partial orderings are sampled over $d$ items parametrized by $\theta^* \in \Omega_b$. 
	For any $\beta_1$ with $ 0 \leq \beta_1 \leq \frac{\ell-2}{\ell}$, define 
	\begin{align} \label{eq:bottoml_2_genreal}
		\gamma_{\beta_1}  \;\;  \equiv \;\; \frac{\ld(\kappa-2)}{(\floor{\ell\beta_1}+1)(d-2)}, \;\; 
	\end{align}
	and for $\gamma_{\beta_1}\leq1$, 
	\begin{eqnarray}
		\chi_{\beta_1} & \equiv&  \big(1-\floor{\ell\beta_1}/\ell\big)^2\Bigg(1 - \exp\bigg(-\frac{(\floor{\ell\beta_1}+1)^2(1-\gamma_{\beta_1})^2}{2(\kappa-2)}\bigg)\Bigg) \;.
	\end{eqnarray}
	If 
	\begin{align} \label{eq:bottoml_1_general}
	n\ell  \;\;  \geq \;\; \bigg(\frac{2^{12}e^{8b}}{\chi_{\beta_1}^2}\frac{d^2}{{\ld}^2}\frac{\kappa}{\ell}\bigg) d\log d\;, \;\;  
	\end{align}
	then the {\em rank-breaking} estimator in \eqref{eq:theta_ml_bl} achieves
	\begin{align} \label{eq:bottoml_3}
	\frac{1}{\sqrt{\ld}}\big\|\widehat{\ltheta} - \ltheta^*\big\|_2 \; \leq  \; \frac{32\sqrt{2}(1+ e^{4b})^2}{\chi_{\beta_1}}\frac{d^{3/2}}{{\ld}^{3/2}}\sqrt{\frac{d\log d}{n\ell} }\;,
	\end{align}
	with probability at least $1 - 3e^3d^{-3}$.
\end{theorem}

Proof is very similar to the proof of Theorem \ref{thm:main}. It mainly differs in the lower bound that is achieved for the second smallest eigenvalue of the Hessian matrix $H(\ltheta)$ of $\Lrb(\ltheta)$, Equation \eqref{eq:likelihood_bl_0}. Equation \eqref{eq:likelihood_bl_0} can be rewritten as
\begin{align}\label{eq:likelihhod_bl}
\Lrb(\ltheta) = \sum_{j=1}^n \sum_{a = 1}^{\ell} \sum_{\substack{i <\i \in S_j \\ : i,\i \in [\ld]}}  \I_ {\big\{(i,\i) \in G_{j,a}\big\}} \lambda_{j,a} \Big(\ltheta_i\I_{\big\{\sigma_j^{-1}(i) < \sigma_j^{-1}(\i)\big\}} + \ltheta_{\i}\I_{\big\{\sigma_j^{-1}(i) > \sigma_j^{-1}(\i)\big\}} - \log \Big(e^{\ltheta_i} + e^{\ltheta_{\i}}\Big) \Big)\;,
\end{align} 
where $(i,\i) \in G_{j,a}$ implies either $(i,\i)$ or $(\i,i)$ belong to $E_{j,a}$.
The Hessian matrix $H(\ltheta) \in \cS^{\ld}$ with $H_{i\i}(\ltheta) = \frac{\partial^2 \Lrb(\ltheta)}{\partial\ltheta_i \partial\ltheta_{\i}}$ is given by
\begin{align} \label{eq:limited_hessian}
H(\ltheta) = -\sum_{j=1}^n \sum_{a = 1}^{\ell} \sum_{\substack{i<\i \in S_j :\\  i,\i \in [\ld]}} \I_{\big\{(i,\i) \in G_{j,a}\big\}}\Bigg( (\le_i - \le_{\i})(\le_i - \le_{\i})^\top \frac{\exp(\ltheta_i + \ltheta_{\i})}{[\exp(\ltheta_i) + \exp(\ltheta_{\i})]^2}\Bigg).
\end{align} 

The following lemma gives a lower bound for $\lambda_2(-H(\ltheta))$.
\begin{lemma} \label{lem:hessian_bottoml}
Under the hypothesis of Theorem \ref{thm:bottoml_upperbound_general}, with probability at least $1 - d^{-3}$,
\begin{align}\label{eq:lambda2_bound_bottoml}
\lambda_2(-H(\ltheta)) \geq \frac{\chi_{\beta_1}}{8(1+ e^{4b})^2} \frac{n\ld\ell^2}{d^2}\;.
\end{align}
\end{lemma}

Observe that although $\ltheta^* \in \reals^{\ld}$, Lemma \ref{lem:gradient_topl} can be directly applied to upper bound $\norm{\nabla\Lrb(\ltheta^*)}_2$. It might be possible to tighten the upper bound, 
given that $\ld \leq d$. However, for $\ell \ll \kappa$, for the smallest preference score item, $i_{\min} \equiv \arg \min_{i \in [d]} \ltheta^*_i$, the upper bound $\P[\sigma^{-1}(i_{\min}) > \kappa-\ell] \leq 1$ is tight upto constant factor 
(Lemma \ref{lem:posl_upperbound}). 
Substituting $\lambda_{j,a} = 1$  and $p_{j,a} = \kappa - \ell +a$ for each $j \in [n]$, $a \in [\ell]$, in Lemma \ref{lem:gradient_topl}, we have that with probability at least $1 - 2e^3 d^{-3}$,
\begin{align}\label{eq:gradient_bound_bottoml}
\norm{\nabla\Lrb(\ltheta^*)}_2  \;\; \leq \;\; (\ell-1)\sqrt{8n\ell\log d}.
\end{align}
Theorem \ref{thm:bottoml_upperbound_general} follows from Equations \eqref{eq:thm_ml_3}, \eqref{eq:lambda2_bound_bottoml} and \eqref{eq:gradient_bound_bottoml}.

\subsubsection{Proof of Lemma \ref{lem:hessian_bottoml}}
Define $\lM^{(j)} \in \cS^{\ld}$,
\begin{align}\label{eq:limited_M_j_def}
\lM^{(j)} = \sum_{\substack{i<\i \in S_j : i,\i \in [\ld]}} \sum_{a = 1}^\ell \I_{\{(i,\i) \in G_{j,a}\}} (\le_i - \le_{\i})(\le_i - \le_{\i})^\top,
\end{align}
and let $\lM = \sum_{j = 1}^n \lM^{(j)}$. Similar to the analysis in Lemma~\ref{lem:hessian_positionl}, we have $\lambda_2(-H(\ltheta)) \geq \frac{e^{4b}}{(1+ e^{4b})^2} \lambda_2(\lM)$. Note that we have $e^{4b}$ instead of $e^{2b}$ as $\ltheta \in \lOmega_{2b}$. We will show a lower bound on $\lambda_2(\E[\lM])$ in \eqref{eq:lambda2_bottoml_expec} and an upper bound on $\norm{\lM - \E[\lM]}$ in \eqref{eq:bottoml_error}. Therefore using $\lambda_2(\lM) \geq \lambda_2(\E[\lM]) - \norm{\lM - \E[\lM]}$, 
\begin{align} \label{eq:bottoml_lambda2_M}
\lambda_2(\lM)  \; \geq\;  \frac{e^{-4b}}{4}\underbrace{(1-\beta_1)^2\Bigg(1 - \exp\bigg(-\frac{(\floor{\ell\beta_1}+1)^2(1-\gamma_{\beta_1})^2}{2(\kappa-2)}\bigg)\Bigg)}_{ \equiv \chi_{\beta_1}}\frac{n\ld\ell^2}{d^2} - 8\ell\sqrt{\frac{n\kappa\log d}{d}} \;.
\end{align}
The desired claim follows from the assumption that $n\ell \geq \big( \frac{2^{12}e^{8b}}{\chi_{\beta_1}^2}\frac{d^2}{{\ld}^2}\frac{\kappa}{\ell} \big) d\log d$, where $\chi_{\beta_1}$ is defined in \eqref{eq:bottoml_1_general}.
To prove the lower bound on $\lambda_2(\E[\lM])$, notice that
\begin{eqnarray} \label{eq:bottoml_hess2}  
\E\big[\lM\big] &=&  \sum_{j = 1}^n \sum_{i<\i \in [\ld]}   \E\Bigg[ \sum_{a = 1}^{\ell} \I_{\big\{(i,\i) \in G_{j,a}\big\}} \Big| (i,\i \in S_j) \Bigg] \P\Big[i,\i \in S_j\Big] (\le_i - \le_{\i})(\le_i - \le_{\i})^\top \;. 
\end{eqnarray}
Since the sets $S_j$ are chosen uniformly at random, $\P[i,\i \in S_j] = \kappa(\kappa -1)/d(d-1)$. Using the fact that $p_{j,a} = \kappa - \ell +a$ for each $j \in [n]$, and the definition of rank breaking graph $G_{j,a}$, we have that
\begin{eqnarray}\label{eq:bottoml_hess3}
\E\Bigg[ \sum_{a = 1}^{\ell} \I_{\big\{(i,\i) \in G_{j,a}\big\}} \Big| (i,\i \in S_j) \Bigg] = \P\Big[\big(\sigma_j^{-1}(i), \sigma_j^{-1}(\i) > \kappa - \ell\big) \Big| (i,\i \in S_j) \Big]\;.
\end{eqnarray}
The following lemma provides a lower bound on $\P[(\sigma_j^{-1}(i),\sigma_j^{-1}(\i)) > \kappa - \ell | (i,\i \in S_j)]$. 
\begin{lemma}\label{lem:prob_bottomlbound}
Under the hypotheses of Theorem \ref{thm:bottoml_upperbound_general}, for any two items $i,\i \in [\ld]$, the following holds:
\begin{align} 
	\label{eq:prob_bottomlbound_eq}
	\P\Big[\sigma^{-1}(i),\sigma^{-1}(\i) > \kappa - \ell \;\Big|\; i,\i \in S\Big] \;\geq\;  
	\frac{e^{-4b}(1-\beta_1)^2(1 - \exp({-\eta_{\beta_1}(1-\gamma_{\beta_1})^2}))}{2}\frac{\ell^2}{\kappa^2}\;,
\end{align}
where $\gamma_{\beta_1} \equiv \ld (\kappa-2)/(\lfloor \ell\beta_1 \rfloor+1 )(d-2) $ and 
$\eta_{\beta_1} \equiv (\lfloor \ell \beta_1 \rfloor +1)^2/2(\kappa-2)$. 
\end{lemma}
Therefore, using Equations \eqref{eq:bottoml_hess2}, \eqref{eq:bottoml_hess3} and \eqref{eq:prob_bottomlbound_eq} we have,
\begin{eqnarray}
	\label{eq:bottoml_expec}
	\E\big[\lM\big] &\succeq&  \frac{e^{-4b}(1-\beta_1)^2(1 - \exp({-\eta_{\beta_1}(1-\gamma_{\beta_1})^2}))}{2}  \frac{\ell^2}{\kappa^2}  \frac{\kappa(\kappa-1)}{d(d-1)} \sum_{j = 1}^n \sum_{i<\i \in [\ld]} (\le_i - \le_{\i})(\le_i - \le_{\i})^\top\;.
\end{eqnarray}
Define $\lL = \sum_{j = 1}^n \sum_{i<\i \in [\ld]} (\le_i - \le_{\i})(\le_i - \le_{\i})^\top$. We have, $\lambda_1(\lL) = 0$ and $\lambda_2(\lL)=\lambda_3(\lL)=\cdots=\lambda_{\ld}(\lL)$. Therefore, using $\lambda_2(\lL) = \Tr(\lL)/(\ld-1) = n\ld$. Using the fact that $\E[\lM]$ and $\lL$ are symmetric matrices, we have, 
\begin{align} \label{eq:lambda2_bottoml_expec} 
\lambda_2(\E\big[\lM\big]) \geq \frac{e^{-4b}(1-\beta_1)^2(1 - \exp({-\eta_{\beta_1}(1-\gamma_{\beta_1})^2}))}{4}  \frac{n\ld\ell^2}{d^2}. 
\end{align}

To get an upper bound on $\norm{\lM - \E[\lM]}$, notice that $\lM^{(j)}$ is also given by,
\begin{align} \label{eq:bottoml_hess4}
\lM^{(j)} \;\; =\;\; \ell\, \diag(\le_{\{I_j\}}) - \le_{\{I_j\}}\le_{\{I_j\}}^\top\;,
\end{align}
where $\le_{\{I_j\}} \in \reals^{\ld}$ is a zero-one vector, with support corresponding to the bottom-$\ell$ subset of items in the ranking $\sigma_j$. $I_j = \{\sigma_j(\kappa-\ell+1),\cdots, \sigma_j(\kappa)\}$ for $j \in [n]$. $(\lM^{(j)})^2$ is given by
\begin{align} \label{eq:bottoml_hess5}
(\lM^{(j)})^2  \;\; =\;\;  \ell^2 \,\diag(\le_{\{I_j\}}) - \ell\, \le_{\{I_j\}}\le_{\{I_j\}}^\top\;.
\end{align} 
Using the fact that sets $\{S_j\}_{j \in [n]}$ are chosen uniformly at random and $\P[i \in \I_j | i \in S_j] \leq 1$, we have $\E[\diag(\le_{\{I_j\}})] \preceq  (\kappa/d) \diag(\le_{\{\vect{1}\}})$. Maximum of row sums of $\E\big[\le_{\{I_j\}}\le_{\{I_j\}}^\top\big]$ is upper bounded by $\ell\kappa/d$. Therefore, from triangle inequality we have $\norm{\sum_{j=1}^n \E[(\lM^{(j)})^2]} \leq 2n\ell^2\kappa/d$. Also, note that $\norm{\lM^{(j)}} \leq 2\ell$ for all $j \in [n]$. Applying matrix Bernstien inequality, we have that
\begin{eqnarray} \label{eq:bottoml_hess6}
\mathbb{P}\Big[\norm{\lM - \E[\lM]} \geq t\Big] \leq d\,\exp\Big(\frac{-t^2/2}{2n\ell^2\kappa/d + 4\ell t/3}\Big). 
\end{eqnarray}
Therefore, with probability at least $1 - d^{-3}$, we have, 
\begin{align} \label{eq:bottoml_error}
\norm{\lM - \E[\lM]} \leq 4\ell\sqrt{\frac{2n\kappa\log d}{d}} + \frac{64\ell\log d}{3} \leq 8\ell\sqrt{\frac{n\kappa\log d}{d}}\;,
\end{align}
where the second inequality follows from the assumption that $n\ell \geq 2^{12}d\log d$. 

\subsubsection{Proof of Lemma \ref{lem:prob_bottomlbound}}
Without loss of generality, assume that $\i < i$, i.e., $\ltheta^*_{\i} \leq \ltheta^*_i$. Define $\Omega$ such that $\Omega = \{j: j\in S, j \neq i,\i\}$. 
For any $\beta_1 \in [0,(\ell-2)/\ell]$, 
define event $E_{\beta_1}$ that occurs if in the randomly chosen set $S$ there are at most $\floor{\ell\beta_1}$ items that have preference scores less than $\ltheta^*_i$, i.e., 
\begin{align} \label{eq:bl_prob_7}
E_{\beta_1}  \; \equiv \; \Big\{\textstyle\sum_{j \in \Omega} \I_{\{\ltheta^*_i > \ltheta^*_j\}} \leq \floor{\ell\beta_1} \Big\} \;.
\end{align}
We have,
\begin{eqnarray} \label{eq:bl_prob_1}
&&\P\Big[\sigma^{-1}(i),\sigma^{-1}(\i)  \;>\;  \kappa - \ell \;\Big|\; i,\i \in S\Big] \nonumber\\
&>&  \P\Big[\sigma^{-1}(i),\sigma^{-1}(\i) > \kappa - \ell \;\Big|\; i,\i \in S; E_{\beta_1} \Big] \P\Big[E_{\beta_1} \;\Big|\; i,\i \in S\Big]
\end{eqnarray}
The following lemma provides a lower bound on $\P[\sigma^{-1}(i),\sigma^{-1}(\i) > \kappa - \ell \;|\; i,\i \in S; E_{\beta_1}]$. 
\begin{lemma} \label{lem:bl_prob1}
Under the hypotheses of Lemma \ref{lem:prob_bottomlbound}, 
\begin{align} \label{eq:bl_prob2}
\P\Big[\sigma^{-1}(i),\sigma^{-1}(\i) \; >  \; \kappa - \ell \;\Big|\; i,\i \in S; E_{\beta_1}  \Big] \;\geq\;  \frac{e^{-4b}(1-\floor{\ell\beta_1}/\ell)^2}{2}\frac{\ell^2}{\kappa^2}\;.
\end{align}
\end{lemma}
Next, we provide a lower bound on $\P[E_{\beta_1} \;|\;i,\i \in S]$. 
Fix $i,\i$ such that $i,\i \in S$. Selecting a set uniformly at random is probabilistically equivalent to selecting items one at a time uniformly at random without replacement. Without loss of generality, assume that $i,\i$ are the $1$st and $2$nd pick. Define Bernoulli random variables $Y_{\j}$ for $ 3 \leq \j \leq \kappa$ corresponding to the outcome of the $\j$-th random pick from the set of $(d-\j-1)$ items to generate the set $\Omega$ such that $Y_{\j} = 1$ if and only if $\ltheta^*_{i} > \ltheta^*_{\j}$. 

Recall that $\gamma_{\beta_1} \equiv \ld (\kappa-2)/(\lfloor \ell\beta_1 \rfloor+1 )(d-2) $ and 
$\eta_{\beta_1} \equiv (\lfloor \ell \beta_1 \rfloor +1)^2/2(\kappa-2)$.
Construct Doob's martingale $(Z_2,\cdots,Z_{\kappa})$ from $\{Y_{\k}\}_{ 3 \leq \k \leq \kappa}$ such that $Z_{\j} = \E[\sum_{\k=3}^{\kappa} Y_{\k}\;|\;Y_3,\cdots,Y_{\j}]$, for $2 \leq \j\leq \kappa$. Observe that, $Z_2 = \E[\sum_{\k=3}^{\kappa} Y_{\k}] \leq \frac{(i-2)(\kappa-2)}{d-2} \leq \gamma_{\beta_1}(\floor{\ell\beta_1}+1)$, where the last inequality follows from the assumption that $i \leq \ld $. Also, $|Z_{\j} - Z_{\j-1}| \leq 1$ for each $\j$.  Therefore, we have
\begin{eqnarray}\label{eq:bl_prob_3}
\P\Big[\textstyle\sum_{j \in \Omega} \I_{\{\ltheta^*_{i} > \ltheta^*_j\}} \leq \floor{\ell\beta_1} \Big]  & = &\P\Big[ \textstyle\sum_{\j=3}^{\kappa} Y_{\j} \leq \floor{\ell\beta_1} \Big]\nonumber\\
& = & 1 - \P\Big[ \textstyle\sum_{\j=3}^{\kappa} Y_{\j} \geq \floor{\ell\beta_1} +1 \Big] \nonumber\\
& \geq & 1 - \P\Big[Z_{\kappa-2} - Z_2 \geq (\ell\beta_1 +1) - \gamma(\floor{\ell\beta_1}+1) \Big] \nonumber\\ 
& \geq & 1 - \exp\Big(-\frac{(\floor{\ell\beta_1}+1)^2(1-\gamma_1)^2}{2(\kappa-2)} \Big) \nonumber\\
& = & 1 - \exp\Big(-\eta_{\beta_1}(1-\gamma_{\beta_1})^2\Big),
\end{eqnarray}
where the inequality follows from the Azuma-Hoeffding bound. Since, the above inequality is true for any fixed $i,\i \in S$, for random indices $i,\i$ we have $ \P[E_{\beta_1} \;|\; i,\i \in S] \geq 1 - \exp(-\eta_{\beta_1}(1-\gamma_{\beta_1})^2)$. Claim \eqref{eq:prob_bottomlbound_eq} follows by combining Equations \eqref{eq:bl_prob_1}, \eqref{eq:bl_prob2} and \eqref{eq:bl_prob_3}.

\subsubsection{Proof of Lemma \ref{lem:bl_prob1}}
Without loss of generality, assume that $\i < i$, i.e., $\ltheta^*_{\i} \leq \ltheta^*_i$. 
Define $\Omega = \{j: j\in S, j \neq i,\i\}$, and event 
$E_{\beta_1} = \{ i,\i \in S; \textstyle\sum_{j \in \Omega} \I_{\{\ltheta^*_i > \ltheta^*_j\}} \leq \floor{\ell\beta_1} \}$. 
Since set $S$ is chosen randomly, $i,\i$ and $j \in \Omega$ are random. 
Throughout this section, we condition on the 
random indices $i,\i$ and the set $\Omega$ such that event $E_{\beta_1}$ holds. 
To get a lower bound on $\P[\sigma^{-1}(i),\sigma^{-1}(\i) > \kappa - \ell ]$, define independent exponential random variables $X_j \sim \exp(e^{\ltheta^*_j})$ for $j \in S$.   Observe that given event $E_{\beta_1}$ holds, there exists a set $\Omega_1 \subseteq \Omega$ such that 
\begin{align} \label{bl_prob_8}
\Omega_1 = \Big\{j\in S :\ltheta^*_i \leq \ltheta^*_j \Big\}\;,
\end{align}
and $|\Omega_1| = \kappa-\floor{\ell\beta_1} -2$. In fact there can be many such sets, and for the purpose of the proof 
we can choose one such set arbitrarily. 
Note that $\floor{\ell\beta_1} +2 \leq \ell$ by assumption on $\beta_1$, so $|\Omega_1| \geq \kappa-\ell$. 
From the Random Utility Model (RUM) interpretation of the PL model,
we know that the PL model is equivalent to ordering the items as per {\em random cost} of each item drawn from 
exponential random variable with mean $e^{\tilde\theta^*_i}$. 
That is, we rank items according to $X_j$'s such that the lower cost items are ranked higher. 
From this interpretation,  we have that
\begin{eqnarray} \label{eq:bl_prob_4}
\P\Big[\sigma^{-1}(i),\sigma^{-1}(\i) > \kappa - \ell \Big] &=& \P\Big[ \sum_{j \in \Omega} \I_{\big\{\min\{X_i, X_{\i}\}\; > \;X_j\big\}} \geq \kappa-\ell \Big] \nonumber\\
&>& \P\Big[ \sum_{\j \in \Omega_1} \I_{\big\{\min\{X_i, X_{\i}\} \;>\; X_{\j}\big\}} \geq \kappa-\ell \Big] 
\end{eqnarray} 
The above inequality follows from the fact that $\Omega_1 \subseteq \Omega$ and $|\Omega_1| \geq \kappa -\ell$. It excludes some of the rankings over the items of the set $S$ that constitute the event $\{\sigma^{-1}(i),\sigma^{-1}(\i) > \kappa - \ell \}$. Define $\Omega_2 = \{\Omega_1,i,\i\}$. Observe that items $i,\i$ have the least preference scores among all the items in the set $\Omega_2$. Therefore, the term in Equation \eqref{eq:bl_prob_4} is the probability of the least two preference score items in the set $\Omega_2$, that is of size $(\kappa-\floor{\ell\beta_1})$, being ranked in bottom $(\ell-\floor{\ell\beta_1})$ positions. 

The following lemma shows that the probability of the least two preference score items in a set being ranked at any two positions is lower bounded by their probability of being ranked at $1$st and $2$nd position. 
\begin{lemma}\label{lem:bl_prob2}
Consider a set of items $S$ and a ranking $\sigma$ over it. Define $i_{\min_1} \equiv \arg \min_{i \in S} \theta_i$, $i_{\min_2} \equiv \arg \min_{i \in S\setminus i_{min_1}} \theta_i$. For all $ 1 \leq i_1, i_2 \leq |S|$, $i_1 \neq i_2$, following holds: 
\begin{align}
\P\Big[ \sigma^{-1}(i_{\min_1}) = i_1, \sigma^{-1}(i_{\min_2}) = i_2 \Big] \geq \P\Big[ \sigma^{-1}(i_{\min_1}) = 1, \sigma^{-1}(i_{\min_2}) = 2 \Big]. 
\end{align}
\end{lemma}
Using the fact that $\i = \arg \min_{j \in \Omega_2} \ltheta^*_j $, $i = \arg \min_{j \in \Omega_2 \setminus \i} \ltheta^*_j$, for all $1 \leq i_1, i_2 \leq \kappa-\floor{\ell\beta_1}$, $i_1 \neq i_2$, we have that
\begin{align}\label{eq:bl_prob_5}
\P\Big[ \sigma^{-1}(\i) = i_1, \sigma^{-1}(i) = i_2 \Big] \geq \P\Big[ \sigma^{-1}(\i) = 1, \sigma^{-1}(i) = 2 \Big] \geq e^{-4b}\frac{1}{\kappa^2}\;, 
\end{align}
where the second inequality follows from the definition of the PL model and the fact that $\ltheta^* \in \lOmega_{2b}$.
Together with Equation \eqref{eq:bl_prob_5} and the fact that there are a total of $(\ell-\floor{\ell\beta})(\ell-\floor{\ell\beta}-1) \geq (\ell-\floor{\ell\beta})^2/2$ pair of positions that $i,\i$ can occupy in order to being ranked in bottom $(\ell-\floor{\ell\beta})$, we have,
\begin{align}
\P\Big[\sigma^{-1}(i),\sigma^{-1}(\i) > \kappa - \ell \Big] \geq \frac{e^{-4b}(1-\floor{\ell\beta_1}/\ell)^2}{2}\frac{\ell^2}{\kappa^2}.
\end{align}
Since, the above inequality is true for any fixed $i,\i$ and $j \in \Omega$ such that event $E$ holds, it is true for random indices $i,\i$ and $j \in \Omega$ such that event $E$ holds, hence the claim is proved.

\subsubsection{Proof of Lemma \ref{lem:bl_prob2}}
Let $\widehat{\sigma}$ denote a ranking over the items of the set $S$ and $\P[\widehat{\sigma}]$ be the probability of observing $\widehat{\sigma}$. Let
\begin{align}
\widehat{\Omega}_1 = \Big\{ \widehat{\sigma}: \widehat{\sigma}^{-1}(i_{\min_1}) = i_1, \widehat{\sigma}^{-1}(i_{\min_2}) = i_2 \Big\} \;\; \text{and} \;\; \widehat{\Omega}_2 = \Big\{ \widehat{\sigma} : \sigma^{-1}(i_{\min_1}) = 1, \sigma^{-1}(i_{\min_2}) = 2 \Big\}.
\end{align}
Now, take any ranking $\widehat{\sigma} \in \widehat{\Omega}_1$ and construct another ranking $\widetilde{\sigma}$ from $\widehat{\sigma}$ as following.  If $i_1 =2, i_2 = 1$, then swap the items at $i_1$-th and $i_2$-th position in ranking $\widehat{\sigma}$ to get $\widetilde{\sigma}$. Else, if $i_1 < i_2$, then first: swap items at $i_1$-th position and $1$st position, and second: swap items at $i_2$-th position and $2$nd position, to get $\widetilde{\sigma}$; if $i_2 < i_1$, then first: swap items at $i_2$-th position and $2$nd position, and second: swap items at $i_1$-th position and $1$st position, to get $\widetilde{\sigma}$.  

Observe that  $\P[\widetilde{\sigma}] \leq \P[\widehat{\sigma}]$ and $\widetilde{\sigma}_1^\ell \in \widehat{\Omega}_2$. Moreover, such a construction gives a bijective mapping between $\widehat{\Omega}_1$ and $\widehat{\Omega}_2$. Hence, the claim is proved.

\section*{Acknowledgements} 
The authors thank the anonymous reviewers for their constructive feedback. 
This work was partially supported by National Science Foundation Grants 
MES-1450848, CNS-1527754, and CCF-1553452. 

\bibliographystyle{plain}
\bibliography{_ranking}

\end{document}